\documentclass[sigconf,9pt]{acmart} % revision和Carmen-Ready的版本可以用9pt,一开始用10pt
\usepackage{threeparttable}
\usepackage{latexsym}
\usepackage{graphicx}
\usepackage{amsmath}
\usepackage{multirow}
\usepackage{booktabs}
\usepackage{comment}
\usepackage[linesnumbered,ruled,vlined]{algorithm2e}
\usepackage{makecell}
\usepackage{newtxmath}
\usepackage{placeins}
\usepackage{booktabs} % For professional looking tables
\usepackage{array}    % For table spacing
\usepackage{pifont}

\usepackage{subcaption}
\usepackage{balance} % For balanced columns on the last page

\newcommand{\circled}[1]{\ding{\numexpr171+#1\relax}}

\def\ourmodel{\textit{FedHybrid }}
\begin{document}
\acmYear{2024}\copyrightyear{2024}
\acmConference[SenSys '24]{ACM Conference on Embedded Networked Sensor Systems}{November 4--7, 2024}{Hangzhou, China}
\acmBooktitle{ACM Conference on Embedded Networked Sensor Systems (SenSys '24), November 4--7, 2024, Hangzhou, China}
\acmDOI{10.1145/3666025.3699346}
\acmISBN{979-8-4007-0697-4/24/11}
% \title{Breaking the Memory Constraint for Heterogeneous Federated Learning via Hierarchical Coordination}
\title{FedHybrid: Breaking the Memory Wall of Federated Learning via Hybrid Tensor Management}
% \title{FedHyb: Heterogeneity-Aware Memory Efficient Federated Learning via Hybrid Scheduling}

% \author{Kahou Tam$^{1}$, Chunlin Tian$^{1}$, Li Li$^{1}$}
% \authornote{Corresponding author. Email: llili@um.edu.mo} 
% \author{Haikai Zhao$^{2}$, ChengZhong Xu$^{1}$}
% % \author{Kahou Tam$^{1}$, Chunlin Tian$^{1}$, Li Li$^{1}$\authornote{Corresponding author.}, Haikai Zhao$^{2}$, ChengZhong Xu$^{1}$}
% \affiliation{%
%  \institution{$^1$ State Key Laboratory of IoTSC, University of Macau, Macau SAR, China}
%  \institution{$^2$ Simon Fraser University, Canada}
% }
\author{Kahou Tam}
\orcid{0000-0001-5816-6837}
\affiliation{%
   \position{State Key Laboratory of IoTSC} 
  \institution{University of Macau}
  \state{Macau SAR}
  \country{China}
}
\email{tamkahou.2023@connect.um.edu.mo}

\author{Chunlin Tian}
\orcid{0009-0009-5220-1609}
\affiliation{%
\position{State Key Laboratory of IoTSC} 
  \institution{University of Macau}
  \state{Macau SAR}
  \country{China}}
\email{yc27402@um.edu.mo}

\author{Li Li}
\authornote{Corresponding author}
\orcid{0000-0002-2044-8289}
\affiliation{%
\position{State Key Laboratory of IoTSC} 
  \institution{University of Macau}
  \state{Macau SAR}
  \country{China}}
\email{llili@um.edu.mo}

\author{Haikai Zhao}
\orcid{0009-0001-3830-4275}
\affiliation{%
  \institution{Simon Fraser University}
  % \state{Macau SAR}
  \country{Canada}}
\email{hza214@sfu.ca}

\author{ChengZhong Xu}
\orcid{0000-0001-9480-0356}
\affiliation{%
\position{State Key Laboratory of IoTSC} 
  \institution{University of Macau}
  \state{Macau SAR}
  \country{China}}
\email{czxu@um.edu.mo}

% \orcid{0000-0001-5816-6837}
% \affiliation{%
%   \institution{University of Macau}
%   % \city{Dublin}
%   \state{Macau SAR}
%   \country{China}
% }
% \email{tamkahou.2023@connect.um.edu.mo}

% \affiliation{%

% \affiliation{%

% \affiliation{%

% \affiliation{%

% \affiliation{%

\begin{abstract}
Federated Learning (FL) emerges as a new learning paradigm that enables multiple devices to collaboratively train a shared model while preserving data privacy. 
However, one fundamental and prevailing challenge that hinders the deployment of FL on mobile devices is the memory limitation. 
This paper proposes \textit{FedHybrid}, a novel framework that effectively reduces the memory footprint during the training process while guaranteeing the model accuracy and the overall training progress. 
Specifically, \textit{FedHybrid} first selects the participating devices for each training round by jointly evaluating their memory budget, computing capability, and data diversity. 
After that, it judiciously analyzes the computational graph and generates an execution plan for each selected client in order to meet the corresponding memory budget while minimizing the training delay through employing a hybrid of recomputation and compression techniques according to the characteristic of each tensor. 
During the local training process, \textit{FedHybrid} carries out the execution plan with a well-designed activation compression technique to effectively achieve memory reduction with minimum accuracy loss. We conduct extensive experiments to evaluate \textit{FedHybrid} on both simulation and off-the-shelf mobile devices. The experiment results demonstrate that \textit{FedHybrid} achieves up to a 39.1\% increase in model accuracy and a 15.5$\times$ reduction in wall clock time under various memory budgets compared with the baselines.

% designed to enable heterogeneity-aware, memory-efficient FL. Unlike previous works that simply assume all the participating devices have sufficient memory to well support on-device local training, Harmony employs a hierarchical structure to conduct effective training on memory-constrained devices while guaranteeing the model accuracy and the overall training progress. Specifically, Harmony features three core components. 
% The Memory-aware Client Selector, deployed on the server side, first selects devices by jointly evaluating their memory capacity and computing capability along with data diversity. After that, the Heterogeneity-aware Graph Optimizer judiciously analyzes the computational graph and generates an execution plan for each selected client in order to accelerate the local training process while meeting the memory constraint. Then the Local Training Engine carries out the plans with advanced activation compression and activation recomputation to further boost runtime efficiency with minimum accuracy loss. We conduct extensive experiments to evaluate Harmony on both simulation and off-the-shelf mobile devices. The experiment results demonstrate that Harmony achieves up to a 39.1\% increase in final model accuracy and a 15.5-fold reduction in wall clock time under the various memory budgets compared with the baselines. 
\end{abstract}

\begin{CCSXML}
<ccs2012>
   <concept>
       <concept_id>10010147.10010919</concept_id>
       <concept_desc>Computing methodologies~Distributed computing methodologies</concept_desc>
       <concept_significance>500</concept_significance>
       </concept>
   <concept>
       <concept_id>10003120.10003138</concept_id>
       <concept_desc>Human-centered computing~Ubiquitous and mobile computing</concept_desc>
       <concept_significance>500</concept_significance>
       </concept>
 </ccs2012>
\end{CCSXML}

\ccsdesc[500]{Computing methodologies~Distributed computing methodologies}
\ccsdesc[500]{Human-centered computing~Ubiquitous and mobile computing}

\keywords{Federated learning, Mobile computing, Memory optimization.}

%% headers. Often, this list is too long and will overlap


\maketitle

\section{introduction}
Federated Learning (FL) \cite{mcmahan2017communication} coordinates multiple mobile devices to collaboratively train a shared model while preserving data privacy  \cite{kairouz2021advances}. 
Most existing FL approaches \cite{lai2021oort,paulik2021federated,shin2022fedbalancer,li2022pyramidfl} simply assume that all the participating clients have sufficient resources to update the local model with their own private data.
However, in real-world cases, a fundamental challenge that impedes the deployment of FL on mobile devices is memory limitation.  
% the devices are usually heterogeneous and have totally different computing resources and memory sizes. Meanwhile, 
During the local training process, the model weight, intermediate activation, and optimizer states are mandated to be stored in the memory. 
For instance, training MobileNetV2 \cite{sandler2018mobilenetv2}, specifically designed for on-device vision tasks, necessitates over 8GB of memory. 
% on ImageNet \cite{deng2009imagenet} 
However, commercial mobile devices typically only have 4GB to 12GB of RAM~\cite{ramstatisticsinfo}.
More importantly, in order to retrieve high analysis quality, the model architecture is becoming deeper and wider and the memory requirement keeps increasing. 
% which means the low-end devices cannot even participate in the training process. 
% We can find that the low-end devices cannot even afford to conduct local training due to memory constraints. 

% \textbf{Critical Issue.} Memory constraints of the participating devices not only degrade the training performance but also deteriorate the overall training progress at the same time. 
\textbf{The Memory Wall in FL.} Unfortunately, the memory limitation deteriorates the performance of FL  from multiple perspectives. 
% Unfortunately, memory constraints of mobile devices deteriorate the training process from multiple perspectives in real-world cases. 
First, the low-end devices cannot even afford to conduct local training. 
Directly dropping them leads to inferior model accuracy as the unique data on them are underrepresented in the global model, especially when the amount of low-end devices is relatively large.  
% In this case they cannot contribute to the global model with their own unique data, 
On the other side, although other devices can afford to conduct local training, the learning process can be severely slowed down.  
This is for the reason that when the memory footprint exceeds a predefined threshold \cite{lim2023swam}, it triggers the operating system to reclaim pages by moving some memory pages occupied by processes to the swap area on the memory or storage devices which leads to high training latency.
Worse still, memory contention caused by the currently running background apps can exacerbate page reclaiming and further slow down the local training process. This deceleration can be amplified in FL as the slowest devices bottleneck the overall convergence process. 
% as the training process proceeds, its continuous demand for memory can exceed around 60\%the system's memory total utilization, 
% In mobile devices, when the memory utilization becomes around 60\%, the ZRAM/Swap secures the available memory space by moving some memory pages occupied by processes to the swap area on the memory or storage devices prompting the operating system to reclaim pages by moving them between the faster main memory and the slower secondary storage.
% Moreover, in real-world scenarios, as the training process proceeds, there are often applications alive in the background in the mobile systems. The background applications consume a significant amount of memory resources, which raises the memory contention with the training process. Such concurrent exacerbates more pages of the training process to be evicted by page-reclaiming and further slows down the training procedure. 
Thus, the memory wall can simultaneously hurt the model performance and training efficiency in FL. 

\textbf{Prior Art.} 
Memory optimization has been widely studied in on-server training. Gradient checkpointing \cite{chen2016training,checkmate_mlsys2020_196,kirisame2020dynamic}, activation compression \cite{chen2021actnn,liu2022gact}, and swapping \cite{peng2020capuchin,wang2018superneurons,huang2020swapadvisor} have been widely employed. However, they rely on architectures that are invalid on mobile devices, such as host-side memory for swapping. Though achieving memory reduction, directly applying them either leads to inferior model accuracy or results in prolonged training progress in FL. 
Recently, in order to surmount the resource constraints of the participating devices, partial training is proposed \cite{diao2020heterofl,alam2022fedrolex,horvath2021fjord}. It first trains a lower-complexity submodel on the devices and then integrates the trained submodels into the full global model. 
 % partial training \cite{diao2020heterofl,alam2022fedrolex,horvath2021fjord} is proposed to surmount the resource constraints of participating devices in FL. The main idea is to train a lower-complexity submodel on the devices and integrate the trained submodels into the full global model. 
However, these approaches adversely affect model performance as numerous filters/parameters must be discarded and the architecture of the model is severely compromised. 
\textit{Thus, a new framework that can effectively guide the training process on memory-constrained devices while jointly guaranteeing the model accuracy and training efficiency is crucial for the deployment of FL in real-world cases. }
\textbf{Challenges.} Designing such a framework is not straightforward and faces the following critical challenges. First, memory constraints of the mobile devices simultaneously impact the model accuracy and training performance. Thus, how to jointly trade off the heterogeneity of memory capacity, computing capability and data distribution across different devices is the first critical challenge. Second, simply dropping the memory-constrained devices can severely hurt the model accuracy. Though directly applying the existing on-server memory-saving techniques, such as swapping and compression, can reduce the memory footprint, they can slow down the training progress or deteriorate the model performance at the same time. Thus, how to achieve memory-efficient training without compromising training efficiency is the second critical challenge. In addition, due to the resource contention caused by the background apps, the memory budget for the training process varies across different training rounds. Thus, how to conduct efficient training in a highly dynamic training environment is the third critical challenge.

\textbf{Our Contribution.} In this work, we propose \textit{FedHybrid}, a new framework that efficiently conducts FL on memory-limited mobile devices through hybrid tensor management. It aims to strike a balance among the 1) memory reduction of the local training process, 2) accuracy of the global model and 3) training efficiency of the overall system.
% To conduct efficient training on memory-constrained devices, in this work, we propose \ourmodel, a heterogeneity-aware hierarchical coordination framework for memory-efficient Federated Learning. It intelligently guides the training process to make it proceed in \ourmodel in order to strike a balance among the 1) memory reduction of the local training process, 2) accuracy of the global model and 3) training efficiency of the overall system. \ourmodel employs a hierarchical structure to conduct effective training on memory-constrained devices while guaranteeing the model accuracy and the overall training progress. 
Specifically, \ourmodel consists of the following three core components. In each training round, the Memory-aware Client Selector first selects the participating devices by jointly evaluating their memory budget, and computing capability along with data diversity. 
% \textcolor{blue}{This selection process employs a fine-grained utility design that ensures even clients with insufficient memory for local training may be selected if their data contribution significantly outweighs their system limitations.}
% After that, for the selected devices, the Heterogeneity-aware Graph Optimizer delicately analyzes the computational graph and generates an execution plan by configuring the \textit{hybird} memory reduction technique according to the characteristic of each tensor and considering the layout transformation-related overhead in order to conduct local training with the given memory budget without compromising the training efficiency. 
After that, for the selected devices, the Heterogeneity-aware Graph Optimizer delicately analyzes the computational graph and generates an execution plan tailored to the features of the mobile platform, which employs a judiciously designed memory optimization strategy through hybrid tensor management in order to meet the memory constraint with minimum training latency. 
After receiving the execution plan, the Local Training Engine directs the local training process with a well-designed activation compression strategy and the recomputation technique to further boost runtime efficiency with minimal accuracy loss. Additionally, this engine incorporates a novel memory budget predictor in order to retrieve the safe memory budget in a dynamic training environment. 
% that integrates mobile dynamic usage features. This predictor provides more accurate forecasts of memory requirements and ensures safe available memory during training, thus optimizing the balance between training efficiency and memory utilization. }
% Our insight is that outlier values in tensors are crucial for ensuring model accuracy during training, a factor overlooked by existing compression methods. Our proposed activation compression method preserves these outlier values while maintaining a high compression ratio and speed. Additionally, we design a new memory budget predictor that integrates mobile dynamic usage features to more accurately forecast memory requirements and provide safe available memory during training, effectively balancing training efficiency and memory usage. 
% This process iterates until the model converges. 
To the best of our knowledge, \ourmodel is the \textbf{\textit{first}} work that conducts memory-efficient FL on a highly heterogeneous and dynamic training environment. 
Specifically, we make the following key contributions: 

$\bullet$ We propose \textit{FedHybrid}, a memory-efficient federated learning framework, in which a hierarchical structure is designed to well balance memory reduction, model accuracy, and the overall training process.

$\bullet$ We design and implement the three core components, the Memory-aware Client Selector, the Heterogeneity-Aware Graph Optimizer, and the Local Training Engine to interact with each other and guide the whole training process. 

$\bullet$ To evaluate the effectiveness of \textit{FedHybrid}, we conduct extensive experiments based on representative DNN models, datasets, and commodity mobile devices.

\section{Background and Motivations}
\begin{figure}[!t] 
	\centering
	\includegraphics[width=0.99\linewidth]{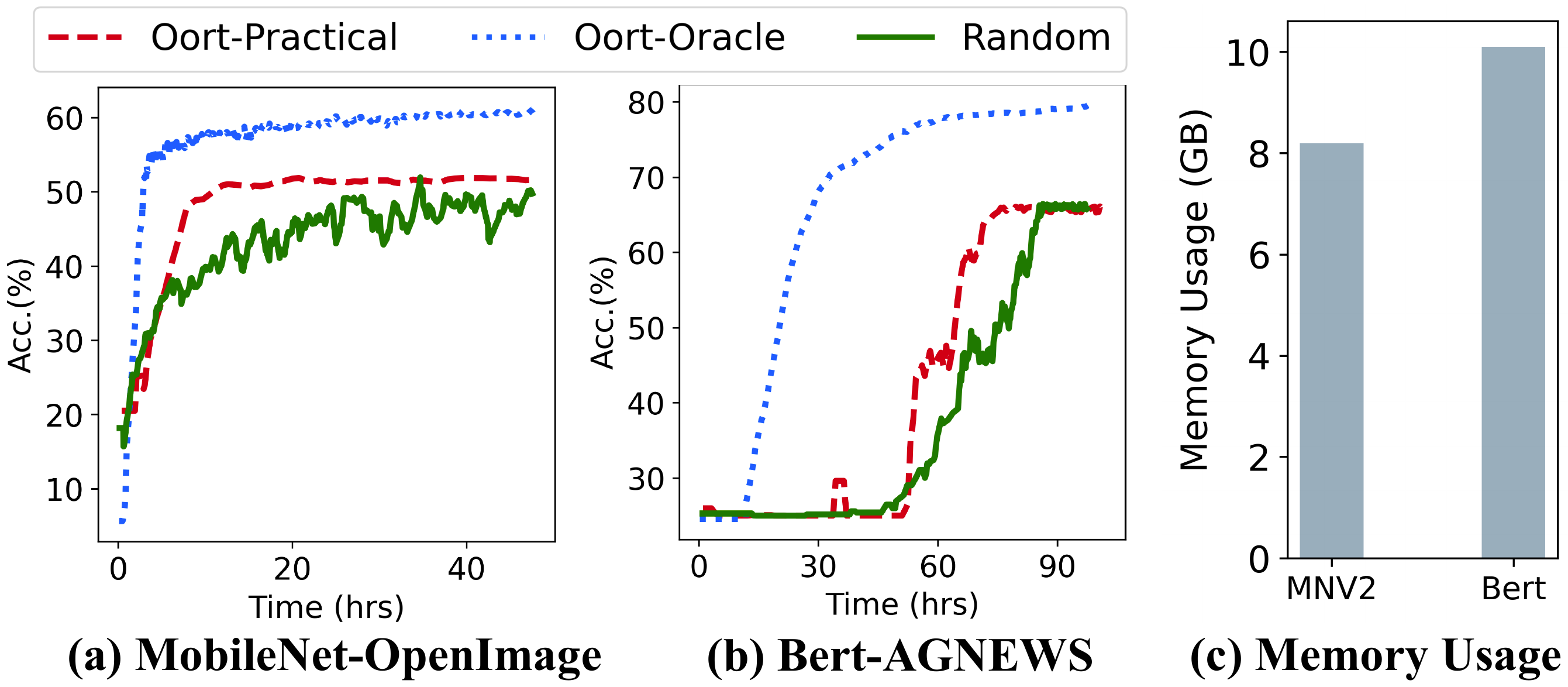}
        \vspace{-10pt}
	\caption{Performance impact of memory constraints on Oort and memory utilization for different models. (a) Accuracy of MobileNetV2 trained on OpenImage with a batch size of 32. (b) Accuracy of Bert trained on AGNEWS with a batch size of 8. (c) Memory utilization on mobile devices during training, measured with MNN\cite{alibaba2020mnn}; MNV2 denotes MobileNetV2.}
	\label{memory_motivation1} 
 \vspace{-15pt}
\end{figure}

\vspace{-0.5em}
\subsection{The Memory Wall in FL}\label{memorywallexp}
One pivotal question required to be explored first is: \textit{how does the memory limitation impact a heterogeneous FL system?} In order to investigate the impact, we conduct the following experiments. 
Specifically, we establish a hybrid FL platform with both simulation and hardware testbed to conduct the following two training tasks:
MobileNetV2 \cite{sandler2018mobilenetv2} on OpenImage \cite{kuznetsova2020open} for image classification and Bert \cite{lan2019albert} on AGNEWS \cite{zhang2015character} for text classification. Leveraging the FedScale \cite{lai2022fedscale} benchmark to simulate the real-world environment, the system contains 6,582 clients for image classification and 2,040 clients for text classification, respectively. 
The hardware configuration including the computing capability and the memory capacity are emulated with the data from AI Benchmark \cite{AIBenchmark}, which provides statistical runtime information of the training process across different types of mobile devices. 
In specific, the memory distributions are as follows: 4GB (15\%), 6GB (25\%), 8GB (30\%), 12GB (25\%), and 16GB (5\%). 
Moreover, off-the-shelf mobile devices, including S22 and Oneplus 10-Pro, are adopted to delve into the impact of memory limitation on the runtime.

\textbf{Observation 1: Memory wall deteriorates the model performance.}
Figure~\ref{memory_motivation1} presents the training performance in three different scenarios including: 1) Oort-
Practical, 2) Oort-Oracle, and 3) Random. 
Specifically, Oort \cite{lai2021oort} is a client selection methodology that jointly considers data and system heterogeneity.
In this case, Oort-Practical directly applies Oort on the memory-constrained devices, whereas Oort-Oracle represents a theoretical baseline, which assumes that all the devices have sufficient memory to complete local training. 
For Random, it just randomly selects a certain number of clients to participate in each training round. It is worth noting that 200 clients are selected in each training round. 
Figure \ref{memory_motivation1} shows that, compared with Oort Ideal, Oort-Practical shows a prominent accuracy decline, 10.8\% for image classification and 14.2\% for text classification. This is for the reason that, as shown in Figure \ref{memory_motivation1}(c), over 8GB of memory is required for MobileNetV2, and 10GB is required for Bert. In this case,  Oort-Practical excludes over 40\% and 70\% of clients for MobileNet and Bert training. This exclusion not only reduces data diversity but also compromises the comprehensiveness and representativeness of the training data.

\begin{figure}[!t] 
	\centering
	\includegraphics[width=0.99\linewidth]{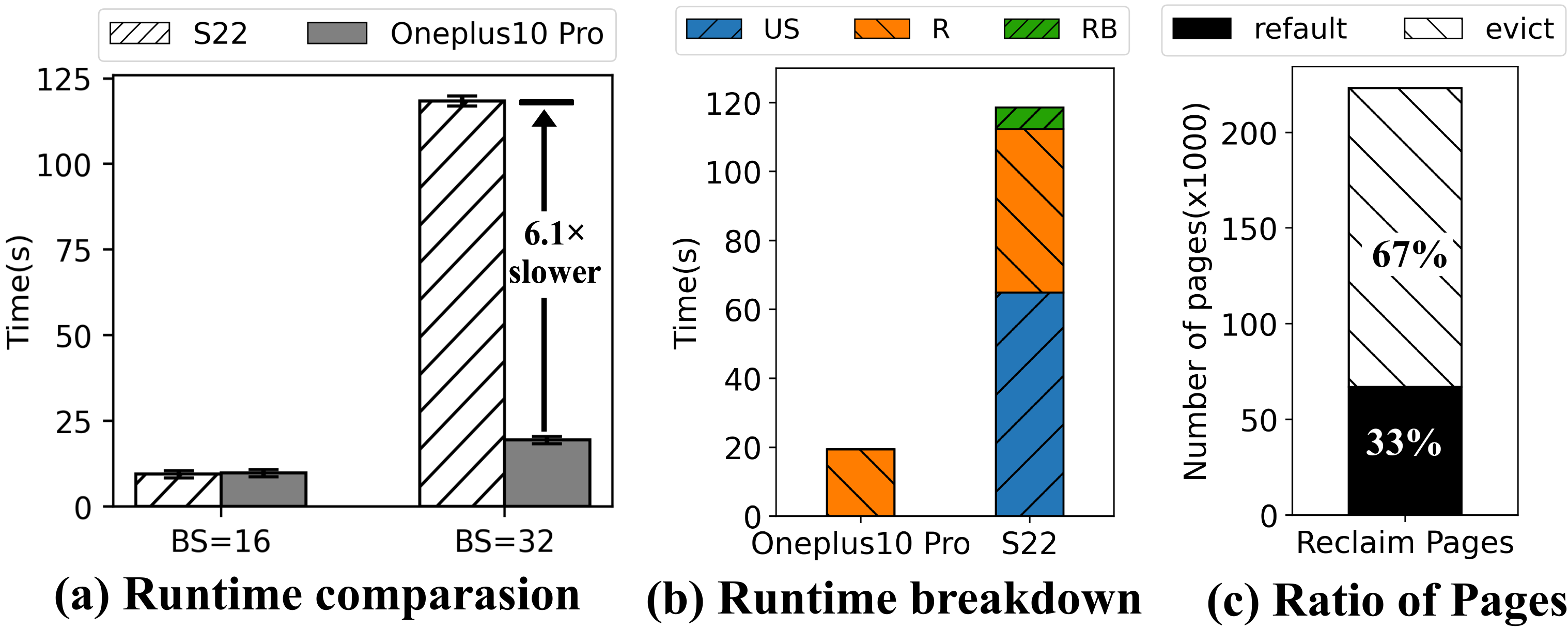}
 \vspace{-10pt}
	\caption{The analysis of local training runtime on mobile devices. We use MNN to conduct the local training without any background application.
(a) Compare the runtimes of different devices under 16 and 32 batch sizes.
(b) Breakdown of the training process's runtime status and average system memory usage during training with 32 batch sizes. The US represents the Uninterruptible Sleep status, R represents the Running status, and RB represents the Runnable status.
(c) Distribution of evicted pages during the training process with 32 batch sizes in S22.
 %96666count
 }
	\label{runtimeoverhead} 
 \vspace{-14pt}
\end{figure}

% \subsubsection{Memory Contention Extend Training's Runtime.}
\textbf{Observation 2: Memory wall impedes the training efficiency.}
Memory constraints not only affect training performance, but also significantly degrade overall training efficiency. 
%改写成training round的时间 5.3min 20.2  117min 280.8 average training round time 
We observe that the average duration for each training round with Oort-Practical extends $3.8 \times$ longer for MobileNetV2 and $2.4 \times$ longer for Bert compared with Oort-Oracle to attain the respective target accuracy.
% As shown in Figure \ref{memory_motivation1}, Oort-Practical takes over $3.8 \times$ longer to train MobileNetV2 (Figure \ref{memory_motivation1}(a)) and $2.4 \times$ longer to train Bert (Figure \ref{memory_motivation1}(b)) than Oort-Oracle to the target accuracy.
To investigate the root cause, we employ the MNN \cite{alibaba2020mnn} framework to perform local training on commercial mobile devices, including Galaxy S22 (8GB) and OnePlus 10 Pro (12GB).  

Figure \ref{runtimeoverhead}(a)  presents the local training completion time of one training round on the two different smartphones, utilizing different batch sizes. 
With a batch size of 16, the runtime for these devices is nearly identical, reflecting their comparable computation capability. 
However, a significant discrepancy is observed when the batch size is increased to 32.
The training completion time of S22 significantly exceeds that of the OP10P, being over 6.1 times slower. 
% However, a significant discrepancy is observed with a batch size of 32, the S22's training completion time significantly exceeds that of the OP10P, being over 6.1 times slower. 
% To further investigate the dynamics of the training process's runtime at this larger batch size, we utilize Perfetto \cite{perfetto} to trace and analyze the scheduling and status of the process within the CPU. This approach enables a precise assessment of CPU engagement and workload distribution, providing a comprehensive insight into the process's interaction with the CPU resources.
Figure \ref{runtimeoverhead}(b) shows the duration the CPU spends in different statuses within the local training procedure including: 1) uninterruptible sleep (UR), which refers to a state in which the process awaits a hardware resource and cannot be interrupted 2) active running (R) denotes that the training process is currently executing on the CPU, and 3) runnable (RB) which signifies readiness for execution pending CPU availability. 
We can find that, for the S22, 56\% of CPU time is dedicated to uninterruptible sleep, 40\% to active running, and 4\% to a runnable state.
% As illustrated in Figure 2(b) \ref{runtimeoverhead}, our trace analysis reveals significant differences in CPU time allocation for the S22's training process: 56\% of CPU time is dedicated to uninterruptible sleep, 40\% to active running, and 4\% to a runnable state. This analysis encounters several CPU statuses critical for understanding the performance and efficiency of the training process. 'Uninterruptible sleep' refers to a state in which the process awaits a hardware condition and cannot be interrupted, often due to waiting for disk or network I/O operations to complete. 'Active running' denotes that the process is currently executing on the CPU, whereas a 'runnable' state signifies readiness for execution pending CPU availability.
The substantial time spent in uninterruptible sleep status, over 50\% for the S22, is primarily attributed to I/O operations related to memory management.  Concurrently, our assessment of average system memory usage indicates that it is at 74\% for the OP10P and increases to 91\% for the S22 as shown in Figure \ref{runtimeoverhead}(b).
% This high proportion suggests that I/O operations significantly impede the training process. 
Under conditions of high memory pressure, the system triggers page reclaim which identifies and reclaims inactive pages. The identified evicted pages are relocated to secondary storage,  and if needed again (`page refault'), they would be reintroduced into main memory, causing substantial delays due to I/O requests. Notably, Figure \ref{runtimeoverhead}(c) shows, S22 experiences 96,666 page refaults\footnote{We instrument the Android kernel source code (Linux kernel version 5.10.136) and use the adb to obtain information on memory allocations and the reclaim process of our evaluated smartphone. }, with 33\% necessitating re-access which prominently slows down the local training process. 
In addition to the substantial memory footprint leading to heightened memory pressure, memory contention from background applications can further exacerbate the runtime overhead. 

\noindent\textbf{Summarized Takeaway: } 
Memory constraints of the participating devices severely deteriorate the model performance and the training efficiency at the same time. Moreover, the apps concurrently running with the local training process can further exacerbate this situation. \textit{Thus, a new FL system that effectively coordinates the training on memory-limited devices with dynamic resource contention is crucial for real-world deployment. }

\begin{figure}[!t] 
        % \vspace{-10pt}
	\centering
	\includegraphics[width=0.8\linewidth]{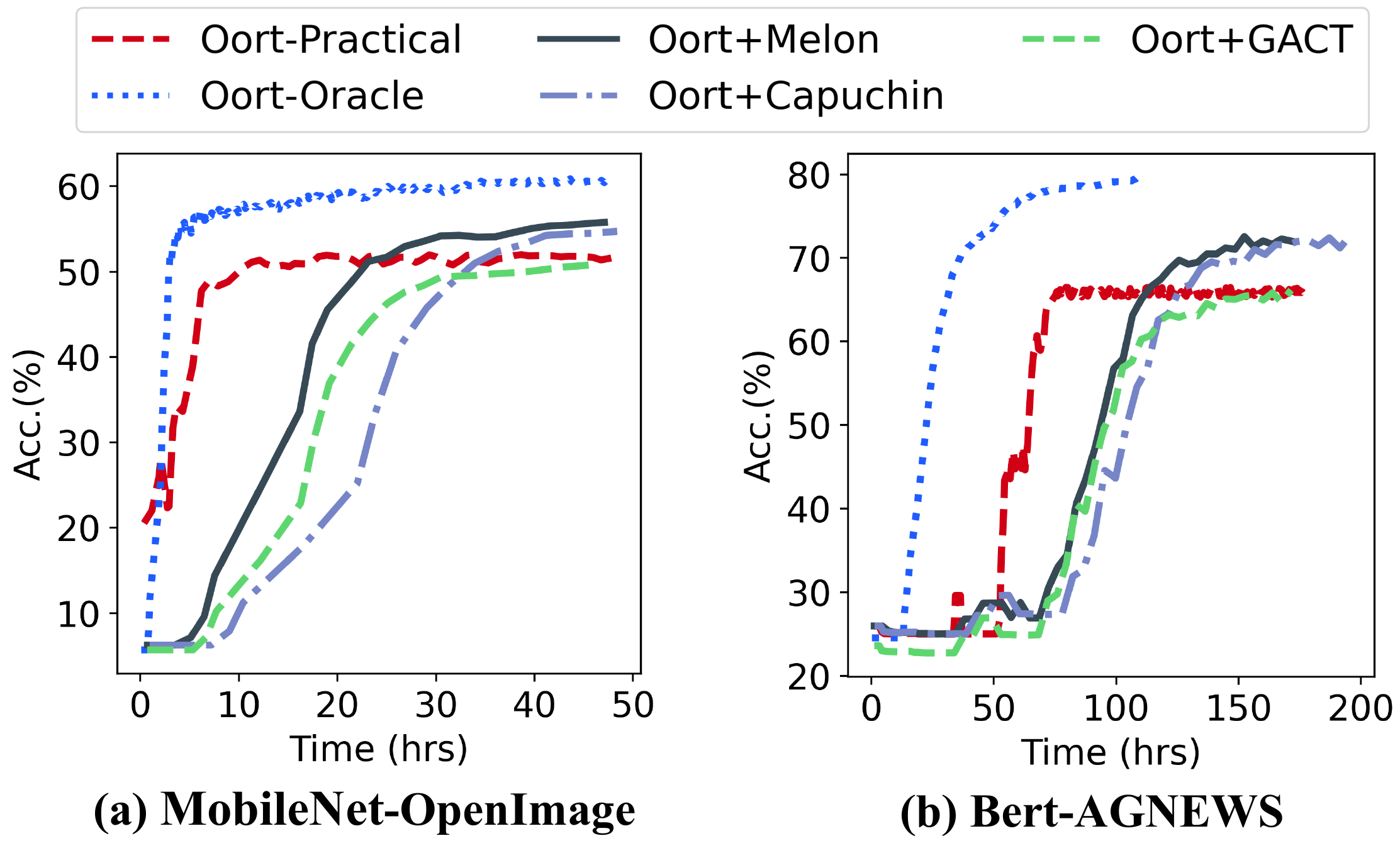}
        \vspace{-12pt}
	\caption{The performance of existing memory-saving techniques applied in FL with memory constraints.}
	\label{memoptmethod} 
 \vspace{-12pt}
\end{figure}

% \begin{figure}[] 
% 	\centering
% 	\includegraphics[width=1\linewidth]{Figures/directreclaim.pdf}

% 	\caption{The number of direct reclaims in the training process with varying numbers of background applications.}
% 	\label{directreclaim} 

% \end{figure}

% \subsection{Observation}
% 这里就需要讨论收益比，引出为什么要用重计算+压缩 ,现在要不
% 1. selector 2. 收益比相加 3. 优化时间  4.压缩
% \vspace{-5pt}
\section{Exploring Existing Memory Saving Techniques in FL}
\label{section:limitation_existing_memory}
Another important question yet to be explored is: \textit{whether the existing memory saving techniques are sufficient to break the memory wall in FL? } In this section, we investigate the following techniques that are widely adopted in on-server training.
% In this section, we aim to answer the following question: \textit{Can existing memory reduction techniques designed for on-server training can be directly applied to FL?} 
% Specifically, the same setting as Section \ref{memorywallexp}. 
% Different memory reduction approaches are directly applied to each client to reduce the memory consumption of local training.  

% first examine the existing memory saving techniques that are originally designed for the cloud

\begin{figure}[!t] 
	\centering
        % \vspace{-10pt}
	\includegraphics[width=1\linewidth]{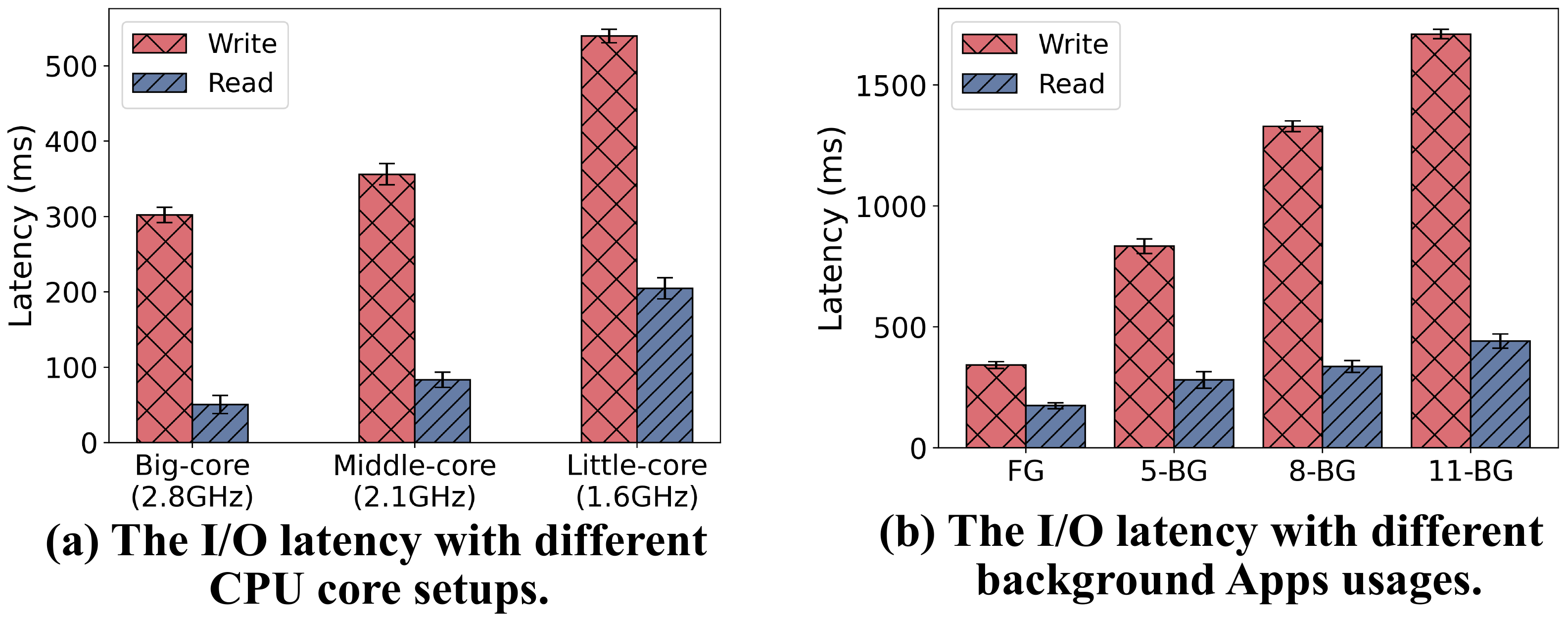}
        \vspace{-18pt}
	\caption{The I/O latency of write and read for a 128 MB File in UFS 3.1: simulation of swapping In and Out a 128 MB tensor. (a) A higher frequency of the CPU core correlates with increased I/O performance. (b) Multi-Apps deteriorate the I/O performance due to contention in the UFS command queue. }
	\label{iolatency} 
 \vspace{-17pt}
\end{figure}

$\bullet$\textbf{Host-device memory swapping.}
Swapping reduces memory usage during training by offloading activations or model parameters from GPU memory to external memory (e.g., CPU memory) in the cloud server. However, mobile devices predominantly utilize integrated memory for all processors, necessitating data swapping between main memory and storage disks. By applying Capuchin \cite{peng2020capuchin}, a widely utilized swapping mechanism designed in on-server training, we observe a notable improvement in accuracy by 4.7\% and 7.2\% for image and text classification tasks, respectively, as depicted in Figure \ref{memoptmethod}. 
This is for the reason that as the memory footprint is reduced, more low-end devices can be involved in the training process and the corresponding private local data can well benefit the global model. 
However, in the meantime, it extends the training duration required to achieve target accuracy by factors of 4.7$\times$ and 1.7$\times$, significantly degrading training efficiency. 
This inefficiency is primarily due to the instability of device I/O for data swapping in mobile environments. 
As shown in Figure \ref{iolatency}(a), the read/write latency is influenced by the CPU issuing the I/O command, with higher CPU core frequencies correlating with decreased I/O latency. 
% This is because the CPU core initiating the read/write needs to run the UFS driver; a higher frequency enables faster processing of UFS-related I/O operations, including handling interrupts and managing queues.
This is because the CPU core initiating the read/write needs to run the UFS driver, which involves several tasks such as handling interrupts and managing queues. A higher frequency allows faster processing of these UFS-related I/O operations.
Additionally, unlike NVMe, UFS storage in mobile devices has a single command queue, lacking internal concurrency capabilities. Current mobile devices frequently engage in background I/O activities, primarily due to the frequent reading and writing of cache files, which are first stored in main memory and then written back to flash storage \cite{liang2022cachesifter}. As shown in Figure \ref{iolatency}(b), such frequent I/O activities result in resource competition between swapping processes and background applications, pushing the hardware to its throughput limit and leading to high latency issues \cite{han2020command,lee2021internal,xue2024powerinfer}. 
Moreover, in FL, the variance in I/O speeds across different devices \cite{kairouz2021advances,liang2020acclaim} can exacerbate these performance bottlenecks.

$\bullet$\textbf{Activation compression.}\label{section:compression}
Activation compression compresses the training activation during the forward pass and then decompresses the saved activation during backpropagation to calculate the gradient. However, this approach requires detailed, model architecture-dependent analysis to minimize compression error, restricting their applicability. In this case, we incorporate GACT \cite{liu2022gact}, a model-universal framework for activation compression that adaptively modulates the compression ratio for each neural network layer. As illustrated in Figure \ref{memoptmethod}, integrating GACT does not lead to enhanced model accuracy, and it simultaneously degrades the training efficiency. 
This accuracy decline results from the error introduced during the compression process. 
While, the extra training latency lies in its uniform compression of all activations for gradient computations during backpropagation as some activations are too large to be processed quickly on mobile devices

\begin{figure}[!t] 
	\centering
        % \vspace{-10pt}
	\includegraphics[width=1\linewidth]{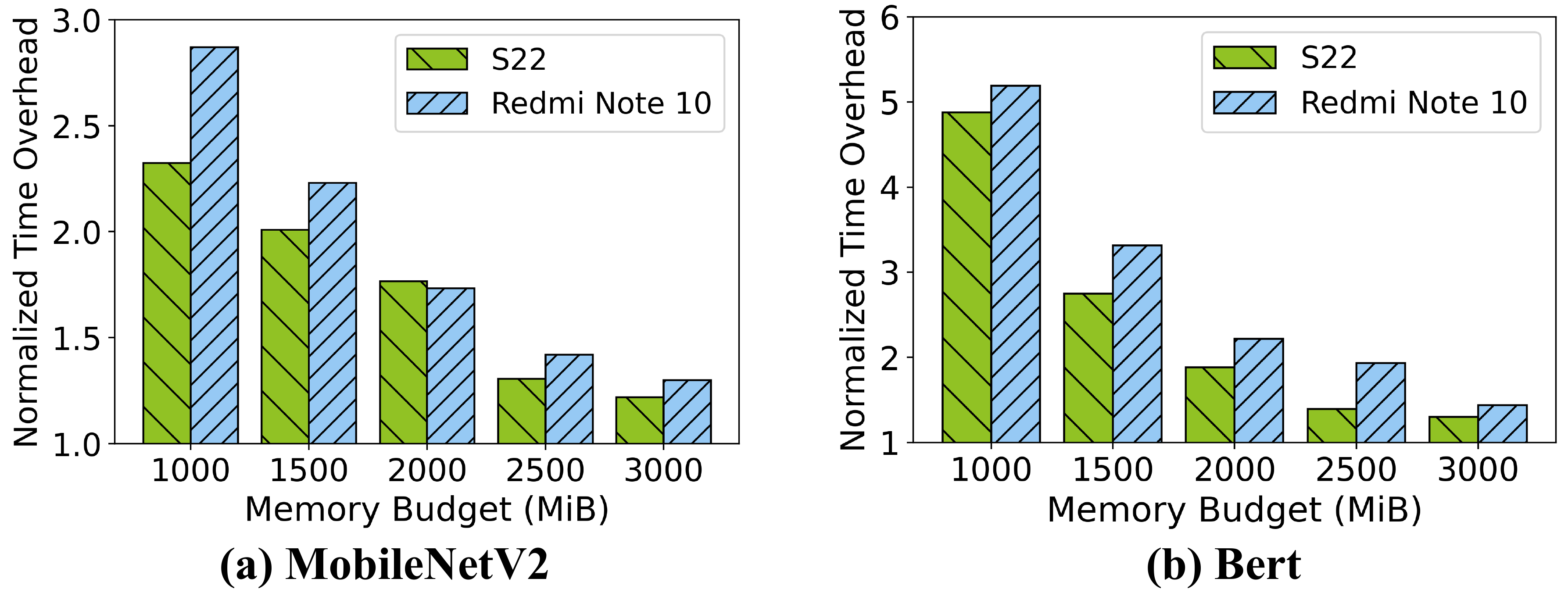}
        \vspace{-18pt}
	\caption{Efficiency analysis of activation recomputation in FL with heterogeneous devices. All the experiments are conducted on Melon \cite{wang2022melon}. Comparison of different memory budgets versus training latency overhead in MobileNetV2 with a batch size of 32, and BERT with a batch size of 8, on the S22 and Note 10, respectively. }
 % (c) An example illustrating the suboptimal memory budget during the training of MobileNetV2 on the S22. (d) The time cost of plan regeneration on the S22 and Note 10 for varying regeneration counts.}}
	\label{recompute_analysis} 
 \vspace{-10pt}
\end{figure}

$\bullet$\textbf{Activation recomputation.}
\label{sec:activation_recomp}
% 介绍melon
% that existing methods do not adequately account for the heterogeneous capabilities of devices, which affects overall FL training efficiency. Firstly, relying solely on recomputation for memory savings proves inefficient, especially in mobile environments under constrained memory conditions. 
% \textcolor{blue}{
Activation recomputation evicts activations during the forward pass and recomputes them as needed during the backward stage. Recent studies \cite{gim2022memory, wang2022melon} have used advanced recomputation methods to reduce the memory footprint for on-device training. Melon \cite{wang2022melon} is a representative recomputation-based approach, which combines recomputation and micro-batch while eliminating fragments to reduce the memory footprint. As illustrated in Figure \ref{memoptmethod}, integrating Melon results in a notable increase in accuracy 5\% and 8\% higher compared to Oort-Practical for training MobileNet and Bert, as more mobile devices can be involved in the training process. However, such a method leads to significantly prolongs the convergence process by 3.5$\times$ and 1.5$\times$, respectively. The primary reason for this slowdown is that through achieving memory reduction, the existing recomputation strategy in Melon involves computation-intensive operators that require repeated calculation and significantly extend the training time. This inefficiency is particularly pronounced when memory budgets are extremely limited. As illustrated in Figure \ref{recompute_analysis} (a) and (b), empirical evidence demonstrates substantial time overheads: the S22 device experiences 2.3$\times$ and 4.8$\times$ increases for MobileNetV2 and BERT respectively, with a memory budget of 1000. These overheads are even more significant for lower-end devices such as the Redmi Note 10, reaching 2.9$\times$ and 5.2$\times$ respectively. Thus, while local training can be completed within the memory budget, the prolonged duration can severely extend each training round. 

\noindent \textbf{Summarized Takeaway:} Though directly applying existing memory reduction techniques can achieve memory saving, it either degrades the model performance or deteriorates the training efficiency due to challenges posed by the unique architecture of mobile devices and an FL system. \textit{Thus, a memory-efficient training framework specifically designed for a mobile-based FL system is urgently required. } 

\vspace{-0.5em}
\section{System Design}
\ourmodel is designed to effectively deploy FL on memory limited mobile devices while jointly guaranteeing the model performance and training efficiency. 
In this section, we first give an overview of \ourmodel and then discuss each key component in detail. 
% \vspace{-1em}
\subsection{Overview}
\begin{figure}[!t] 
	\centering
	\includegraphics[width=\linewidth]{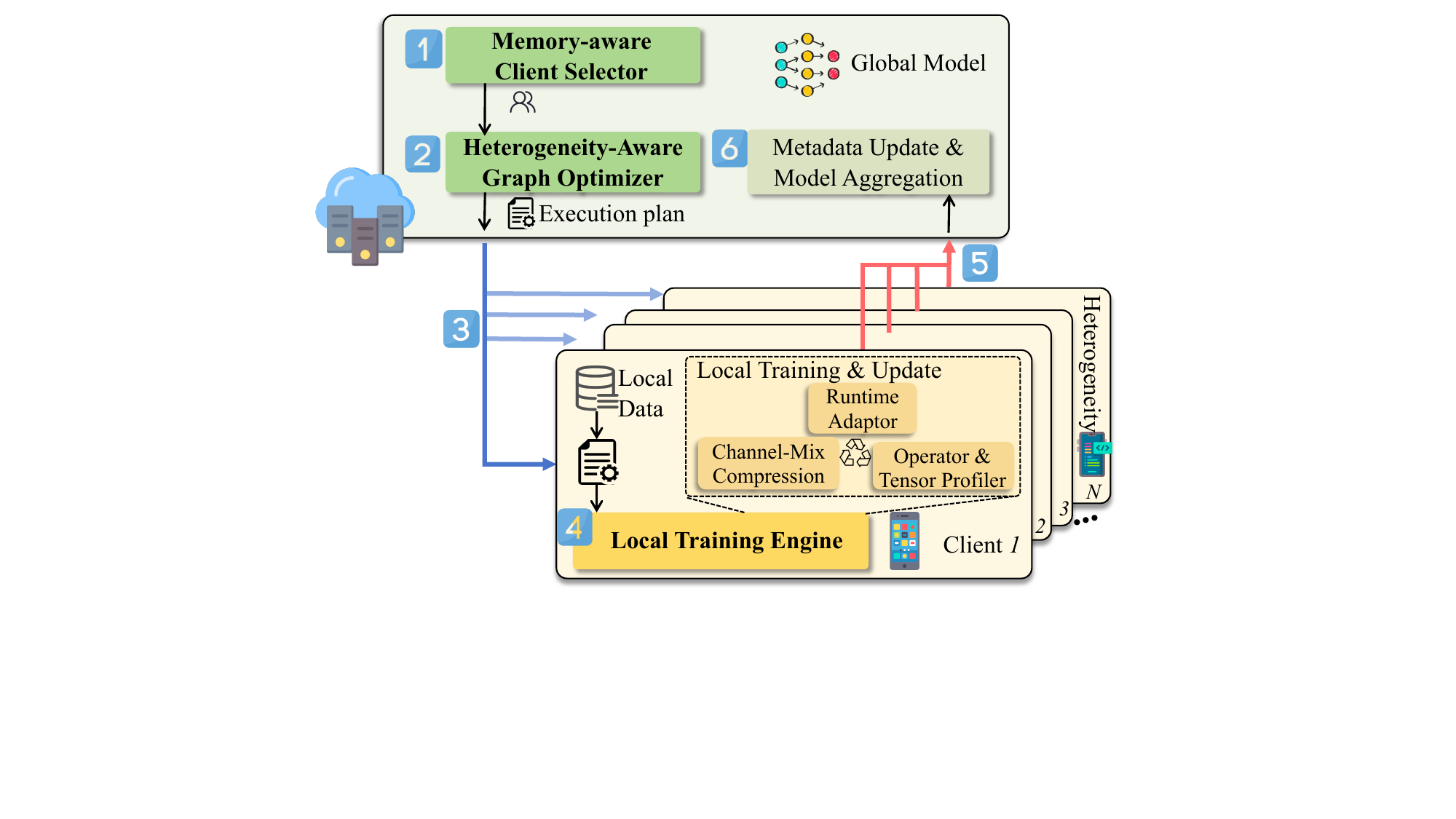}
  \vspace{-20pt}
	\caption{The workflow of \textit{FedHybrid}.	}
	\label{workflow} 
 \vspace{-15pt}
\end{figure}

% \textbf{Design Goal.} \ourmodel aims to effectively conduct FL on a highly heterogeneous and dynamic environment consisting of mobile devices with limited memory capacity in order to jointly guarantee the model performance and the overall training efficiency. 
Figure \ref{workflow} depicts the architecture of \textit{FedHybrid}.
Following the standard schema of FL, \ourmodel employs a server/client architecture and consists of the following three core components:
1) Memory-aware Client Selector, 2) Heterogeneity-aware Graph Optimizer hosted by the central server and the 3) Local Training Engine deployed on each participating device. 
% 1) Memory-aware Client Selector deployed on the server side, 2) Heterogeneity-aware Graph Optimizer hosted by the central server, 3) Memory budget predictor and 4) Local Training Engine deployed on each participating mobile device. ,
These three core components coordinate with each other to direct the overall training procedure in order to strike a balance between memory reduction, model performance, and training efficiency.

% designs three novel components to mitigate the challenge of memory limitation: 1) Memory-aware Global Clien t Selector (Section \ref{memoryclientselection}), 2) Heterogeneity-aware Graph Optimizer (Section \ref{graphopt}) and 3) Memory-Aware Efficient Traning Engine (Section \ref{trainingengine}). The Memory-aware Global Client Selector dynamically chooses clients by balancing statistical and system efficiency considerations.
% Given the diverse runtime conditions of the selected clients (e.g., Free Memory, Computation Resources, and background services), the Heterogeneity-aware Graph Optimizer is devised to create an optimal execution plan for clients within the memory limitations, minimizing the extra time needed to address memory overhead. The Graph Optimizer includes multiple steps to analyze the model's computational graph and resolve the optimization challenge, as detailed in Section \ref{graphopt}.
% To ensure the theoretical efficiency of the optimal execution graph, we have developed a highly efficient training engine for on-device client training. This training engine orchestrates computational and memory-saving techniques to enhance runtime efficiency while preserving the model's accuracy.
% 这里最后首先讲讲每个component怎么联系在一起，因为东西太多逻辑不够好，每个component之间的关系以及输入输出是怎么样的，最好重新画那个图
\textbf{Workflow.} The overall workflow can be divided into the following main stages:
\circled{1} \textit{Initialization. } At the initialization stage, the central server first initializes the global model to be trained and oversees the status (e.g., hardware configuration) of the participating devices. 
\circled{2} \textit{Participant Selection}: the Memory-aware Client Selector selects the participating clients for the current training round through jointly evaluating their computing capability, memory budget and data utility in order to well balance the model accuracy and training efficiency. 
% by evaluating their computation ability, available memory, and data utilization.
\circled{3} \textit{Execution Graph Optimization}: 
For each selected client, Heterogeneity-aware Graph Optimizer judiciously analyzes the computation graph and configures the right memory saving strategy for each tensor in order to minimize the latency of the local training process given a specific memory budget. 
% To maximize the local training speed given a memory budget, 
% Heterogeneity-aware Graph Optimizer analyzes DNN's computational graphs and generates the optimal execution plan to perform the \emph{right} operation for each tensor.
\circled{4} \textit{Model\&Plan Transmission}: The server then broadcasts the model parameters together with the execution plan to the corresponding client. 
\circled{5} \textit{On-device Local Training}: After receiving the execution plan, the client employs the Memory-Aware Efficient Training Engine, crafted for executing detailed training plans. It optimizes runtime efficiency by employing a carefully orchestrated channel-wise mixed compression alongside activation recomputation. At the same time, it monitors the system status and predicts the memory budget for the next training round. 
% while minimizing the impact on the precision of training outcomes. .
% the client leverages the Memory-Aware Efficient Training Engine to train, which enhances runtime efficiency by utilizing a carefully orchestrated channel-wise mixed compression method alongside activation recomputation, aiming for the slightest impact on training precision. 
% To adapt the dynamic available memory, the operator profiler and runtime monitor record each operator's runtime info(e.g., memory de/allocation, page allocation time, and execution time) and the background app's workload status. 
% It also accommodates the dynamic available memory by logging runtime information (e.g., memory allocation/deallocation, page allocation time, and execution time) for each operator and monitoring the workload status of background applications by the operator profiler and runtime monitor.
\circled{6} \textit{Model aggregation \& Client status uploading}:  After completing local training, the client sends the updated model, memory budget, operators latency, and data utility back to the server. The central server then aggregates the local models and updates the global model.
% \circled{6} \textit{Model aggregation}: server aggregates updates from participants and updates the meta-info from the clients.
This process iterates until the model converges. 

% \vspace{-1em}
\subsection{Memory-aware Client Selection}\label{memoryclientselection}
Existing client selection schemes simply assume that all the devices have sufficient memory to well support the local learning process and fail to consider the impact on both model accuracy and training efficiency according to the discussion in Section 2.2. In this section, we first introduce the utility function and then discuss how \textit{FedHybrid} selects the clients through jointly taking into account the memory budget, computing capability and data utility.

\noindent{\textbf{Memory Utility.}} The memory utility of a device is defined as follows: 
\vspace{-3pt}
\begin{equation}
    \text{MemStat(i)}=  \left(\frac{M_i}{M_{G}} \right)^{\mathbf{1}(M_G>M_{i})}
\end{equation}
where $M_i$ is the memory budget of client $i$, representing the available memory for the training process as specified by the application or os and determined by the client's memory budget predictor (see Section \ref{sectionmempreditor}). This predictor estimates the client's memory budget over time, through taking into account the resource contention caused by the concurrently running applications in a dynamic training environment. $M_G$ is the memory requirement to train the model in sufficient memory conditions. $\mathbf{1}(x)$ is an indicator function that takes value 1 if x is true and 0 otherwise. This metric evaluates whether a device has sufficient memory (\(M_{i} \geq M_{G}\)) to train the model without any system overhead, yielding a memory utility of 1, indicating full capability. In contrast, a memory utility less than 1 (\(M_{i} < M_{G}\)) represents that a device lacks sufficient memory, highlighting potential limitations in its contribution to the training process due to memory constraints. 
The lower the value is, the higher the probability the system will trigger page reclaim during the local training process and thus lead to high training latency.
% This design is based on observation 2 in Section\ref{memorywallexp} which the intensive memory utilization can trigger the system to reclaim the page and then the training process reacces the evicted page from the slower storage. Thus sufficient memory resources can guarantee the swapping can not incur the extra runtime overhead for local training. 

\noindent{\textbf{Data Utility.}} To tackle the data heterogeneity of the clients, we measure the statistical utility of client $i$ by:
\begin{equation}
    \text{Stat(i)}= \sqrt{  \frac{1}{|B_i|} \sum_{k \in B_i} \text{Loss}(k)^2 }
\end{equation}
where the $|B_i|$ is the local training batch size of clients and the $\text{Loss}(k)$ indicates the training loss of data sample. A higher statistical utility indicates greater data importance due to the inclusion of challenging or underrepresented samples, which are essential for improving model robustness and generalization \cite{lai2021oort,kairouz2021advances,shin2022fedbalancer,li2022pyramidfl}.

\noindent{\textbf{Computing Utility.}} The computing utility of client i is defined as: 
\begin{equation}
    % \text{CompStat(i)}= \frac{1}{\mathrm{C}_i^{comp}}
    \text{CompStat(i)}= \frac{1}{\sum_{o \in \mathrm{CM}_i}^M t_o}
\end{equation}
where $\mathrm{CM}_i$ represents the set of unique operators, $M$ in total , employed within the training model, and $t_o$ is the average kernel computation time of operator $o$ on device $i$ collected from the local training engine.  A higher \(\text{CompStat(i)}\) suggests client \(i\) has a strong computation capability, leading to shorter computation times for NN operators. Conversely, a lower \(\text{CompStat(i)}\) indicates a weaker computational capability, resulting in longer computation times for executing the same NN operations. 
% discuss the DVFS, 
% where $\mathrm{CM}_i$ is the set containing the $M$ size of unique operator of the training model 

We unify the above utility models and generate the comprehensive utility function of client $i$ as follows: 
% \begin{equation}\label{newselection}
% \text{Util}(i) = \text{Stat(i)} \times\text{MemStat(i)}\times\text{CompStat(i)}
% \end{equation}
\begin{equation}\label{newselection}
\text{Util}(i) = \sqrt{  \frac{1}{|B_i|} \sum_{k \in B_i} \text{Loss}(k)^2 } \times \left( \frac{M_i}{M_{G}} \right)^{\mathbf{1}(M_i>M_{G})} \times \left( \frac{1}{\sum_{o \in \mathrm{CM}_i}^M t_o}\right)
\end{equation}

Through this fine-grained design of utility, even if a client lacks sufficient memory for local training, our memory-aware selection criterion might still choose that client if its statistical utility outweighs its system utility. This approach involves a larger, more diverse set of clients, significantly enhancing data diversity during training. Our selection strategy is grounded in theoretical insights: sampling high-loss samples reduces variance similarly to gradient norm-based sampling, resulting in faster convergence compared to naive random sampling \cite{cho2020client}.
To maintain system efficiency and fairness while enhancing the data diversity of the participant pool, we design an exploration-exploitation strategy based on the Multi-Armed Bandit with Upper Confidence Bound (MAB-UCB) model \cite{chen2013combinatorial} for client selection.
In this model, each device is treated as an ``arm'' of the bandit, and the set of selected arms is referred to as the super arm \( N \). We first calculate the utilities of all devices based on Equation \ref{newselection} and sort them from high to low. Then, we select \(\epsilon\) fraction of total selection clients from the high-utility participants. Additionally, we sample an \(1-\epsilon\) fraction of participants to explore potential participants that have not been selected before. These unexplored participants lack data utility, but we can prioritize those with rich system resources, as clients regularly check in and send memory and compute information as a "heartbeat" in existing FL systems \cite{lai2021oort,paulik2021federated,bonawitz2019towards}. This check-in only occurs when the phone is charging and connected to an unmetered network such as WiFi.  
% \textcolor{blue}{Throughout this process, \ourmodel adheres to FL principles by ensuring data privacy, only collecting necessary client states (memory budget and operator execution time) and data utility, and is compatible with common FL privacy enhancements such as differential privacy \cite{dwork2008differential}, secure aggregation \cite{feng2023does}, and homomorphic encryption \cite{zhang2020batchcrypt}.}
% \subsection{Discussion}
% \noindent\textbf{Privacy concerns.} \ourmodel adheres to FL principles by ensuring that raw training data never leaves the devices, thereby preserving data privacy. \ourmodel only collects clients' states (e.g., predicted memory budget and operator execution time) and data utility. We acknowledge that collecting clients' system information may raise privacy concerns. To address this, we can seamlessly integrate \ourmodel with common FL privacy enhancements, including differential privacy \cite{dwork2008differential}, secure aggregation \cite{feng2023does}, and homomorphic encryption \cite{zhang2020batchcrypt}.
All information is encrypted \cite{dwork2008differential,zhang2020batchcrypt} to ensure user privacy and anonymized. 

\vspace{-10pt}
\begin{algorithm}[!t]
    \caption{Plan Generation Mechanism.}
    \label{algo:plan_generation}
    % \SetKwFunction{Prepare}{prepare}
    \SetKwFunction{Recover}{recover}
    \SetKwFunction{Compute}{compute}
    \SetKwFunction{Alloc}{alloc}
    \SetKwFunction{MaxMPS}{MaxMPS}
    \SetKwFunction{compress}{compress}
    \SetKwFunction{Evict}{evict}
    
    \SetKwProg{Fn}{Function}{:}{}
    \KwIn{$origin\_comp\_seqs$, $memory\_budget$, $memory\_pool$}
    \KwOut{optimized plan}
    
    \Fn{\Compute{$comp\_op$}}{
        \For{$t$ \textbf{in} $comp\_op.input\_tensors$}{
            \If{$t.state$ == $COMPRESSED$}{
                % $t.lock()$\;
                $t.address \gets alloc(t.size)$\;
                % $result\_seqs.append(compress(t))$\;
                % $t.unlock()$\;
            }
            \ElseIf{$t.status$ == $EVICTED$}{
                $\Recover(t)$\;
            }
        }
        \For{$t$ \textbf{in} $comp\_op.output\_tensors$}{
            \If{$t$ is not allocated}{
                $t.address \gets alloc(t.size)$\;
            }
            % $t.lock()$\;
        }
    }

    % \Fn{\Recover{$tensor$}}{
    %     $op \gets tensor.from\_op$\;
    %     $\Prepare(op);\Compute(op)$\;
    % }

    \Fn{\Alloc{$comp\_op$}}{
        \While{$\text{pool\_size} + size \geq \text{memory\_budget}$}{
            $tensor \leftarrow \MaxMPS()$\;
            \If{$tensor.MPS_{compute} > tensor.MPS_{compress}$}{
                $\Evict(tensor)$\;
            }
            \Else{
                $\compress(tensor)$\;
            }
        }
    }

    \SetKwProg{Main}{Main}{:}{}
    \Main{}{
        \For{$op_i$ \textbf{in} $origin\_comp\_seqs$}{
            % $\Prepare(op_i)$\;
            $\Compute(op_i)$\;
        }
    }

\end{algorithm}

\subsection{Heterogeneity-aware Graph Optimizer}\label{graphopt}
% According to the discussion in Section 4.2, mobile devices that cannot provide sufficient memory can also be selected if they have high data utility. Thus, 
For a selected client, the Heterogeneity-aware Graph Optimizer is designed to conduct efficient training on devices with a given memory budget. 
According to the discussion in Section \ref{section:limitation_existing_memory}, directly applying the existing memory-saving techniques either deteriorates the model accuracy or slows down the overall training process. 
% For the selected clients, a training strategy that can effectively achieve memory reduction in order to meet the memory constraints without compromising the training efficiency is critical. 
% As outlined in our experimental analysis (Section \ref{section:limitation_existing_memory}), though individually applying existing memory reduction techniques can achieve memory saving, they often significantly extend the overall training process. 
% they often lead to a decline in model performance or a decrease in training efficiency.
Such observations lead us to a pivotal inquiry: \textit{Is there an approach that can effectively reduce the memory footprint without compromising the model accuracy and training efficiency for FL? }
% hybrid approach that leverages the specific tensor characteristics of different operators to enhance efficiency?
% According to experimental analysis in Section \ref{section:limitation_existing_memory}, although existing memory reduction techniques can achieve memory saving, they either degrade the model performance or deteriorate the training efficiency due to challenges posed by the unique architecture of mobile devices. This raises a key question that 有没有一种hybrid 可能根据不同operator的tensor特性来提高，因此我们来探究这个问题

% Thus, a memory-efficient training framework specifically designed for a mobile-based FL system is urgently required.  

% In this section, we aim to generate an optimal execution plan for all participant clients with the whole operator graph as input. We firstly explore solution of the corresponding optimization problem at the client level. Then we extend this approach to heterogenous clients by server-client coordinating.
% 这里要引入两种insight 来引入这个optimizer，首先我们要说明的是为什么现在单个策略不行（Mobile的策略来引入比较好），IO不稳定，其次我们要从Mobile的计算图出发，说明目前的在Mobile计算图上的特点，主要说明和Server的区别，然后提出新的策略
% \subsubsection{Balance training efficiency and memory reduction.} 
\begin{figure}[!t] 
	\centering
        % \vspace{-10pt}
	\includegraphics[width=0.99\linewidth]{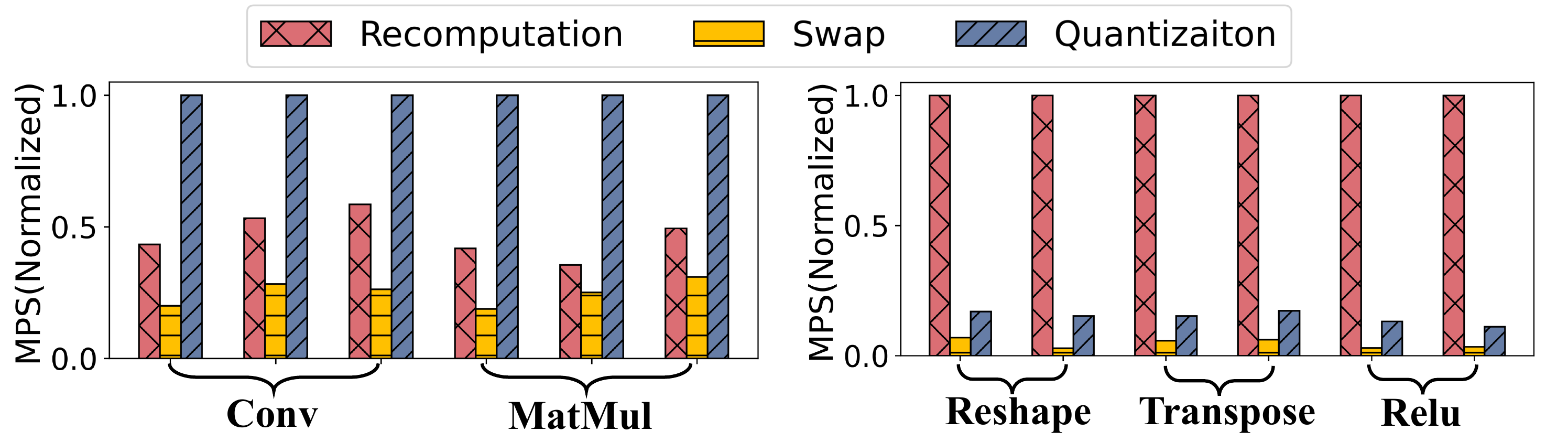}
   \vspace{-10pt}
	\caption{Comparison of MPS with different tensors from MobileNetV2.}
    \vspace{-15pt}
	\label{fig:op_compare} 
\end{figure}

\subsubsection{Opportunity.} 
To further investigate this issue, we propose a new metric, MPS (Memory reduced Per Second) to quantify the reward (RW) of the application of memory-saving techniques—namely recomputation, compression, and swapping-on targeted tensor $d$ which is defined as follows:
\begin{equation}\label{eq:metric}
% \begin{split}
 \text{RW}(d)= SM(d) \times FLT(d) , \ \text{MPS}(d) = \frac{\text{RW}(d)}{PC(d) + RC(d)} 
% \end{split}
\end{equation}
where \(SM\) (Saved Memory) represents the amount of memory conserved by using a particular technique, while \(FLT\) (Freed Lifetime) indicates the duration for which the tensor remains unused in memory, thus contributing to overall memory efficiency. The effectiveness of MPS lies in its ability to balance memory savings with computational overhead. \(PC\) (Purge Cost) and \(RC\) (Regain Cost) are critical in this metric, as they account for the time required to purge tensors and regenerate them when needed. By incorporating these factors, MPS provides a comprehensive tensor measurement that reflects not just the memory conserved but also the efficiency of the process in terms of computational time. 
% where \(SM\) refers to the saved Memory, the amount of memory saved with a certain memory reduction technique, \(FLT\) is Freed Lifetime, the lifetime of the tensor stored in memory until it is freed, \(PC\) is Purge Cost, the time cost required to eliminate data or tensors not currently needed, \(RC\) is Regain Cost, the time cost required to recover or regenerate the eliminated data or tensors for further operations. 
% Unlike the regeneration cost defined in server methods \cite{peng2020capuchin}, they ignore this cost since some regeneration techniques (such as swapping between GPU and CPU) can be asynchronous, thereby eliminating the cost. However, these techniques cannot be fully asynchronous due to limitations in the existing DNN framework and mobile hardware, necessitating the consideration of these additional costs. This metric is precise for understanding the trade-offs between memory efficiency and computational overhead in mobile training environments. 
A high MPS value signifies efficient memory savings with minimal purge or regeneration time costs, indicating a well-balanced trade-off between memory conservation and computational efficiency in mobile devices. Conversely, a low MPS value points to insufficient memory savings or high time costs, suggesting a need for improved strategies or computational optimizations tailored for mobile hardware constraints.
Figure \ref{fig:op_compare} represents the MPS of different memory-saving techniques on various tensors. Specifically, we select the 20 tensors in MobileNetV2 with the highest reward as case study, grouping them by their operator type and using different shapes to reduce duplication, resulting in 12 unique tensors. These high-reward tensors are selected because they have larger sizes and longer freed lifetimes, meaning discarding these tensors can free up more space during training, thus improving overall memory efficiency.

The following key findings direct us to the design of a graph optimizer. 
1) Low MPS for swapping. 
% we found that swap operations demonstrate consistently lower MPS values than both recomputation and compression across all tensors. This highlights the swap technique's limitations in efficiently balancing memory savings with computational demands.
The swapping technique consistently demonstrates the lowest MPS values across all tensors compared to recomputation and compression. This is for the reason that, although the embedded Universal Flash Storage (eUFS) 3.1 offers throughput of 2,100 MB/s read and 1,200 MB/s write, the main problem with swapping in a mobile environment is the unstealable I/O contention from background apps, which is due to the restriction of a single command queue and the lack of internal concurrency.
% This limitation highlights the swap technique's inefficiency in balancing memory savings with computational demands.
2) Various MPS in Recomputation and Compression. The utility of recomputation and compression highly depends on the specific tensor, with compression showing superior performance for complex operations like convolution and matrix multiplication. This superiority is attributed to the inefficiencies of mobile convolution operators, suggesting that compression is more adept at managing computational demands. Thus, \textit{it is essential to select the optimal technique (either recomputation or compression) for specific tensors within the computation graph.}

\begin{figure}[!t]
    \centering
    \begin{minipage}[!t]{0.45\linewidth}
        \centering
        \includegraphics[width=0.95\linewidth]{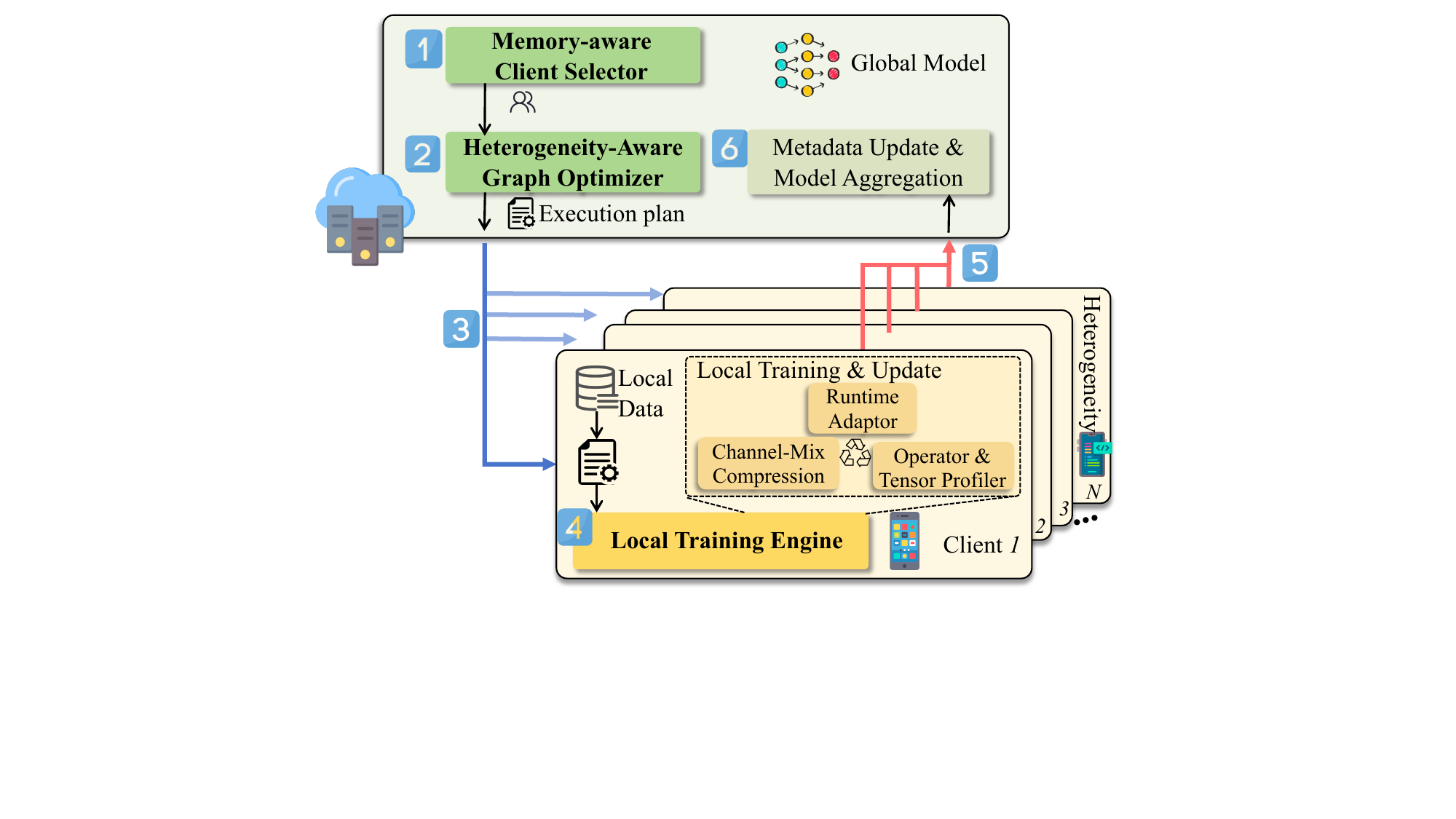} % 替换为你的图片文件
        % \caption{示例图片}
         \vspace{-10pt}
        \caption{Examples of layout transformation in DNNs.}
        \label{layout_example}
         \vspace{-12pt}
    \end{minipage}
    \hfill
    \begin{minipage}[!t]{0.45\linewidth}
        \centering
        \vspace{-10pt}
        % \caption{Android devices for the testbed experiments.}
        \captionof{table}{Latency and transformation breakdown across various models. `Layout Tran.' refers to the latency associated with explicitly transforming the tensor’s layout. `Other  Comp' indicates the latency attributed to the remaining operators. FT refers to finetine.}
        \vspace{-10pt}
        \LARGE
        \renewcommand{\arraystretch}{1.1} % Adjusts row spacing
        \resizebox{\linewidth}{!}{
        \begin{tabular}{|c|cc|}
\hline
% \toprule[2.5pt] % Top thick line
\multirow{2}{*}{Model} & \multicolumn{2}{c|}{Lat. breakdown (\%)}      \\ \cline{2-3} 
                       & \multicolumn{1}{c|}{Layout Tran.} & Other Comp. \\ \hline
ResNet34               & \multicolumn{1}{c|}{19.7}        & 81.3       \\ \hline
MobileNetV2            & \multicolumn{1}{c|}{14.2}        & 85.8       \\ \hline
ResNet50               & \multicolumn{1}{c|}{22.8}        & 77.2       \\ \hline
Bert                   & \multicolumn{1}{c|}{57.8}        & 42.2       \\ \hline
Swin (FT*)             & \multicolumn{1}{c|}{69.6}        & 30.4       \\ \hline
% ViT (FT*)              & \multicolumn{1}{c|}{54.5}        & 45.5       \\ \hline\bottomrule[2.5pt] % Bottom thick line
\end{tabular}}
        
        \label{layout_op_latency}
        \vspace{-10pt}
    \end{minipage}
\end{figure}

\textbf{Impact of Layout Transformation.} In addition to the regeneration efficiency issues related to operator types, we found that data layout transformations can introduce significant overhead during tensor recomputation in mobile training. 
Figure \ref{layout_example} illustrates a computation graph where the model explicitly includes Reshape and Transpose operations between the Conv and MutMal layers, and implicitly adds Gather and Reshape operations between the MutMal and MaxPool layers. These transformations ensure compatibility between different operators within a DNN's computation graph, as certain operators, especially those optimized for high execution performance on CPUs or GPUs, require specific data memory layouts such as NCHW or NC4HW4.
Layout transformations consume substantial memory bandwidth and disrupt the locality of subsequent operations. While individual transformations are relatively fast, their cumulative effect leads to significant performance overhead. The tensors generated by these transformations are not stored and have very short lifetimes. Therefore, each time a recomputed operator requires a layout transformation, it must be recalculated, further adding to the recomputation overhead. As shown in Table \ref{layout_op_latency}, which details the latency and transformation breakdown across various models using recomputation in MNN, a significant portion of time is spent on layout transformations, particularly in Transformer-based models. 
Previous recomputation strategies \cite{wang2022melon, peng2020capuchin} primarily attribute the cost of tensor recomputation to the computational effort of the tensor's operator alone. However, they overlook the additional costs associated with layout transformations. This oversight can lead to considerable performance overheads in various DNNs. To the best of our knowledge, no existing strategies account for the influence of layout transformations during tensor recomputation. This gap can result in suboptimal graph optimization policies and inefficiencies in the recomputation process. Thus, \textit{it is crucial to consider the layout transformation cost when using recomputation to balance memory reduction and training efficiency.}

Guided by the above design principles, the Heterogeneity-aware graph optimizer aims to minimize the overall execution cost while ensuring that peak memory usage remains below the specified budget by incorporating a \textbf{hybrid} tensor-saving strategy. This approach is detailed in Algorithm \ref{algo:plan_generation}.
When determining which tensor needs to be discarded, we exploit the MPS (Equation \ref{eq:metric}) to estimate the regeneration cost. To accurately measure the recomputation cost for tensor \( d \) generated by operator \( j \), we propose new metric \( RC_{compute}(d) \) for computing \( MPS_{compute} \):
% \vspace{-3pt}
\begin{equation}
     RC_{compute}(d)=  OpTime(j) + \psi \sum_{i=0}^{Dep(d)} OpTime(i)
\end{equation}
% \vspace{-2pt}
where \( OpTime(j) \) is the time cost of recomputing operator \( j \), \( \psi = 2 \) when including CPU/GPU transformation, otherwise \( \psi = 1 \). \( Dep(d) \) is the Depth-first search (DFS) depth from the output of operator \( j \) to the final operator \( z \) that takes tensor \( d \) as input, including any intermediate layout transformation operators.
When generating the computation graph, we attribute a property to tensors requiring layout transformations that record the original operator index. This is achieved through layout check functions in MNN, which inspect the tensor's layout. In addition, we compare the original computation graph with the framework-generated graph to identify these layout transformation operators. The final transformed tensors retain the original operator's index, enabling the DFS to trace back to the original tensor's operator. This ensures all layout transformation operators are included in the recomputation cost calculation.

The graph optimizer starts by taking the entire computation graph as input (lines 18-19). When memory is exhausted, it initiates a memory reclamation procedure (lines 11-16). During this process, the optimizer evaluates each tensor's Memory Pressure Score (MPS) to decide whether to evict or compress the tensor with the highest MPS (line 12). Each tensor is assigned a recomputation MPS ($MPS_{compute}$) and a compression MPS ($MPS_{compress}$). If $MPS_{compute}$ exceeds $MPS_{compress}$, the tensor is evicted; otherwise, it is compressed (lines 13-16). This process repeats until memory usage is within the budget (lines 1-8). When a compressed tensor required by the current operator is needed, it is decompressed, and memory is allocated accordingly (lines 3-4). If a tensor is missing from memory, the optimizer recomputes it from its original tensors, potentially recomputing source tensors if they are also unavailable (lines 5-6). This mechanism ensures that logical dependencies are maintained during operations while adhering to memory constraints.

\textbf{Plan generation for large scale of clients.} As discussed in Section \ref{section:limitation_existing_memory}, generating an execution plan frequently is time-consuming for mobile devices, even when using heuristic-based algorithms. To address this issue, \ourmodel offloads this procedure to a central server, thereby reducing computational burden for the participating devices. In FL, generating an execution plan for each participant is challenging and time-consuming due to varying available memory and computing capabilities. To efficiently manage large-scale clients, we capitalize on the heterogeneous nature of the client pool. We employ clustering techniques, such as k-means, based on clients' memory and computing characteristics. Within each cluster, we select the client with minimum capability to generate an execution plan, ensuring feasibility for all cluster members. This approach is optimal as the resulting plan is universally applicable within the cluster without compromising performance on more capable devices. To further enhance efficiency, we implement a cache table for rapid plan generation and retrieval. This table stores previously generated plans based on different budgets and client device information, enabling quick adaptation to various client scenarios. This strategy ensures efficient use of resources while maintaining the quality of the FL process across a wide range of client capabilities.
\subsection{Local Training Engine}\label{trainingengine}
% \vspace{-5pt}
% \subsection{Channel-wise Mix Compression}\label{channelcompress}
The Local Training Engine is designed to conduct the local training according to the received execution plan generated by the Heterogeneity-aware Graph Optimizer. The Local Training Engine then sends the memory budget, as predicted by the memory budget predictor, along with the operator's cost time, which is the average time collected from the last batch of all training rounds. All information is encrypted to ensure user privacy and anonymized.
However, efficiently performing the execution plan faces the following critical challenges: 
1) According to the execution plan, compression or recomputation is selected to apply on a specific tensor in order to achieve memory reduction while guaranteeing training efficiency. However, we find that existing compression techniques can introduce compression errors at the same time as they overlook the outlier value in tensor, making quantization ineffective. \textit{Thus, how to conduct effective compression without minimum accuracy loss is the first critical challenge.} 
2) The resource contention caused by concurrently running apps can dynamically change the available system memory. \textit{Thus, how to precisely determine the safe memory budget is another critical challenge.} 
In order to conduct efficient local training, we propose a new compression technique called Channel-wise Mix Compression and memory budget predictor.  
% The Local Training Engine is designed to direct the local training on each client according to the received execution plan. In specific, it is designed for the following two purpose:  1) reducing the compression error in tensor without compromising the efficiency in compression ratio and speed. 2) adaptive to switch the plan to deal with dynamic memory budget at runtime. Accordingly, we propose the following techniques.
% the dynamic memory in runtime and improve the training efficiency while saving memory in runtime, the Memory-Aware Efficient Training Engine has the two following objective: 1) 
% deal with the compression error in tensor
% is then used in the selective execution module.

% In this section, we first investigate why the existing activation compression methods are not effective in FL. Then based on the analysis, we propose our new channel-wise mix compression for FL. 
\begin{figure}[!t] 
	\centering
  % \vspace{-15pt}
	\includegraphics[width=1\linewidth]{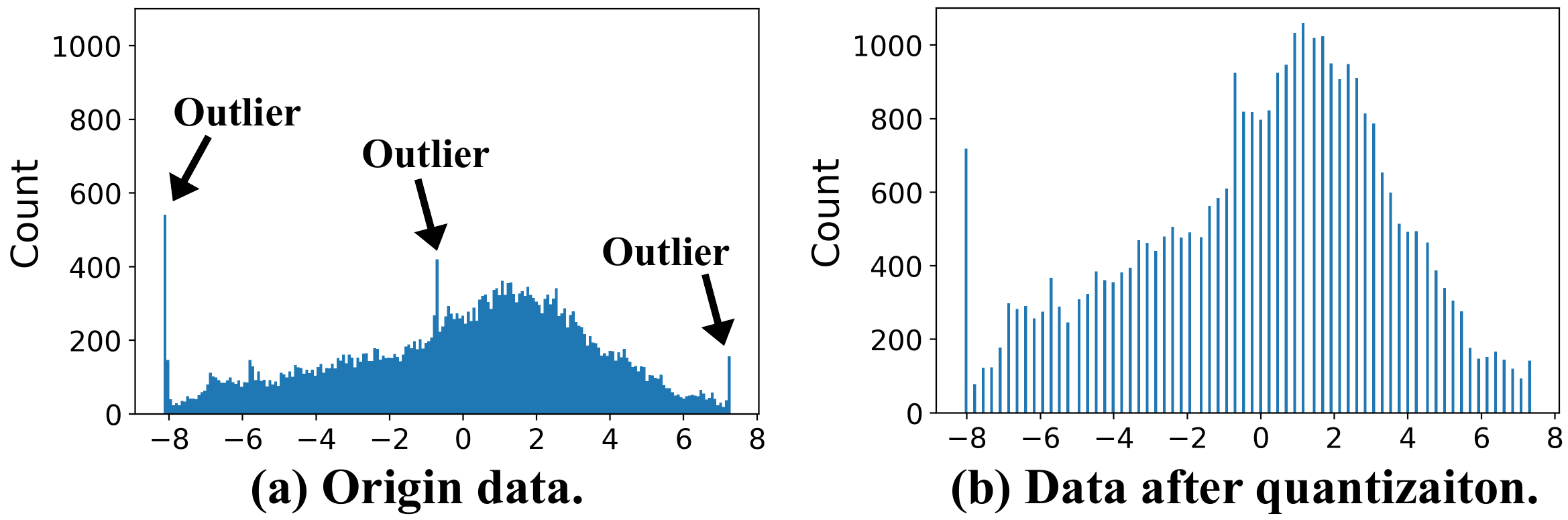}
   \vspace{-20pt}
	\caption{The data distribution of before and after quantization with outlier.}
     \vspace{-18pt}
	\label{outliershow} 
\end{figure}
% % and how they effect the model
\subsubsection{Channel-wise Mix Compression}\label{channelcompress}
Current activation compression methods \cite{liu2022gact,evans2021ac,chen2021actnn} use lossy compression to reduce memory via quantization, but they often neglect the impact of tensor outliers, leading to significant errors. Outliers, play a crucial role in DNNs, with their values in CNNs and Transformers \cite{guo2023olive} exceeding standard values by over $81 \times$ and $347 \times$. 
% as shown in Figure \ref{outlierdistribute}
This discrepancy causes substantial deviation in post-quantization data from the original, as seen in Figure \ref{outliershow}, resulting in a loss of information.
% This discrepancy causes substantial deviation in post-quantization data from the original, as seen in Figure \ref{outliershow}, resulting in a loss of information.
This problem is exacerbated in FL, where data compression across multiple clients amplifies these errors. Additionally, the significance of outliers in determining layer outputs and gradient updates \cite{heo2023rethinking, lin2023awq} highlights the necessity for improved compression techniques that consider outlier preservation to maintain data integrity and model accuracy.

\begin{figure}[!t] 
	% \centering
 % \vspace{-14pt}
	\includegraphics[width=1\linewidth]{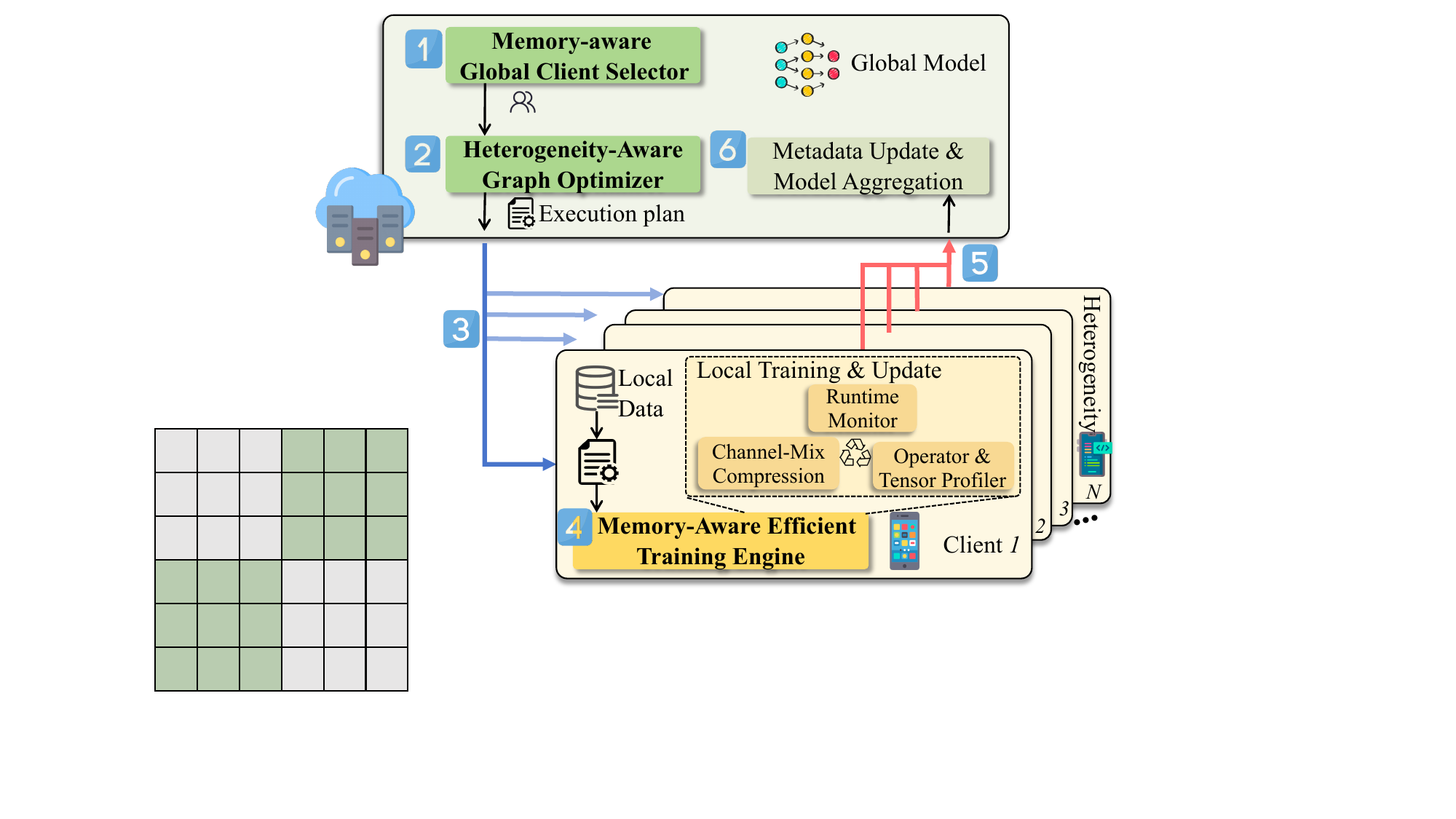}
        \vspace{-20pt}
	\caption{Workflow of Channel-wise Mix Compression.}
	\label{workflowchannelcompress} 
 \vspace{-12pt}
\end{figure}

To balance accuracy and compression without compromising quantization effectiveness, we design adaptive channel-wise mix compression, as shown in Figure \ref{workflowchannelcompress}. This approach exploits the observation that activation values are mostly concentrated \cite{guo2023olive}, allowing precise quantization of normal values. Outliers are identified using the empirical 3\(\sigma\) rule \cite{6895997rule}, acknowledging their sparsity and channel-wise clustering rather than treating them individually. This insight guides us to classify channels into outlier-rich (salient) and normal, applying a mix-quantization strategy that significantly compresses data while retaining crucial model performance information. Normal channels, characterized by concentrated values, undergo quantization to achieve compression, ensuring an efficient balance between memory usage and data integrity.

For outlier-rich channels, which typically have many zero values, we cannot directly use existing lossy compression methods \cite{evans2020jpeg,chen2021actnn} since we observe that zero values among the channels can cause significant errors during compression. Specifically, these errors arise because zero values are not preserved accurately after decompression. This inaccuracy can result in high variance in the gradients, preventing the training process from converging. To address this problem, we design a nuanced block-wise compression method that preserves outliers while achieving high accuracy. We first categorize the non-zero values of a channel feature map (\(h \times w\)) into dense and sparse blocks by starting with a binary mask matrix to distinguish non-zero values. This step addresses the local sparsity of DNN activations. To improve operation speed, we use average pooling to process the feature map instead of element-wise counting. Mean filtering through average pooling (window size \(n \times n\)) helps identify sparse blocks by comparing them against a sparsity threshold \(\tau\). Sparse blocks are then efficiently compressed using the Compressed Sparse Row (CSR) format, balancing data integrity with compression effectiveness.

In dense blocks, we implement predictive compression, leveraging the local data relationship for more efficient compression within an error-bound, distinct from the conventional linear scale approach. This process encompasses three stages to effectively compress the data. Initially, for each element \(x\) in block \(B_n\), the Lorenzo predictor is employed to estimate its value \(\hat{x}\) using neighboring elements, formalized as:
\begin{equation}
    P_{n}=\{ p^* | p^*=f_l(x_k^{SR}), x_k\in B_{n}\},
\end{equation}
% \vspace{-5pt}
where \(P_n\) represents the predicted value for block \(B_n\), \(x_k^{SR}\) denotes the surrounding values of \(x_k\), and \(f_l\) is the Lorenzo predictor, which is defined by:
\begin{equation}
    \sum_{
\substack{
0 \leq k_1, \ldots, k_m \leq n \\
k_1, \ldots, k_d \neq 0
}}
\left(
\prod_{j=1}^m (-1)^{k_j+1}
\binom{n}{k_j}
\right)
\cdot x_{c_1-k_1, \ldots, c_d-k_d},
\end{equation}
ensuring that the sum of coefficients equals one. The predictive model ranges from simple 1-dimensional first-order predictions to more complex multidimensional constructs.
Subsequently, we assess the prediction error \(\varphi_{n}\) between the original and predicted values, confined within the error bound \(\epsilon\), to determine:
\begin{equation}
    \varphi_{n}=\{\delta|\delta= p^*-x_k,\delta<\epsilon,p^*\in P_{n}, x_k\in B_{n}\},
\end{equation}
where \(\delta\) is the prediction error. Values within \(\epsilon\) are quantized to conserve memory, while outliers are preserved intact, ensuring compression accuracy as predictions utilize decompressed rather than original values. To enhance the compression ratio, especially for outliers requiring additional bits, a Huffman tree is constructed from the quantization array for encoding, coupled with lossless compression for the Lorenzo predictor's coefficients, facilitating decompression.

Decompression involves Huffman decoding, followed by the reconstruction of floating-point values from quantization codes and their prediction. This predictive compression not only optimizes memory use but also captures and leverages data relationships beyond simple scaling, offering a nuanced approach to data compression in dense NN blocks.

\begin{figure}[!t] 
	\centering
        % \vspace{-12pt}
	% \includegraphics[width=1\linewidth]{Figures/memory_predictor.pdf}
        \includegraphics[width=0.99\linewidth]{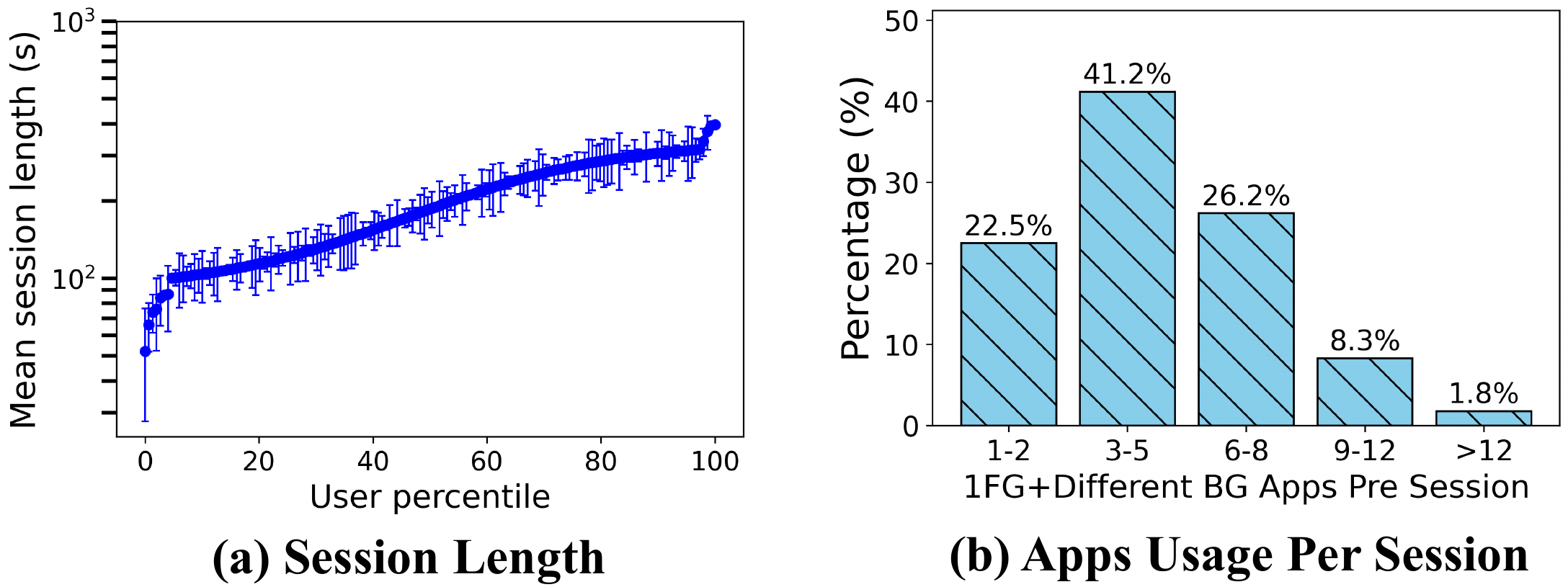}
 % \vspace{-1.2em}
    \vspace{-10pt}
	\caption{Analysis of Users' Long-Term App Usage in the Carat dataset \cite{oliner2013carat}. (a) The mean and standard deviation of the duration of individual interaction sessions across all users, presented on a logarithmic scale. (b) The number of applications accessed during each interaction session.}
 % , highlighting that all users engage with one foreground application while varying the number of background applications available for switching.}}  To ensure a comprehensive trace analysis, we selected data from 150 users, each providing at least 30 days of usage information.
	\label{memoryusageanalysis} 
    \vspace{-5pt} 
\end{figure}

\begin{figure}[!t] 
	\centering
        % \vspace{-12pt}
	% \includegraphics[width=1\linewidth]{Figures/memory_predictor.pdf}
        \includegraphics[width=0.99\linewidth]{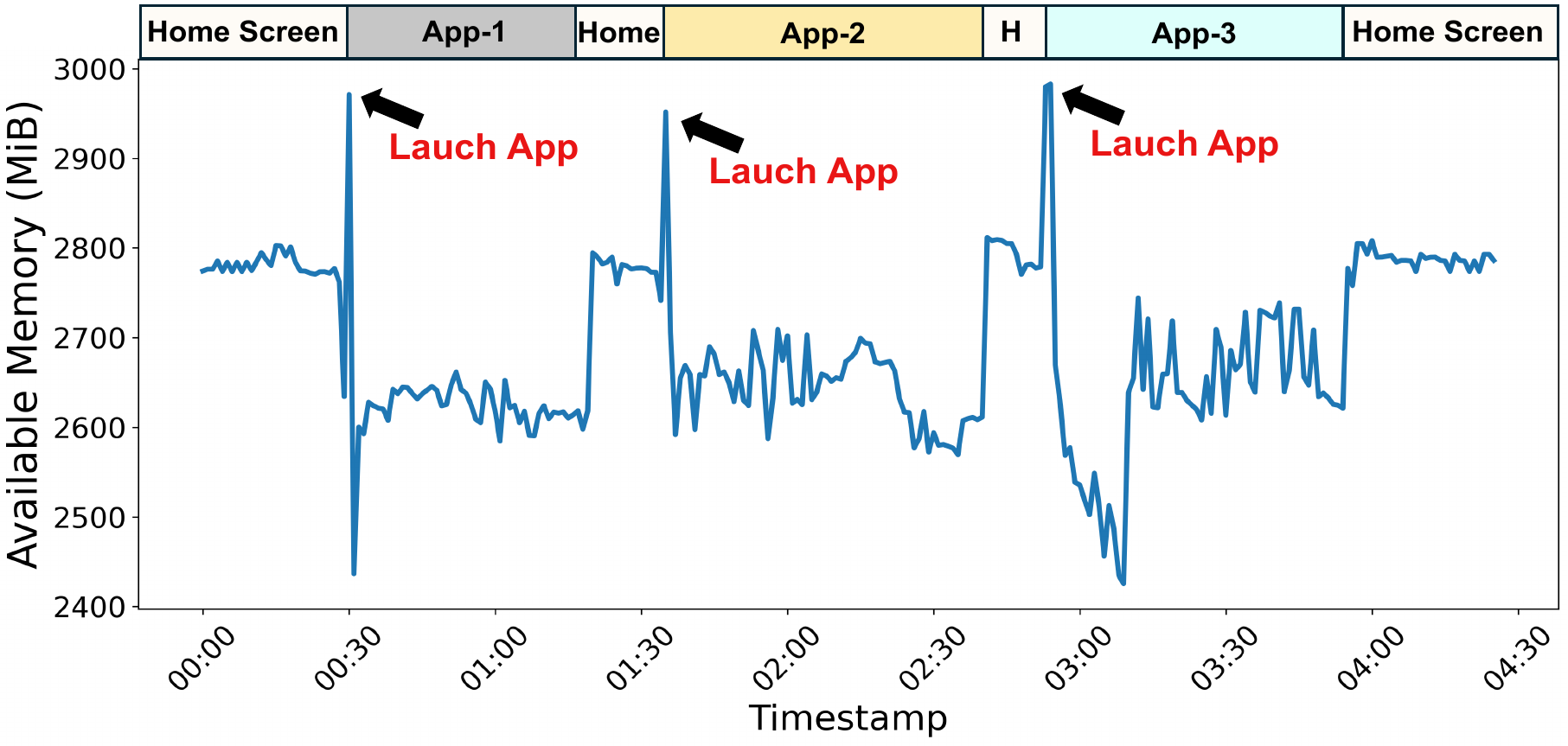}
 % \vspace{-1.2em}
    \vspace{-12pt}
	\caption{Analysis of Memory Usage Patterns. We select a sample user usage trace from the Carat dataset \cite{oliner2013carat} and emulate it on the S22 mobile device using the Monkey testing tool \cite{Monkey}. Caching background app re-launch, with the top bar representing the trace process.}
 % (b) Pattern 2: Launching a new app while three background applications remain active.}}
	\label{memoryapp_usagetrace} 
    \vspace{-10pt} 
\end{figure}

\subsubsection{Memory Budget Predictor}\label{sectionmempreditor}
Memory budget of the training process can be impacted by the resource contention caused by the concurrently running apps during user interaction. User interaction with mobile apps often follows discernible patterns over time \cite{tu2018your,li2021understanding}.
% While mobile device memory usage can appear random due to diverse user interactions, we hypothesize that user behavior often follows discernible patterns over time \cite{tu2018your,li2021understanding}. 
To make the investigation, we analyze long-term app usage data from 150 users over 30 days using the Carat dataset \cite{oliner2013carat,li2021understanding}. Our analysis reveals consistent patterns in both session length and app usage. As illustrated in Figure \ref{memoryusageanalysis}(a), over 70\% of users have an average session duration of 153 seconds, with 73\% falling between 126 to 412 seconds, suggesting relatively fixed and brief interaction times. Furthermore, Figure \ref{memoryusageanalysis}(b) demonstrates that users typically operate with one foreground app and 3-5 background apps during sessions. 
% This finding indicates that despite variations in the specific apps used by each user, the number of active apps during each session remains relatively constant.
These observations reveal a consistent memory usage pattern where a significant portion of device memory is typically occupied by a fixed number of active apps, corresponding to certain usage patterns. This consistency in app usage behavior provides a foundation for predicting memory budget in dynamic mobile environments.
% This indicates that despite variations in the specific apps used by each user, the number of apps active during each session remains relatively constant. It reveals a consistent memory usage pattern where a significant portion of device memory is typically occupied by a fixed number of active apps, corresponding to certain usage patterns. 
% This suggests that users tend to maintain a small but stable set of applications running in the background during their interactions with a device. 
% , indicating a consistent memory usage pattern where a significant portion of device memory is occupied by a fixed number of active apps.

To investigate how these patterns translate to actual memory usage, we emulated user behavior on the S22 using the Monkey tool \cite{Monkey}. Figure \ref{memoryapp_usagetrace} illustrates the resulting memory usage during app launches and transitions.
% It is essential to note that in Android, when a user returns an app to the background, it does not exit completely; instead, it remains active and cached in the background, allowing for quick reactivation without needing a full restart. Android's memory management system causes brief spikes when launching apps, followed by stabilization. The cached apps show a sharp spike and quick recovery as depicted in Figure \ref{memoryapp_usagetrace} (a). 
In Android's ecosystem, apps don't fully exit when moved to the background; instead, they remain cached for quick reactivation. This design leads to distinctive memory usage patterns. App launches cause brief memory spikes followed by stabilization, while switching between cached apps results in sharp spikes with quick recovery, as shown in Figure \ref{memoryapp_usagetrace} (a). These spikes indicate that memory demand is highly dynamic during app switching and launches, but once a foreground app is active, memory consumption stabilizes. This insight reveals that short-term memory spikes during app launches and switches significantly increase the variance of available memory. However, these spikes are transient and do not reflect the stable memory usage during active app use. Therefore, these fluctuations should be filtered out when determining real-time memory budgets. 

\begin{figure}[!t] 
	\centering
        % \vspace{-12pt}
	% \includegraphics[width=1\linewidth]{Figures/memory_predictor.pdf}
        \includegraphics[width=0.94\linewidth]{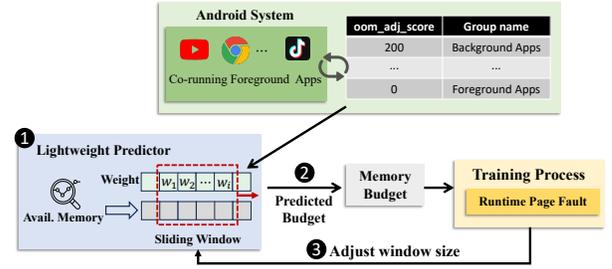}
 % \vspace{-1.2em}
    \vspace{-10pt}
	\caption{The workflow of Memory Budget Predictor.}
	\label{memoryPredictor} 
    \vspace{-12pt} 
\end{figure}

% To predict system memory availability accurately, we introduce an adjusted memory metric that accounts for the impact of various swap types on page reclamation timing. Swap type significantly affects when page reclamation occurs.
Based on the above-observed memory usage pattern, we propose the lightweight memory budget predictor specifically designed to account for the fluctuating workloads of the participating devices in FL. Figure \ref{memoryPredictor} shows the architecture and workflow of the memory budget predictor. It not only estimates available memory under dynamic usage conditions but also coordinates with the graph optimizer to determine the optimal moments for plan regeneration, balancing overall efficiency with regeneration overhead. To effectively predict the system's available memory, we first introduce an adjusted available memory metric $M_{safe}$, which considers the impact of different swap types on page reclamation. In practice, we observe that swap type significantly influences the timing of page reclamation.
For example, compression-based swaps like ZRAM \cite{zram} use a portion of DRAM as swap space, reducing available DRAM and triggering earlier memory reclamation.
To address this, we define adjusted available DRAM memory: 
$M_{safe} = M_{Avail} - \alpha \cdot \text{Watermark}_{high}$.
\( M_{Avail} \) is the current available DRAM memory from \textit{/proc/meminfo}, and \(\text{Watermark}_{high}\) is a threshold for stopping background page reclamation, typically a fixed value proportional to total DRAM pages. The parameter \(\alpha\) represents swap activity type: \(\alpha = 1\) for disk-based swap and \(\alpha = 2\) for ZRAM.

% Using the adjusted available DRAM memory $M_{safe}$ as an input with sampling period $T_{sample}$, we then apply a weighted moving average method to capture the most recent and important memory usage patterns. % \begin{equation}
%     w_i=\sum_{j=1}^{B} \frac{1000}{\text{oom\_adj\_socre}_j}
% \end{equation}
To mitigate the short-term fluctuations caused by app switching and launching, we employ a weighted moving average method to capture the most recent and important memory usage patterns. For each sampling period $T_{sample}$, we calculate the average adjusted available DRAM memory $M_{safe}$. The window size, $W$, is initiated as the duration of one training epoch, ensuring that memory predictions align with the training process. The sliding size is $T_{slide}$. The weights in the moving average are calculated based on the importance of the applications running during each window. We use the \textit{oom\_adj\_score} \cite{oomscore}, which prioritizes processes for termination in the Android system, with higher scores indicating lower importance. We calculate the weights $w_i$ at time $i$ by: $w_i=\sum_{j=1}^{B} \frac{\text{Max\_Score}}{\text{oom\_adj\_socre}_j}$.
$B$ is the total application's activities \footnote{We only calculate processes with oom\_adj\_score $\geq$ 0. The maximum score is 1,000. Foreground applications have an oom\_adj\_score of 0, which is set to 1 when calculating the weight.} collected by \textit{ActivityManager} \cite{ActivityManager}. Foreground applications and critical background applications (such as music playback, downloads, and synchronization tasks) are given higher weights, while less critical background applications receive normal weights. For example, when an app is running in the foreground, its \textit{oom\_adj\_score} is 0 (indicating highest priority), but when the user switches to the home screen or another app, the \textit{oom\_adj\_score}  for the previous app increases, reducing its importance in memory calculations. Additionally, we found that processes associated with the home launcher \cite{HOME_APP_ADJ} (e.g., the home screen) have significantly higher \textit{oom\_adj\_score} than foreground apps. As a result, the transition from the home screen to an app launch is assigned lower weights, making the impact of this transition less significant in the memory prediction. This weighting scheme inherently assigns lower weights to short-term fluctuations, as memory spikes are often caused by temporary background activities with higher \textit{oom\_adj\_score}.
This approach helps monitor short-term trends in DRAM memory usage and allows us to adjust memory budgets effectively. 
% in response to the training progress. 
The prediction process can be described as follows:
% \vspace{-5pt}
\begin{equation}
\text{M}_{Pred}^t = \frac{\sum_{i=t-W+1}^{t} w_i \cdot M_{safe}^{i}}{\sum_{i=t-W+1}^{t} w_i}
\end{equation}
where ${M}_{Pred}^t$ represents the predicted weighted moving average of the adjusted available DRAM memory $M_{safe}$ at time $t$, $M_{safe}^{i}$ represents the adjusted available DRAM memory at the time $i$, and $w_{i}$ are the importance weights of the task applications at time $i$.

\begin{table}[!t]
% \vspace{-12pt}
\caption{Android devices for the testbed experiments.}
\vspace{-10pt}
\label{phoneinfo}
\resizebox{\linewidth}{!}{
\begin{tabular}{c|lcccc}
\hline
                         & Year & Mem & SoC                & OS           \\ \hline
Samsung Galaxy S23 Ultra & 2023 & 8G  & Snapdragon 8 Gen 2 & Android 13          \\ \hline
Samsung Galaxy S22       & 2022 & 8G  & Snapdragon 8 Gen 1 & Android 12      \\ \hline
OnePlus 10 Pro           & 2022 & 12G & Snapdragon 8 Gen 1 & Android 12      \\ \hline
Google Pixel 6           & 2021 & 8G  & Google Tensor      & Android 12    \\ \hline
Xiaomi Redmi Note 10     & 2021 & 6G  & Snapdragon 678     & Android 11        \\ \hline
Honor V30 Pro            & 2019 & 8G  & Kirin 990          & HarmonyOS 3       \\ \hline
Realme GT Neo 2          & 2021 & 8G  & Snapdragon 870     & Android 11     \\ \hline
\end{tabular}
}
\vspace{-12pt}
\end{table}

\noindent\textbf{Plan Regeneration Rule.} To balance runtime efficiency with the dynamics of available memory, we set the following criteria for clients to regenerate the plan from the server:
1). The current round's training process total page faults count exceeds the previous round's average by more than $TP_1$.
2). The training process is terminated by the Low Memory Killer (LMK) more than $TP_2$ times.
If plan regeneration is needed, the predictor first adjusts the sliding window size with factor $WS_{adj}$as:
$W_{new} = W_{current} \times WS_{adj}$.
Then, a new budget is generated.

\section{Evaluation}
% 这部分要补充超参数
% We evaluate \ourmodel to answer the following research questions:
% \begin{itemize}
%     \item \textbf{Q1:} Can \ourmodel surpass the performance benchmarks set by current State-Of-The-Art (SOTA) methods and the conventional random client selection approach across a variety of FL scenarios?
%     \item \textbf{Q2:} Which intrinsic components of \ourmodel are pivotal for its success, and how do they individually contribute to its efficiency
%     \item \textbf{Q3:} To what extent does \ourmodel contribute to reducing the overall system costs in terms of resource utilization?

% \end{itemize}

\subsection{Implementation}
% , a hybrid tensor computation engine,
We implemented \ourmodel using a combination of C++ and Python. For the local training engine, which includes channel-wise mix compression and the memory budget predictor, we build on top of MNN (version 2.3.0) \cite{alibaba2020mnn}, adding 5.8K lines of code (LOC). For the memory-aware client selection and heterogeneity-aware graph optimizer, we extend FedScale \cite{lai2022fedscale} with an additional 1.5K LOC. Additionally, we implement three baselines (Capuchin, GACT, and HeteroFL) based on prior literature. Since some of these models were designed for servers rather than mobile devices, we adapte and re-implemented them on top of MNN for a fair comparison.

\begin{figure}[!t] 
	
 % \vspace{-18pt}
 \centering
   \includegraphics[width=0.97\linewidth]{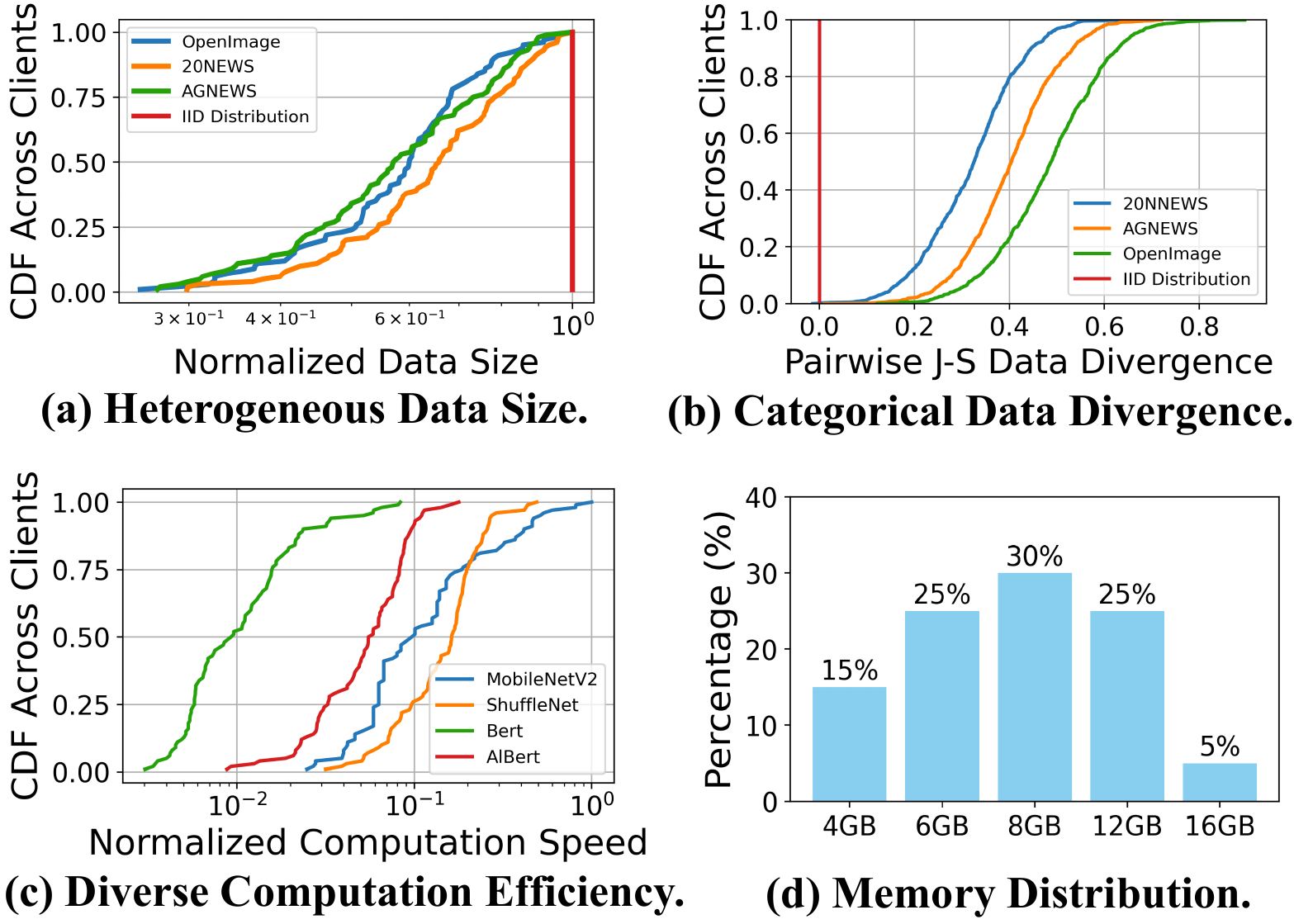}
 \vspace{-10pt}
	\caption{Experiment setup. (a) per-client quantity of samples, (b)  deviation of per-client categorical distributions using Jensen-Shannon divergence, (c) distribution of the computation efficiency across
clients, (d) distribution of memory distribution across clients. }
	\label{noniiddis} 
 \vspace{-12pt}
 \label{exp_setup}
\end{figure}

\begin{table*}[!t]
% \vspace{-13pt}
\centering
\caption{Summary of improvements on time to accuracy. We take the highest accuracy that random selection can achieve as the target, which is moderate due to the high task complexity and lightweight models. 
% All speed-up comparisons are benchmarked against random selection's time that achieves the target accuracy level.
}
\vspace{-1em}
\label{table1}
\LARGE
\renewcommand{\arraystretch}{1.4} % Adjusts row spacing
\resizebox{\linewidth}{!}{
% \usepackage{multirow}
% \begin{table}[]
\begin{tabular}{cccccccccccccccccc}
\toprule[2.5pt] % Top thick line
\multirow{2}{*}{Dataset} &
  \multirow{2}{*}{Model} &
  \multicolumn{2}{c}{Random+FedProx} &
  \multicolumn{2}{c}{Oort+FedProx} &
  \multicolumn{2}{c}{HeteroFL+FedProx} &
  \multicolumn{2}{c}{\textit{FedHybrid}+FedProx} &
  \multicolumn{2}{c}{Random+Yogi} &
  \multicolumn{2}{c}{Oort+Yogi} &
  \multicolumn{2}{c}{HeteroFL+Yogi} &
  \multicolumn{2}{c}{\textit{FedHybrid}+Yogi} \\ 
 &
   &
  Speedup &
  \multicolumn{1}{c}{Acc} &
  Speedup &
  \multicolumn{1}{c}{Acc} &
  Speedup &
  \multicolumn{1}{c}{Acc} &
  Speedup &
  Acc &
  Speedup &
  \multicolumn{1}{c}{Acc} &
  Speedup &
  \multicolumn{1}{c}{Acc} &
  Speedup &
  \multicolumn{1}{c}{Acc} &
  Speedup &
  Acc \\ \hline
\multirow{2}{*}{OpenImg} &
  MobileNetV2 &
  1$\times$ &
  51.56 &
  2.89$\times$ &
  52.75 &
  - &
  38.65 &
  \textbf{15.55$\times$} &
  \textbf{68.09} &
  1$\times$ &
  53.42 &
  2.02$\times$ &
  54.25 &
  - &
  38.76 &
  \textbf{11.34$\times$} &
  \textbf{70.12} \\
 &
  ShuffleNet &
  1$\times$ &
  51.03 &
  2.19$\times$ &
  52.09 &
  - &
  38.25 &
  \textbf{9.86$\times$} &
  \textbf{66.03} &
  1$\times$ &
  52.01 &
  1.86$\times$ &
  53.95 &
  - &
  37.35 &
  \textbf{8.43$\times$} &
  \textbf{68.83} \\ \hline
\multirow{2}{*}{20NEWS} &
  Bert &
  1$\times$ &
  31.71 &
  1.14$\times$ &
  36.72 &
  0.80$\times$ &
  31.92 &
  \textbf{9.05$\times$} &
  \textbf{70.42} &
  1$\times$ &
  33.52 &
  1.08$\times$ &
  35.22  &
  1.12$\times$ &
  33.89  &
  \textbf{8.57$\times$} &
  \textbf{72.68} \\
 &
  Albert &
  1$\times$ &
  26.53 &
  1.05 $\times$&
  31.22 &
  - &
  15.36 &
  \textbf{7.85$\times$} &
  \textbf{63.39} &
  1$\times$ &
  28.71 &
  1.03$\times$&
  32.56&
  - &
  20.21 &
  \textbf{7.03$\times$}&
  \textbf{65.12}  
   \\ \hline
\multirow{2}{*}{AGNEWS} &
  Bert &
  1$\times$ &
  66.47  &
   1.21$\times$&
  67.21 &
  - &
  54.75 &
  \textbf{3.70$\times$} &
  \textbf{86.97} &
  1$\times$ &
  67.89 &
  1.22$\times$ &
  69.21 &
  - &
  53.34 &
  \textbf{4.22$\times$} &
  \textbf{88.34} \\
 &
  Albert &
   1$\times$ &
   65.63&
   1.14 $\times$&
   66.84&
   - &
   50.23&
   \textbf{4.52$\times$}&
   \textbf{82.69}&
   1$\times$ &
   66.74&
   1.08$\times$ &
   67.96&
   -&
   51.25&
   \textbf{4.13$\times$} &
   \textbf{84.24} 
    \\
\bottomrule[2.5pt] % Bottom thick line
\end{tabular}
% \end{table}
}
\vspace{-12pt}
\end{table*}

% \usepackage{multirow}
\begin{table}[!t]
\caption{Summary of improvements. Take the highest accuracy that random+prox can achieve as the target.  
% All speed-up comparisons are measured against the time taken by \ourmodel+prox to reach the target accuracy level.
}
\label{tbl:improvements}
\vspace{-1em}
\LARGE
\renewcommand{\arraystretch}{1.5} % Adjusts row spacing
\resizebox{\linewidth}{!}{
\begin{tabular}{cccccccccc}
\toprule[2.5pt] % Top thick line
\multirow{2}{*}{Dataset} &
  \multirow{2}{*}{Model} &
  \multicolumn{2}{c}{\textbf{\textit{FedHybrid}+FedProx}} &
  \multicolumn{2}{c}{Oort+Melon} &
  \multicolumn{2}{c}{Oort+Capuchin} &
  \multicolumn{2}{c}{Oort+GACT} \\ 
 &
   &
  \multicolumn{1}{c}{\textbf{Speedup}} &
  \textbf{Acc.} &
  \multicolumn{1}{c}{Speedup} &
  Acc. &
  \multicolumn{1}{c}{Speedup} &
  Acc. &
  \multicolumn{1}{c}{Speedup} &
  Acc \\ \hline
\multirow{2}{*}{OpenImg} & MobileNetV2 & \textbf{15.55$\times$} & \textbf{68.09} & 0.83$\times$ & 57.62 & 0.53$\times$ & 57.40                      & 0.62$\times$ & 51.84 \\
                         & ShuffleNet  & \textbf{9.86$\times$}  & \textbf{66.03} & 0.94$\times$ & 54.59 & 0.57$\times$ & 53.95                      & 0.61$\times$ & 51.63 \\ \hline
\multirow{2}{*}{20NEWS}  & Bert        & \textbf{9.05$\times$}  & \textbf{70.42} & 0.68$\times$ & 44.53 & 0.48$\times$ & \multicolumn{1}{l}{42.74} & 0.51$\times$    & 32.68 \\
                         & Albert      &  \textbf{7.85$\times$}   & \textbf{63.39}  &  0.64$\times$  &42.72  & 0.52$\times$    & 40.45         & 0.56$\times$             & 30.62      \\ \hline
\multirow{2}{*}{AGNEWS}  & Bert        & \textbf{3.70$\times$}  & \textbf{86.97} & 0.77$\times$ & 76.55 & 0.67$\times$ & 76.23                      & 0.59$\times$             & 67.04 \\
                         & Albert      & \textbf{4.52$\times$}  & \textbf{82.69} & 0.86$\times$ & 72.68 & 0.76$\times$ &    71.89                     & 0.58$\times$             &66.39      \\
\bottomrule[2.5pt] % Bottom thick line
\end{tabular}
}
\vspace{-15pt}
\end{table}

% \vspace{-5pt}
\subsection{Experiment Setup}\label{exp_detail}
% \textbf{Infrastructure.} To evaluate the effectiveness of \ourmodel, we construct a testbed designed to accurately replicate both data and system heterogeneity as encountered in real-world scenarios. \textit{To emulate the system heterogeneity}, we establish an FL system consisting of 8 Android smartphones with different hardware specs, detailed in Table \ref{phoneinfo}. Meanwhile, we leverage AI Benchmark \cite{AIBenchmark} performance data for diverse deep learning model runtimes across various devices to represent real-world device runtime variances. To closely replicate runtime device usage, we enhance AI Benchmark's data by incorporating training time measurements from smartphones with active background applications, refining our emulation of system performance. The computation distribution across the clients is shown in Figure ~\ref{exp_setup}(c). \textit{To emulate the memory limitation}, devices are set up with a standardized ZRAM \cite{zram} configuration, using 1 GB of memory for swap space with lz4 compression \cite{zram}. Additionally, a 3 GB swap partition is allocated on NAND storage. The device's memory capacity distribution is shown in Figure ~\ref{exp_setup}(d). 
% \vspace{-5pt}
\textbf{Setup.} 
% To evaluate the effectiveness of \textit{FedHybrid}, we construct a testbed designed to replicate both data and system heterogeneity encountered in real-world scenarios. 
To accurately simulate memory impact on training across diverse devices and emulate system heterogeneity, we combine simulation with physical testing. While we use AI-Benchmark \cite{AIBenchmark} to model device heterogeneity, similar to \cite{lai2021oort,li2022pyramidfl,shin2022fedbalancer}, we enhance accuracy by conducting real on-device training using MNN on a testbed of devices with 7 hardware configurations (Table \ref{phoneinfo}). We then extrapolate these results to a wider range of SoCs by scaling measured training times based on AI-Benchmark's theoretical computation time ratios, applying this only to devices with matching memory configurations. This hybrid approach provides a more realistic assessment of training performance across varied mobile devices. The computation distribution across clients is shown in Figure \ref{exp_setup}(c).
To simulate realistic dynamic memory conditions, we vary background applications based on real-world distributions \cite{liang2020acclaim} and simulate foreground app usage using the Carat dataset \cite{oliner2013carat}. The devices are configured with standardized ZRAM \cite{zram}, using 1 GB of memory for swap space with lz4 compression \cite{zram}. Additionally, a 3 GB swap partition is allocated on NAND storage. The memory capacity distribution is shown in Figure \ref{exp_setup}(d).
% This approach accurately represents both OS memory overhead and user interaction patterns, capturing impacts on training that theoretical estimates alone cannot reflect. 
% \textit{To emulate system heterogeneity}, we establish an FL system with 30 Android devices with 7 different hardware configurations (detailed in Table \ref{phoneinfo}). The computation distribution across clients is shown in Figure \ref{exp_setup}(c).
% We use AI Benchmark \cite{AIBenchmark} performance data to represent deep learning model runtimes across different devices. 
% Additionally, we incorporate training time measurements from smartphones with active background applications to refine our emulation of system performance. 
% \textit{To emulate memory limitation}, devices are configured with standardized ZRAM \cite{zram}, using 1 GB of memory for swap space with lz4 compression \cite{zram}. Additionally, a 3 GB swap partition is allocated on NAND storage. The memory capacity distribution is shown in Figure \ref{exp_setup}(d).
\textit{To emulate the data heterogeneity}, we categorize training datasets into different levels of non-identical distribution (Non-IID) with Dirichlet distribution \cite{kairouz2021advances} $\alpha=0.1$, and employ Jensen-Shannon divergence \cite{li2022pyramidfl,kairouz2021advances} to measuring sample counts and categorical distribution variations, as illustrated in Figure ~\ref{exp_setup}(a)(b). We also build a simulator based on the Server/Client architecture utilizing a GPU server equipped with $8 \times$ NVIDIA H800 GPUs as the established FL practices \cite{lai2021oort,li2021hermes,li2022pyramidfl,shin2022fedbalancer}.

% In line with established practices in FL research \cite{lai2021oort,li2021hermes,li2022pyramidfl,shin2022fedbalancer}, our experimental evaluation adopts an emulation approach, utilizing a GPU server equipped with 8x NVIDIA H800 GPUs.

% For fair comparisons, \ourmodel and baseline frameworks ran solely on the devices' high-performance cores.

\noindent \textbf{Datasets and Models.}
% Several well-known DNN and Transformer models from the computer vision (CV) and natural language processing (NLP) domains are employed for evaluation. For CV tasks, we train OpenImg \cite{kuznetsova2020open} dataset, which includes 1.5 million images across 600 categories, on MobileNetV2 \cite{sandler2018mobilenetv2} and ShuffleNet \cite{zhang2018shufflenet}, which are pre-optimized for resource-limited mobile devices. 200 of 6,582 clients are selected each round for this task. For NLP tasks, we utilize the AGNEWS \cite{zhang2015character} and 20NEWS \cite{defferrard2016convolutional} datasets, comprising over 127K and 18K items, respectively, for classification. 
% % The number of training samples for each class is 30K and testing 1.9K.  
% We train them on the Albert\cite{lan2019albert} and Bert\cite{devlin2018bert} for the text classification. For this task, 20 of 100 clients are selected when using 20NEWS and 200 of 2040 clients for AGNEWS each round.
We evaluate several well-known DNN and Transformer models in computer vision (CV) and natural language processing (NLP). For CV tasks, we train the OpenImg dataset \cite{kuznetsova2020open}, which includes 1.5 million images across 600 categories, on MobileNetV2 \cite{sandler2018mobilenetv2} and ShuffleNet \cite{zhang2018shufflenet}, optimized for resource-limited mobile devices. Each round selects 200 out of 6,582 clients. For NLP tasks, we utilize the AGNEWS \cite{zhang2015character} and 20NEWS \cite{defferrard2016convolutional} datasets, with over 127K and 18K items respectively, training on Albert \cite{lan2019albert} and Bert \cite{devlin2018bert} for text classification. The number of client selection varies by dataset: for 20NEWS, 20 out of 100 clients are chosen each round, while for AGNEWS, 200 out of 2,040 clients are selected.
% Each round, 20 of 100 clients are selected for 20NEWS, and 200 of 2040 clients for AGNEWS.

\noindent \textbf{Hyper-parameters.}
We set $\epsilon =0.9$,  \( TP_1 = 2 \) , \( TP_2 = 3 \) and $WS_{adj}=0.9$. For CV tasks, the batch size is 64, with 500 rounds and a local training epoch of 10 at a learning rate of 0.045. For NLP tasks, the batch size is 8, with 400 rounds and a local training iteration of 1 at a learning rate of 0.1. The max sequence len gth is 128 for both 20NEWS and AGNEWS.

\noindent\textbf{Baselines.} We compare \ourmodel with the following baselines. Addressing FL heterogeneity: (1) Random \cite{mcmahan2017federated} selects clients randomly each round. (2) Oort \cite{lai2021oort} optimizes device selection by combining statistical utility with device training time. (3) HeteroFL \cite{diao2020heterofl} adjusts convolutional layer channels to fit diverse memory capacities. Enhancing on-device memory efficiency: (4) Melon \cite{wang2022melon} uses micro-batching and recomputation to minimize memory usage. (5) Capuchin \cite{peng2020capuchin} employs swap and recomputation strategies for memory reduction. (6) GACT \cite{liu2022gact} dynamically compresses activations, adjusting compression ratios per network layer. 
\begin{figure}[!t] 
	\centering
	\includegraphics[width=0.75\linewidth]{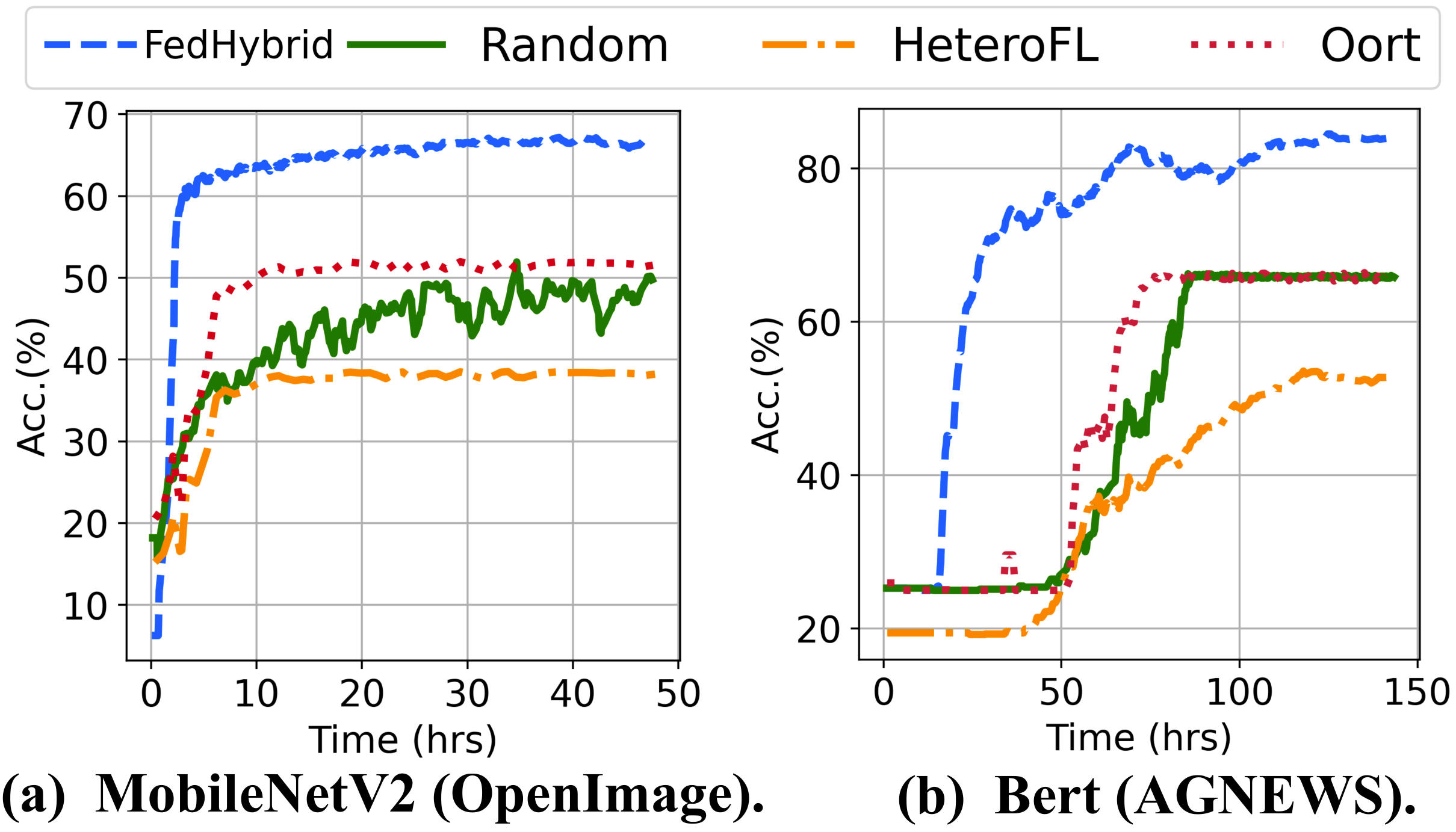}
 \vspace{-1em}
	\caption{Time-to-accuracy comparison with FL baselines.}
	\label{ttaccbl1} 
 \vspace{-10pt}
\end{figure}

\begin{figure}[!t] 
	\centering
	\includegraphics[width=0.75\linewidth]{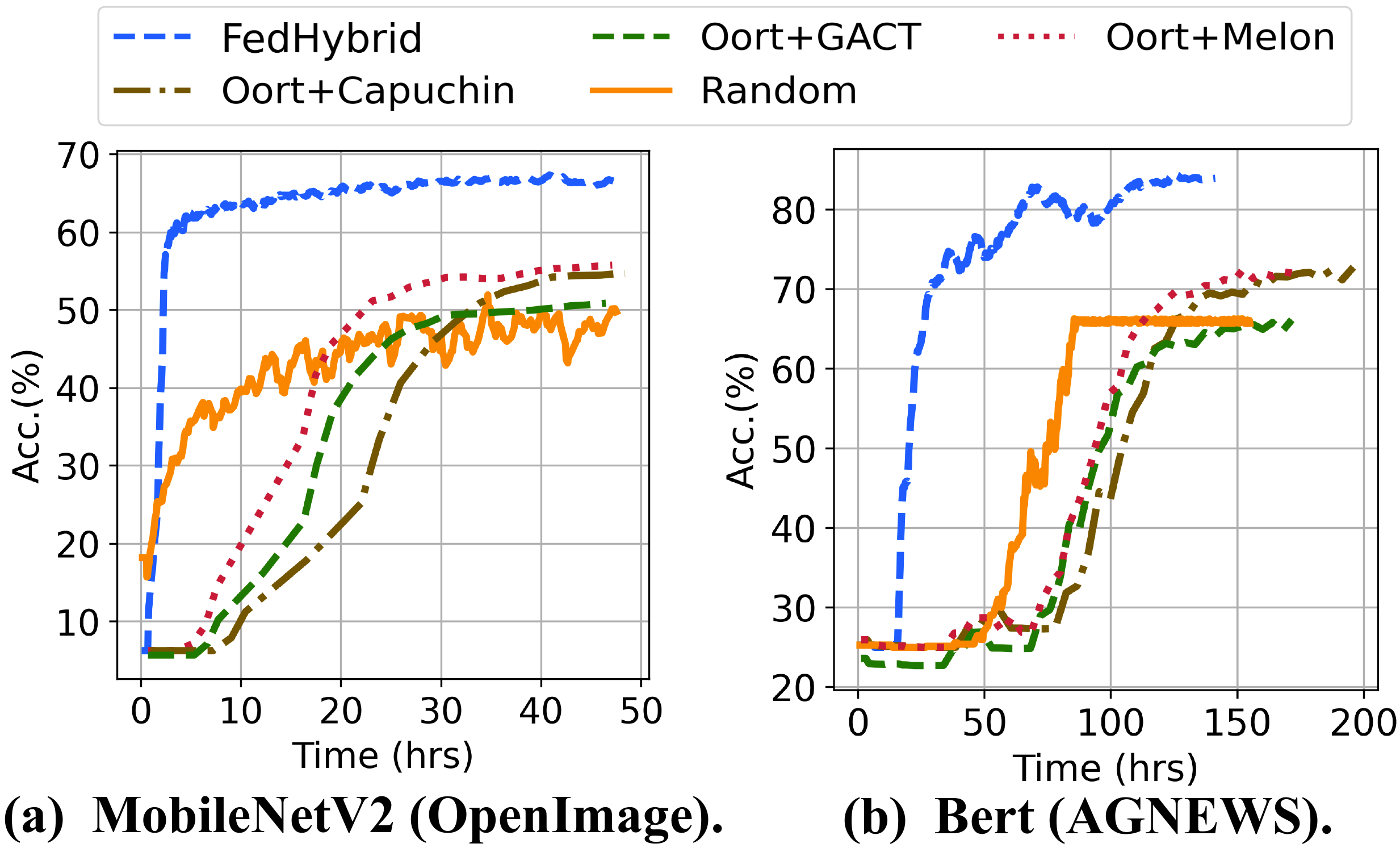}
 \vspace{-1em}
	\caption{Time-to accuracy-comparison of FL integrated memory saving techniques.}
	\label{ttaccbl2} 
 \vspace{-17pt}
\end{figure}

\subsection{End-to-end Performance}
% \vspace{-5pt}
\noindent\textbf{\ourmodel significantly outperforms existing FL baselines across all domains and tasks.} Table \ref{table1} presents the end-to-end performance of \ourmodel in comparison to several FL baselines. Figure \ref{ttaccbl1} reports the accuracy achieved by \ourmodel and FL baselines across different models. We observe that:
(a) Superior Accuracy: \textit{FedHybrid}+FedProx achieves 15.3\% and 33.7\% higher accuracy than Oort+FedProx for training MobileNetV2 and Bert, respectively. This improvement is primarily due to \textit{FedHybrid}'s ability to enable participation from more memory-constrained clients, which other methods fail to incorporate effectively. Both Oort and Random selection methods exclude memory-constrained clients, leading to reduced sample diversity. HeteroFL, while attempting to accommodate memory limitations, excessively prunes submodel parameters to fit constrained devices, resulting in performance degradation. This underscores HeteroFL's reliance on a certain number of full-model capable clients for optimal performance. (b) Faster Convergence: When training MobileNetV2, \ourmodel achieves the target accuracy 5.38× and 15.55× faster than Oort+FedProx and Random+FedProx, respectively. This acceleration is driven by two main factors: \ourmodel optimizes the computation graph based on the system workload of each client, and the local engine minimizes the overhead associated with memory optimization operations. Oort and Random selection methods are slowed down by memory-constrained clients in each round, extending overall training time. The speedup is even more pronounced when training Bert, as its large model size necessitates significant client data, which hinders the overall training progress in alternative methods. 

\noindent\textbf{\ourmodel Significantly Improves Time-to-Accuracy Compared to SOTA FL Frameworks with Memory Saving Optimization.} Figure \ref{ttaccbl2} reports the accuracy achieved by \ourmodel and FL baselines combined with memory saving methods across different models. As shown in Table \ref{tbl:improvements}, we compare \textit{FedHybrid}'s performance against Random and Oort (each integrated with Melon, Capuchin, and GACT). All methods surpass Random, highlighting the "memory wall" that limits participant involvement and impairs model performance. However, \ourmodel outperforms Oort integrated with Melon, Capuchin, and GACT in terms of accuracy. Specifically, when training Bert on the 20NEWS dataset, \ourmodel achieves 25.89\%, 27.68\%, and 37.74\% higher accuracy, respectively. This is due to the simplistic application of memory reduction strategies in FL training, which overlooks client heterogeneity. Additionally,  Oort's selection bias towards clients with shorter training completion times, while sidelining those with less memory but valuable data, further differentiates \textit{FedHybrid}. Therefore, Oort's time-prioritized selection may suboptimally choose clients, as it might exclude those whose completion times are prolonged due to memory limitations. These factors, combined with the extra computational demands of these strategies, hinder their convergence speed. Consequently, they require 13.3$\times$, 18.8$\times$, and 17.7$\times$ longer training times for Bert on the 20NEWS dataset compared to \textit{FedHybrid}.

\begin{figure}[!t] 
% \vspace{-15pt}
	\centering
	\includegraphics[width=0.95\linewidth]{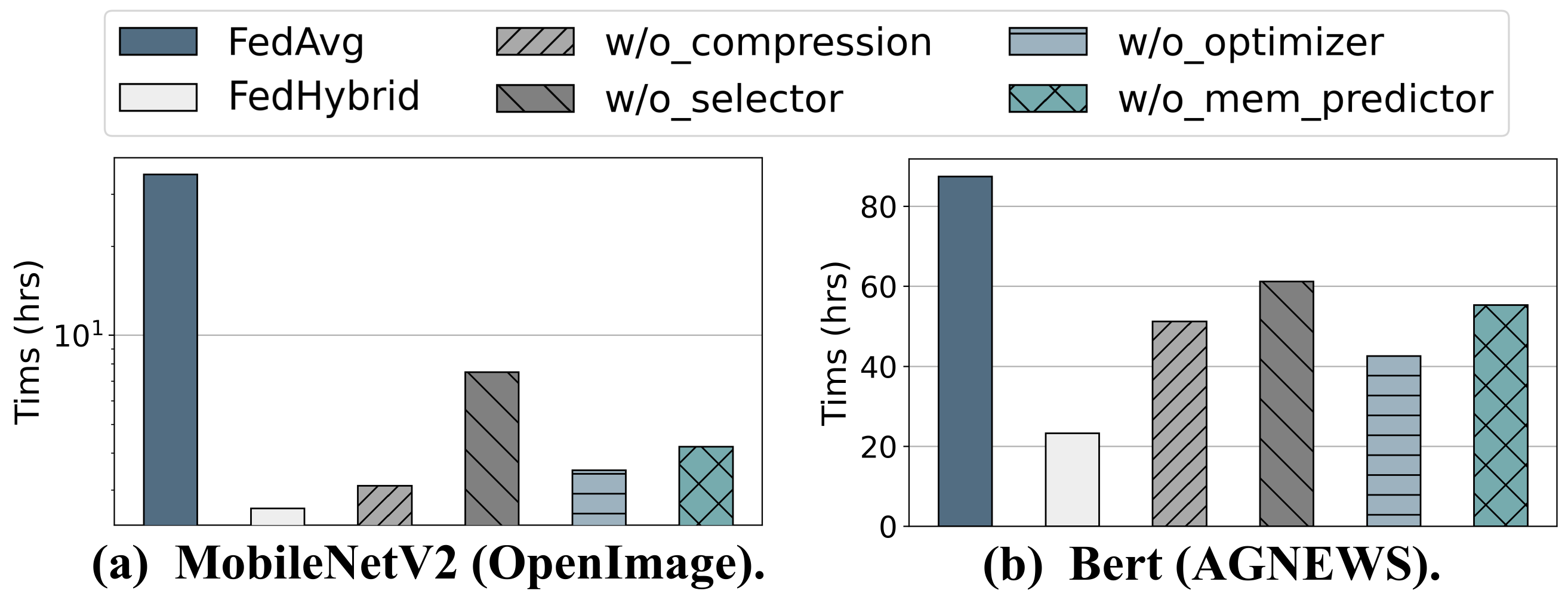}
 \vspace{-10pt}
	\caption{Significance of Key Designs of \textit{FedHybrid}.
 %Time-to-accuracy comparison for ablation study. 
 % w/o compression: deactivate channel-wise mix compression; w/o gbpot: disable global optimizer; w/o selector: without memory-aware client selector.
 }
        \label{abltation1} 
        \vspace{-17pt}
         % \vspace{-1.5em}
\end{figure}

% \vspace{-12pt}
\subsection{Effectiveness of Each Component}
% \vspace{-6pt}
% We implement two breakdown versions of \ourmodel to evaluate and understand the effectiveness of each of the key components.
% \noindent\textbf{Impact of memory-aware client selector.} We replace the client selector with the random selector, which randomly chooses clients for training in each round.  Figure \ref{abltation1} shows the deteriorations on final accuracy and convergence speed. For example, in Figure \ref{abltation1}(a), \ourmodel w/o selector shows slower to \ourmodel, with convergence speeds 3$\times$ slower. The fact demonstrates that the more abundant client data helps boost the overall performance. This suggests that considering the statistical utility of clients is crucial for achieving optimal accuracy. 
\noindent\textbf{Impact of Memory-aware Client Selector.} Replacing the memory-aware client selector with a random selector significantly slows convergence speed, as shown in Figure \ref{abltation1}. In both tasks depicted, \ourmodel without the selector converges 3$\times$ and 2.6$\times$ slower than \textit{FedHybrid}. This demonstrates that leveraging more abundant client data enhances overall performance, indicating that considering both the statistical and system utility of clients is crucial for achieving optimal accuracy.
% \ourmodel w/o selector does not improve the model accuracy as quickly as \ourmodel and Oort. The convergence speed is also 9$\times$ and 4.3$\times$ slower compared to both. This results from \ourmodel w/o selecting clients with slow training speed possibly. Meanwhile, we observer that final accuracy is 12.3\% higher than Oort but 4.5\% lower than \ourmodel. The fact demonstrates that the more abundant client data helps boost the overall performance, but the statistical utility of clients still needs to be comprehensively considered to achieve optimal accuracy.

\begin{figure}[!t]
    \centering
    % \begin{minipage}{0.45\linewidth}
    %     \includegraphics[width=1\textwidth]{Figures/mobilenet_32_oneplus_middle_core.pdf}
    %      \vspace{-1em}
    % \end{minipage}
    % \begin{minipage}{0.45\linewidth}
    %     \includegraphics[width=1\textwidth]{Figures/bert_8_oneplus_middle_core.pdf}
    %     \vspace{-1em}
    % \end{minipage}   
    % \vspace{-1pt}
    % \includegraphics[width=\linewidth]{Figures/Execution_time_vsbudget.pdf}
    \includegraphics[width=0.99\linewidth]{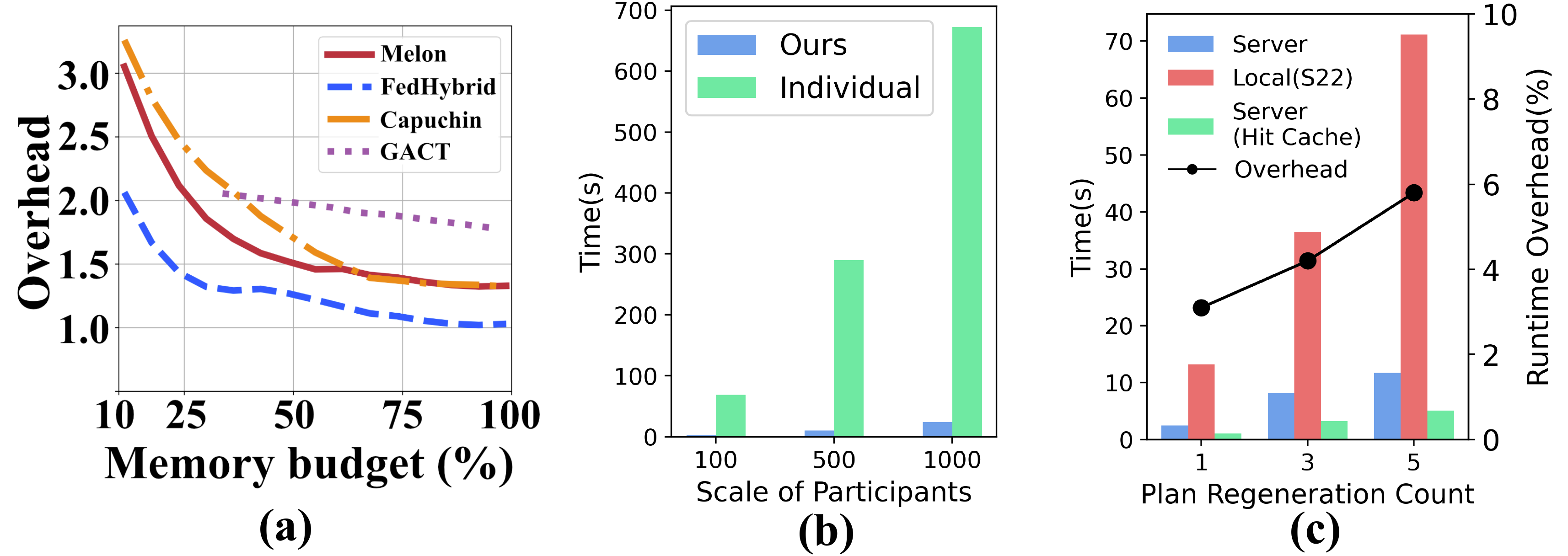}
     \vspace{-7pt}
    % \caption{Execution time under various budget ratio for MobileNetV2 (Left) and Bert (Right).}
    \caption{Overhead Analysis for \textit{FedHybrid}. (a) Execution time under various budget ratio for MobileNetV2 in S22. (b) Overhead of graph optimizer with different scale of clients. (c) The runtime cost of plan regeneration with various counts in S22 (connected 100Mbps WIFI).
    }
   \vspace{-17pt}
    \label{fig:overhead_with_budget}
\end{figure}

% \noindent \textbf{Impact of heterogeneity-aware graph optimizer.} In this part, a global optimizer is removed, requiring that each client generates an execution plan itself. Figure~\ref{abltation1} shows that without the optimizer, the convergence speed is 6.1$\times$ slower compared to our model. The results mostly stem from that execution plan generation is a time-consuming process, which impedes some clients from finishing their training tasks and uploading the model timely. However, only 1.7\% of accuracy degradation is noticed, which is mostly because our selector consistently chooses clients with large statistical utility.
\noindent \textbf{Impact of Heterogeneity-aware Graph Optimizer.} Removing the global optimizer and requiring each client to generate its own execution plan using Melon's plan generation results in a 1.3$\times$ and 1.8$\times$ slower convergence speed in two tasks, as shown in Figure~\ref{abltation1}. This slowdown is primarily due to the time-consuming nature of execution plan generation (as shown in Figure \ref{fig:overhead_with_budget}(c), $9\times$ slower than server's generation), which hinders some clients from completing their training tasks and uploading the model on time. Moreover, existing plan generation methods often overlook the cost of layout transformation, resulting in less effective plans.

% \noindent \textbf{Impact of channel-wise mix compression}: We remove the benefits by prohibiting compression of the optimizer. Figure~\ref{abltation1} illustrated the convergence speed is 1.1$\times$ and 2.2$\times$ slower in both task due to the significant recomputaiton overhead for each client, further exacerbating straggler problem and a 2.8\% decrease in accuracy. We also plot the overhead comparison of representative tensors and find that the time required for quantization and \ourmodel's compression is comparable, while both  are much lower than that of convolution. This underscores the superiority of improving training efficiency by adopting compression. Futhermore, Figure~\ref{noniiddis} depicts a decrease in gradient variance of restored tensor after decompression, which results in small accuracy degradation.

\begin{figure}[!t] 
 % \vspace{-20pt}
	\centering
\includegraphics[width=1\linewidth]{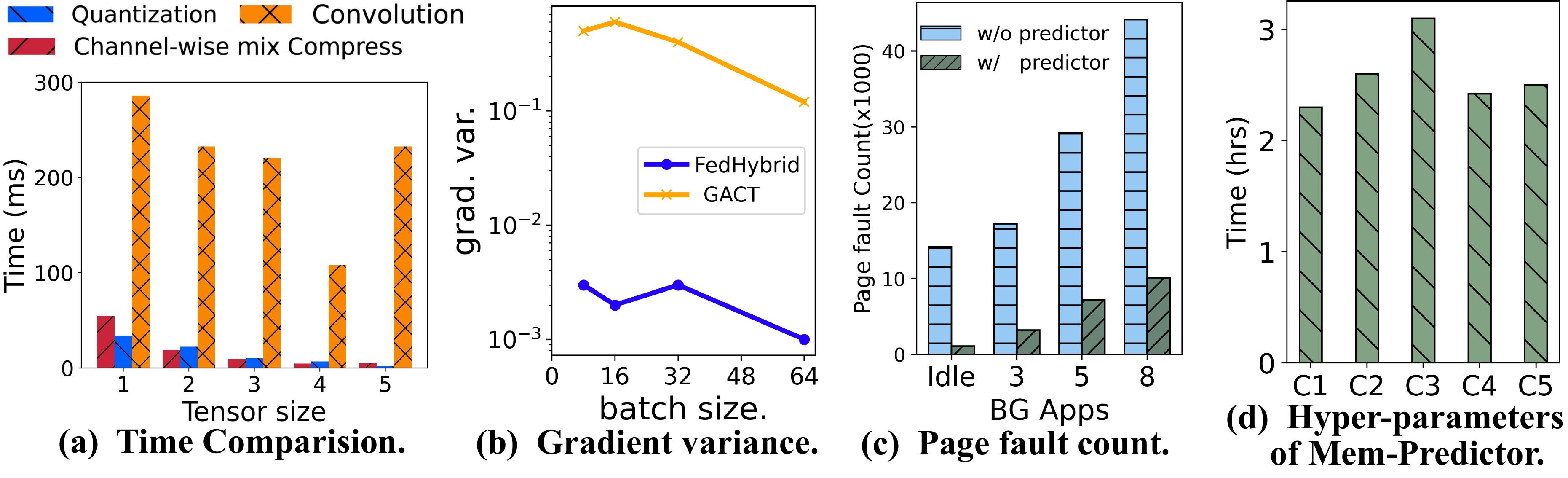}
 \vspace{-19 pt}
	\caption{Ablation study of Local training engine. (a) The tensor sizes from id = 1 to 5 are: 1) 64,112,112 2) 64,56,56 3) 128,28,28 4) 256,14,14 5) 512,7,7. (d) Hyper-parameter of ($TP_1$, $TP_2$): C1(2,3), C2(1.5,3), C3(3,3), C4(2,4), C5(2,5).}
	\label{noniiddis} 
 \vspace{-5pt}
\end{figure}

\begin{figure}[!t] 
 % \vspace{-20pt}
	\centering
\includegraphics[width=1\linewidth]{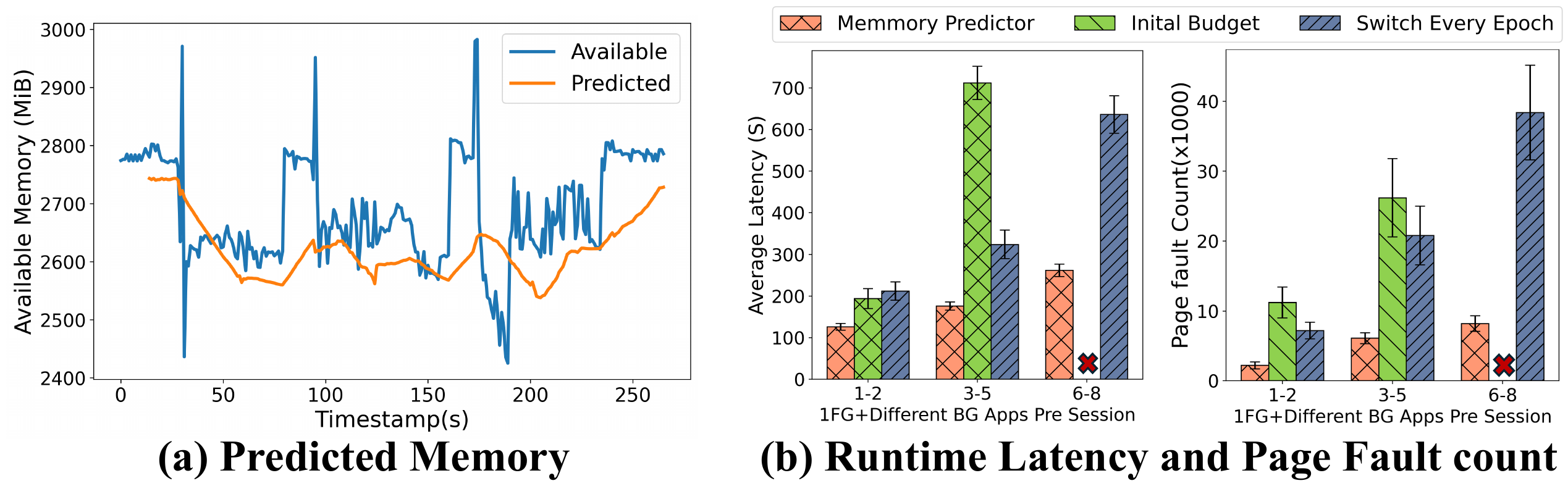}
 \vspace{-20 pt}
	\caption{Evaluation of the Memory Budget Predictor on real user traces from the Carat \cite{oliner2013carat} dataset. We conducted 5 training iterations on 30 example user app usage traces on the S22. (a) The predicted memory budget for an example trace from the predictor. (b) The mean training runtime latency and page fault count among the 30 example traces.}
	\label{memorypredicotr_exp} 
 \vspace{-12pt}
\end{figure}

\noindent \textbf{Impact of Channel-wise Mix Compression.} Disabling compression of the optimizer significantly slows convergence speed, as shown in Figure~\ref{abltation1}, with speeds 1.1$\times$ and 2.2$\times$ slower across both tasks due to the recomputation overhead for each client. A comparison of representative tensors reveals that the time required for quantization and \textit{FedHybrid}'s compression is comparable and much lower than that for convolution, highlighting the efficiency gains from compression, as shown in Figure~\ref{noniiddis}(a). Figure~\ref{noniiddis}(b) shows a decrease in gradient variance of restored tensors after decompression, leading to minimal accuracy degradation.
% This demonstrates the effectiveness of compression in improving training efficiency while maintaining accuracy.

\noindent \textbf{Impact of Memory Budget Predictor.} Replacing the Memory Budget Predictor with direct memory measurement significantly slows convergence speed, as shown in Figure~\ref{abltation1}, with speeds 1.6$\times$ and 2.4$\times$ slower in both tasks due to inaccurate memory measurements affected by dynamic background applications on mobile devices. This inaccuracy impacts client selection and execution plan generation. Figure \ref{noniiddis}(c) demonstrates that using the Memory Budget Predictor decreases page faults, reducing runtime overhead. Additionally, Figure \ref{noniiddis}(d) shows the total training time for MobileNetV2 with different hyperparameters $TP_1$ and $TP_2$. Increasing $TP_1$ reduces the frequency of plan regeneration triggers, leading to more time spent on disk swapping due to memory contention from inaccurate local memory measurements. To further validate the predictor's effectiveness, we use 30 real-world user traces from the Carat \cite{oliner2013carat} dataset with S22 for 5 rounds of on-device training on MobileNetV2. Figure \ref{memorypredicotr_exp}(a) shows the predicted memory according to an example trace. The predictor provides a stable memory budget range for dynamic memory usage, with available memory variance lower than 172.2MB compared to the 576.8MB in practice. This demonstrates the predictor's ability to mitigate fluctuations caused by dynamic usage, applying a lower window weight to fluctuations. Figure \ref{memorypredicotr_exp}(b) reports the average runtime overhead among 30 traces with different training budgets. We compare the memory budget predictor with two baselines: (1) using initial available memory as a fixed budget, and (2) changing the budget every training round. The results show that without the memory predictor, training latency can increase up to 6.9$\times$ when using initial available memory, and up to 2.2$\times$ overhead when changing the budget every round, particularly when there are many background apps. Thus, the memory predictor can effectively handles the dynamicity of diverse client environments and provides more stable and efficient training performance

\subsection{Overhead Analysis}
% \vspace{-3pt}
\noindent \textbf{Client's Overhead.} The client's overhead consists of two main parts. First, the Memory Budget Predictor runs as a background service in the Android system, monitoring available memory asynchronously to prevent page access suspension. On the S22 device, it averages 3-5\% CPU usage, as measured by Perfetto\cite{Perfetto}. Second, dynamic changes in the memory budget may require plan regeneration, incurring additional communication costs. Figure \ref{fig:overhead_with_budget}(c) shows that even with five regeneration requests per ten local training rounds, the total communication time remains below 6\% of the overall training time, indicating minimal additional overhead.

\begin{figure}[!t]
    \centering
    % \begin{minipage}{0.45\linewidth}
    %     \includegraphics[width=1\textwidth]{Figures/mobilenet_32_oneplus_middle_core.pdf}
    %      \vspace{-1em}
    % \end{minipage}
    % \begin{minipage}{0.45\linewidth}
    %     \includegraphics[width=1\textwidth]{Figures/bert_8_oneplus_middle_core.pdf}
    %     \vspace{-1em}
    % \end{minipage}   
    % \vspace{-1pt}
    % \includegraphics[width=\linewidth]{Figures/Execution_time_vsbudget.pdf}
    \includegraphics[width=0.95\linewidth]{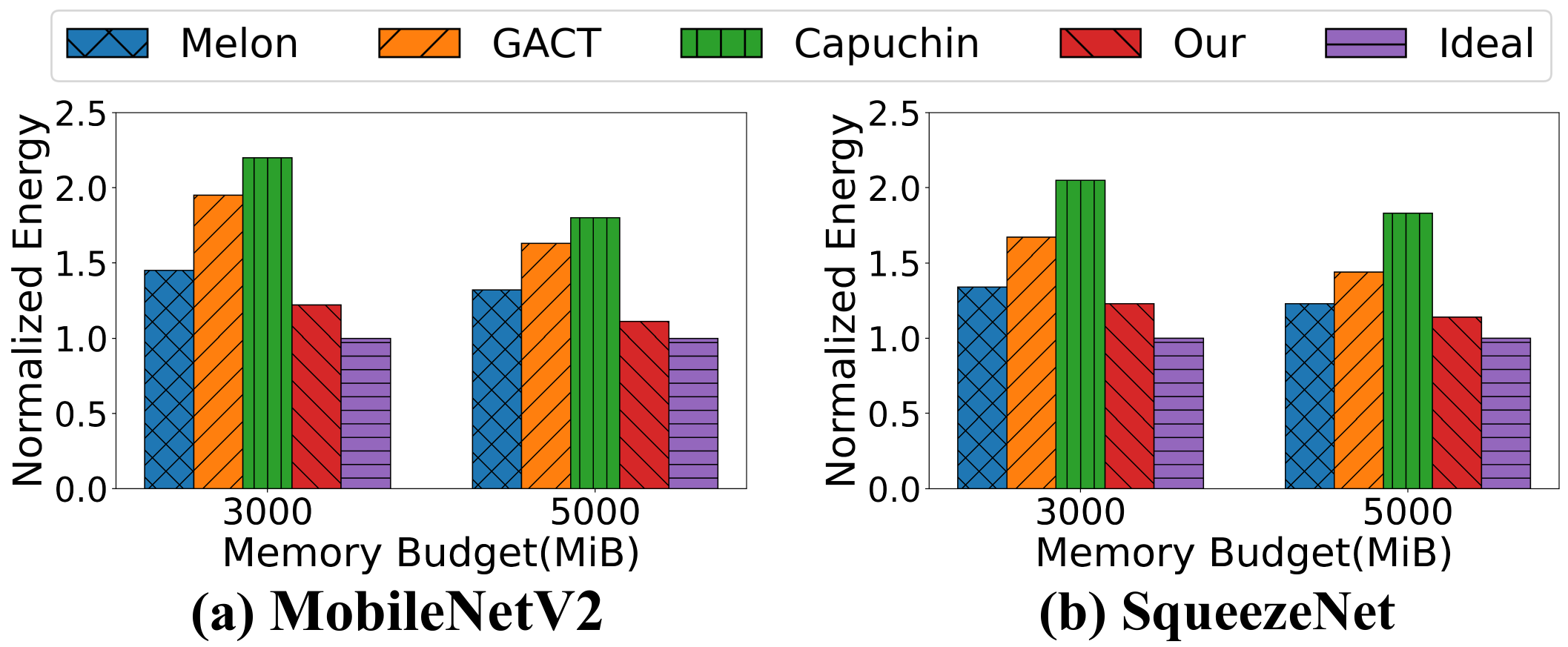}
     \vspace{-10pt}
    % \caption{Execution time under various budget ratio for MobileNetV2 (Left) and Bert (Right).}
    \caption{The energy consumption of \textit{FedHybrid}. This Ideal scenario represents local training with sufficient memory capacity to complete normal training without any page fault overhead. All results are normalized to this Ideal baseline.}
   \vspace{-15pt}
    \label{fig:overhead_with_energy}
\end{figure}

\noindent\textbf{Energy Consumption.} We measure energy usage using the Monsoon Power Monitor \cite{High_voltage_power_moniter}, testing MobileNetV2 and SqueezeNet with a batch size of 16 on an OnePlus10 Pro device under varying memory budgets (Figure \ref{fig:overhead_with_energy}). This Ideal scenario represents local training with sufficient memory capacity to complete normal training without any page fault overhead. \ourmodel reduces energy consumption by 1.1$\times$-1.8$\times$ compared to baselines. Compared to an ideal baseline, \ourmodel only increases energy use by an average of 14.2\% in different memory budgets. This moderate increase in energy consumption is attributed to our method's ability to significantly reduce the time overhead associated with a hybrid-tensor saving strategy. By substantially shortening training completion times, \ourmodel effectively minimizes overall energy consumption. In contrast, other methods like Capuchin, while having lower power consumption for I/O operations during swapping compared to computation, ultimately consume more energy due to prolonged training time. 
% Our approach demonstrates that by optimizing training speed, we can achieve notable energy savings even when implementing memory-saving strategies, outperforming methods that may have lower instantaneous power draw but longer overall execution times.
% While it increases energy use by an average of 16.4\% under limited memory compared to an ideal baseline, this increase is only 18.3\% with sufficient memory. 
% The main energy costs come from compression and recomputation, but \ourmodel remains more efficient than swapping-based methods due to less energy-intensive read/write operations and reduced overall training time.

% \noindent \textcolor{blue}{\textbf{Energy Consumption.} To measure the energy usage of our approach, we used the Monsoon Power Monitor \cite{High_voltage_power_moniter} to evaluate power consumption during local training with varying memory budgets. We tested two models (MobileNetV2 and BERT) on the S22 device, with results illustrated in Figure \ref{fig:overhead_with_energy}. Our results show that \ourmodel reduces energy consumption by 1.2$\times$-2.6$\times$ compared to the baselines. While the energy consumption of \ourmodel increases by 52.4\% on average compared to an ideal baseline under limited memory budgets, it is as low as 18.3\% when memory is sufficient. The primary energy consumption stems from compression and recomputation, but \ourmodel remains more efficient than other methods using swapping. This is because read/write operations are less energy-intensive than computation, and \ourmodel also reduces overall training time.}

% \vspace{-1pt}
\noindent \textbf{Server's Overhead.} 
The server's overhead primarily arises from generating plans for a large number of clients. Figure \ref{fig:overhead_with_budget}(b) illustrates the server's plan generation time across different participant scales. The results highlight the importance of our clustering and cache-based optimization, demonstrating that it can be up to 647$\times$ faster than generating plans individually.

% \vspace{-5pt}

\section{Related Works}
% \vspace{-3pt}
% \textcolor{blue}{
\textbf{Memory-Efficient Training on Mobile Devices}: 
Several approaches \cite{gim2022memory,wang2022melon} have recently been proposed to reduce the memory footprint for on-device training. 
% Although a number of studies have developed memory-saving techniques such as gradient checkpointing \cite{chen2016training,checkmate_mlsys2020_196,kirisame2020dynamic}, activation compression \cite{chen2021actnn,liu2022gact}, and swapping \cite{peng2020capuchin,wang2018superneurons,huang2020swapadvisor} for server-based training, only a few focus on memory efficiency for on-device training on mobile devices. 
For instance, Melon \cite{gim2022memory} reduces the memory footprint of on-device training through recomputation and a lifetime-aware memory pool. Sage \cite{gim2022memory} adopts multiple memory management and saving techniques, including micro batch, operator fusion, and gradient checkpointing, to surmount memory constraints on mobile devices. 
% However, these methods primarily focus on single-device scenarios. 
% However, these methods, primarily designed for single-device, face significant challenges in FL environments. First, micro-batch techniques they use can degrade model performance in FL due to limited local data and non-IID distribution across clients. Their recomputation strategies significantly increase training latency, especially problematic given the heterogeneous capabilities of FL clients. Moreover, their memory management approaches struggle to adapt to the dynamic memory environments typical in FL, where device availability fluctuates due to concurrent applications. These issues can lead to extended training times, increased energy consumption, and limited scalability in FL settings. In contrast, our approach combines recomputation with novel Channel-wise Mix Compression for faster on-device training and incorporates a dynamic memory budget predictor tailored for FL, addressing both memory efficiency and training speed across diverse devices in FL environments.
However, directly employing these memory-saving techniques designed for a single device not only degrades the model performance but also deteriorates the training efficiency due to the following characteristics of the FL system. Micro-batch techniques can degrade model performance in FL due to limited local data and non-IID distributions across clients, leading to noisy gradient estimates and poor generalization. Recomputation strategies substantially increase training latency in each FL round, particularly problematic for heterogeneous client devices with varying computational capabilities. This prolongs overall training time and limits FL's scalability in large-scale deployments. Furthermore, the memory management approaches of Melon and Sage are not well-suited for the dynamic memory environments typical in FL. Clients have varying and fluctuating memory availability due to concurrent applications, making static or arbitrary memory budgets ineffective. While these methods have mechanisms to adapt to changing memory budgets, frequent adjustments introduce additional computational overhead, increasing energy consumption and training duration—particularly problematic for battery-powered devices. 
To address these challenges, our approach combines recomputation with a novel Channel-wise Mix Compression technique and incorporates a dynamic memory budget predictor, effectively balancing memory efficiency and training speed across diverse devices in FL environments.

\noindent
% \textcolor{blue}{
\textbf{System Optimization for FL:} Mobile devices often face resource constraints that lead to performance bottlenecks in FL from a system perspective. Several studies \cite{lai2021oort,li2022pyramidfl,shin2022fedbalancer} have aimed to tackle system heterogeneity in FL, primarily focusing on optimizing device compute speeds and communication bandwidths by client selection and coordination. Oort \cite{lai2021oort} designs a guide-based client selection metric to choose participants for FL based on devices' resource conditions or training data distributions.
However, this type of approach tends to tend to overlook the critical issue of available memory, which is essential for a device to participate in training. Without sufficient memory, a device is excluded from the training process, thereby limiting the diversity and effectiveness of the FL system. Recently, Partial training \cite{alam2022fedrolex,diao2020heterofl,horvath2021fjord} has been introduced to mitigate computational resource limitations by training lower-complexity submodels on devices. For example, HeteroFL \cite{diao2020heterofl} trains a lower complexity submodel according to device capability and integrates it into the full global model.
However, this approach reduces memory usage at the expense of model performance, as it requires discarding numerous filters and parameters, compromising the model architecture. In contrast, our work directly addresses memory constraints while preserving model integrity and performance. By dynamically selecting memory-constrained clients and optimizing their memory usage, our approach enables more devices to participate effectively in FL.
\section{Conclusion}

In this paper, we present \textit{FedHybrid}, a memory-efficient federated learning framework tailored for devices with limited memory. By incorporating a Memory-aware Client Selector, a Heterogeneity-aware Graph Optimizer, and a Local Training Engine, \ourmodel significantly enhances training efficiency. Our experiments demonstrate that \ourmodel boosts model accuracy by up to 39.1\% and cuts training time by up to $15.5 \times$ compared to conventional methods.

% \section{Acknowledgement}
\begin{acks}
We sincerely appreciate the anonymous shepherd and reviewers for their valuable comments. 
This research was supported in part by the MYRG-GRG2023-00211-IOTSC-UMDF and SRG2022-00010-IOTSC grants from the University of Macau.
\end{acks}
% (Robust Federated Learning for Noisy Labels)
% (AutoPerf: End-to-End Driving Performance Control Middleware for Autonomous Vehicle).
% National Key R\&D Program of China under Grant No. 2022YFF0604502, the National Natural Science Foundation of China under Grant No.62122095, 62341201, 62072472, U19A2067 and 92067206, and by a grant from the Guoqiang Institute, Tsinghua University.

% \clearpage

\balance
\bibliographystyle{ACM-Reference-Format}
\bibliography{reference}


\begin{thebibliography}{59}

%%% the Web address.

\ifx \showCODEN    \undefined \def \showCODEN     #1{\unskip}     \fi
\ifx \showDOI      \undefined \def \showDOI       #1{#1}\fi
\ifx \showISBNx    \undefined \def \showISBNx     #1{\unskip}     \fi
\ifx \showISBNxiii \undefined \def \showISBNxiii  #1{\unskip}     \fi
\ifx \showISSN     \undefined \def \showISSN      #1{\unskip}     \fi
\ifx \showLCCN     \undefined \def \showLCCN      #1{\unskip}     \fi
\ifx \shownote     \undefined \def \shownote      #1{#1}          \fi
\ifx \showarticletitle \undefined \def \showarticletitle #1{#1}   \fi
\ifx \showURL      \undefined \def \showURL       {\relax}        \fi
% invisible to TeX
\providecommand\bibfield[2]{#2}
\providecommand\bibinfo[2]{#2}
\providecommand\natexlab[1]{#1}
\providecommand\showeprint[2][]{arXiv:#2}

\bibitem[Alam et~al\mbox{.}(2022)]%
        {alam2022fedrolex}
\bibfield{author}{\bibinfo{person}{Samiul Alam}, \bibinfo{person}{Luyang Liu}, \bibinfo{person}{Ming Yan}, {and} \bibinfo{person}{Mi Zhang}.} \bibinfo{year}{2022}\natexlab{}.
\newblock \showarticletitle{Fedrolex: Model-heterogeneous federated learning with rolling sub-model extraction}.
\newblock \bibinfo{journal}{\emph{Advances in Neural Information Processing Systems}}  \bibinfo{volume}{35} (\bibinfo{year}{2022}), \bibinfo{pages}{29677--29690}.
\newblock


\bibitem[Bonawitz et~al\mbox{.}(2019)]%
        {bonawitz2019towards}
\bibfield{author}{\bibinfo{person}{Keith Bonawitz}, \bibinfo{person}{Hubert Eichner}, \bibinfo{person}{Wolfgang Grieskamp}, \bibinfo{person}{Dzmitry Huba}, \bibinfo{person}{Alex Ingerman}, \bibinfo{person}{Vladimir Ivanov}, \bibinfo{person}{Chloe Kiddon}, \bibinfo{person}{Jakub Kone{\v{c}}n{\`y}}, \bibinfo{person}{Stefano Mazzocchi}, \bibinfo{person}{Brendan McMahan}, {et~al\mbox{.}}} \bibinfo{year}{2019}\natexlab{}.
\newblock \showarticletitle{Towards federated learning at scale: System design}.
\newblock \bibinfo{journal}{\emph{Proceedings of machine learning and systems}}  \bibinfo{volume}{1} (\bibinfo{year}{2019}), \bibinfo{pages}{374--388}.
\newblock


\bibitem[Chen et~al\mbox{.}(2021)]%
        {chen2021actnn}
\bibfield{author}{\bibinfo{person}{Jianfei Chen}, \bibinfo{person}{Lianmin Zheng}, \bibinfo{person}{Zhewei Yao}, \bibinfo{person}{Dequan Wang}, \bibinfo{person}{Ion Stoica}, \bibinfo{person}{Michael Mahoney}, {and} \bibinfo{person}{Joseph Gonzalez}.} \bibinfo{year}{2021}\natexlab{}.
\newblock \showarticletitle{Actnn: Reducing training memory footprint via 2-bit activation compressed training}. In \bibinfo{booktitle}{\emph{Proceedings of ICML}}. PMLR, \bibinfo{pages}{1803--1813}.
\newblock


\bibitem[Chen et~al\mbox{.}(2016)]%
        {chen2016training}
\bibfield{author}{\bibinfo{person}{Tianqi Chen}, \bibinfo{person}{Bing Xu}, \bibinfo{person}{Chiyuan Zhang}, {and} \bibinfo{person}{Carlos Guestrin}.} \bibinfo{year}{2016}\natexlab{}.
\newblock \showarticletitle{Training deep nets with sublinear memory cost}.
\newblock \bibinfo{journal}{\emph{arXiv preprint arXiv:1604.06174}} (\bibinfo{year}{2016}).
\newblock


\bibitem[Chen et~al\mbox{.}(2013)]%
        {chen2013combinatorial}
\bibfield{author}{\bibinfo{person}{Wei Chen}, \bibinfo{person}{Yajun Wang}, {and} \bibinfo{person}{Yang Yuan}.} \bibinfo{year}{2013}\natexlab{}.
\newblock \showarticletitle{Combinatorial multi-armed bandit: General framework and applications}. In \bibinfo{booktitle}{\emph{Proceedings of ICML}}. PMLR, \bibinfo{pages}{151--159}.
\newblock


\bibitem[Cho et~al\mbox{.}(2020)]%
        {cho2020client}
\bibfield{author}{\bibinfo{person}{Yae~Jee Cho}, \bibinfo{person}{Jianyu Wang}, {and} \bibinfo{person}{Gauri Joshi}.} \bibinfo{year}{2020}\natexlab{}.
\newblock \showarticletitle{Client selection in federated learning: Convergence analysis and power-of-choice selection strategies}.
\newblock \bibinfo{journal}{\emph{arXiv preprint arXiv:2010.01243}} (\bibinfo{year}{2020}).
\newblock


\bibitem[Defferrard et~al\mbox{.}(2016)]%
        {defferrard2016convolutional}
\bibfield{author}{\bibinfo{person}{Micha{\"e}l Defferrard}, \bibinfo{person}{Xavier Bresson}, {and} \bibinfo{person}{Pierre Vandergheynst}.} \bibinfo{year}{2016}\natexlab{}.
\newblock \showarticletitle{Convolutional neural networks on graphs with fast localized spectral filtering}.
\newblock \bibinfo{journal}{\emph{Advances in neural information processing systems}}  \bibinfo{volume}{29} (\bibinfo{year}{2016}).
\newblock


\bibitem[Devlin et~al\mbox{.}(2018)]%
        {devlin2018bert}
\bibfield{author}{\bibinfo{person}{Jacob Devlin}, \bibinfo{person}{Ming-Wei Chang}, \bibinfo{person}{Kenton Lee}, {and} \bibinfo{person}{Kristina Toutanova}.} \bibinfo{year}{2018}\natexlab{}.
\newblock \showarticletitle{Bert: Pre-training of deep bidirectional transformers for language understanding}.
\newblock \bibinfo{journal}{\emph{arXiv preprint arXiv:1810.04805}} (\bibinfo{year}{2018}).
\newblock


\bibitem[Diao et~al\mbox{.}(2020)]%
        {diao2020heterofl}
\bibfield{author}{\bibinfo{person}{Enmao Diao}, \bibinfo{person}{Jie Ding}, {and} \bibinfo{person}{Vahid Tarokh}.} \bibinfo{year}{2020}\natexlab{}.
\newblock \showarticletitle{Heterofl: Computation and communication efficient federated learning for heterogeneous clients}.
\newblock \bibinfo{journal}{\emph{arXiv preprint arXiv:2010.01264}} (\bibinfo{year}{2020}).
\newblock


\bibitem[Dwork(2008)]%
        {dwork2008differential}
\bibfield{author}{\bibinfo{person}{Cynthia Dwork}.} \bibinfo{year}{2008}\natexlab{}.
\newblock \showarticletitle{Differential privacy: A survey of results}. In \bibinfo{booktitle}{\emph{International conference on theory and applications of models of computation}}. Springer, \bibinfo{pages}{1--19}.
\newblock


\bibitem[Evans and Aamodt(2021)]%
        {evans2021ac}
\bibfield{author}{\bibinfo{person}{R~David Evans} {and} \bibinfo{person}{Tor Aamodt}.} \bibinfo{year}{2021}\natexlab{}.
\newblock \showarticletitle{Ac-gc: Lossy activation compression with guaranteed convergence}.
\newblock \bibinfo{journal}{\emph{Advances in Neural Information Processing Systems}}  \bibinfo{volume}{34} (\bibinfo{year}{2021}), \bibinfo{pages}{27434--27448}.
\newblock


\bibitem[Evans et~al\mbox{.}(2020)]%
        {evans2020jpeg}
\bibfield{author}{\bibinfo{person}{R~David Evans}, \bibinfo{person}{Lufei Liu}, {and} \bibinfo{person}{Tor~M Aamodt}.} \bibinfo{year}{2020}\natexlab{}.
\newblock \showarticletitle{Jpeg-act: accelerating deep learning via transform-based lossy compression}. In \bibinfo{booktitle}{\emph{Proceedings of ACM/IEEE ISCA}}. IEEE, \bibinfo{pages}{860--873}.
\newblock


\bibitem[{Gary Sims}(2023)]%
        {ramstatisticsinfo}
\bibfield{author}{\bibinfo{person}{{Gary Sims}}.} \bibinfo{year}{2023}\natexlab{}.
\newblock \bibinfo{title}{{How much RAM does your Android phone really need in 2023?}}
\newblock \bibinfo{howpublished}{\url{https://www.androidauthority.com/how-much-ram-do-i-need-phone-3086661/}}.
\newblock
\newblock
\shownote{Accessed: 2023.12}.


\bibitem[Gim and Ko(2022)]%
        {gim2022memory}
\bibfield{author}{\bibinfo{person}{In Gim} {and} \bibinfo{person}{JeongGil Ko}.} \bibinfo{year}{2022}\natexlab{}.
\newblock \showarticletitle{Memory-efficient DNN training on mobile devices}. In \bibinfo{booktitle}{\emph{Proceedings of the 20th Annual International Conference on Mobile Systems, Applications and Services}}. \bibinfo{pages}{464--476}.
\newblock


\bibitem[Google(2023a)]%
        {zram}
\bibfield{author}{\bibinfo{person}{Google}.} \bibinfo{year}{2023}\natexlab{a}.
\newblock \bibinfo{title}{{samsung-galaxy-s21-ram-plus-update}}.
\newblock \bibinfo{howpublished}{\url{https://developer.android.com/topic/performance/memory-management }}.
\newblock
\newblock
\shownote{Accessed: 2023.12}.


\bibitem[Google(2023b)]%
        {Perfetto}
\bibfield{author}{\bibinfo{person}{Google}.} \bibinfo{year}{2023}\natexlab{b}.
\newblock \bibinfo{title}{{System profiling, app tracing and trace analysis}}.
\newblock \bibinfo{howpublished}{\url{https://perfetto.dev/}}.
\newblock
\newblock
\shownote{Accessed: 2023.12}.


\bibitem[Google(2023c)]%
        {Monkey}
\bibfield{author}{\bibinfo{person}{Google}.} \bibinfo{year}{2023}\natexlab{c}.
\newblock \bibinfo{title}{{UI/Application Exerciser Monkey.}}
\newblock \bibinfo{howpublished}{\url{https://developer.android.com/studio/test/other-testing-tools/monkey}}.
\newblock
\newblock
\shownote{Accessed: 2023.12}.


\bibitem[Google(2024a)]%
        {ActivityManager}
\bibfield{author}{\bibinfo{person}{Google}.} \bibinfo{year}{2024}\natexlab{a}.
\newblock \bibinfo{title}{ActivityManager}.
\newblock \bibinfo{howpublished}{\url{https://developer.android.com/reference/android/app/ActivityManager.AppTask }}.
\newblock
\newblock
\shownote{Accessed: 2023.12}.


\bibitem[Google(2024b)]%
        {oomscore}
\bibfield{author}{\bibinfo{person}{Google}.} \bibinfo{year}{2024}\natexlab{b}.
\newblock \bibinfo{title}{{oom-adj-score}}.
\newblock \bibinfo{howpublished}{\url{https://developer.android.com/topic/performance/memory-management\#low-memory_killer }}.
\newblock
\newblock
\shownote{Accessed: 2023.12}.


\bibitem[Guo et~al\mbox{.}(2023)]%
        {guo2023olive}
\bibfield{author}{\bibinfo{person}{Cong Guo}, \bibinfo{person}{Jiaming Tang}, \bibinfo{person}{Weiming Hu}, \bibinfo{person}{Jingwen Leng}, \bibinfo{person}{Chen Zhang}, \bibinfo{person}{Fan Yang}, \bibinfo{person}{Yunxin Liu}, \bibinfo{person}{Minyi Guo}, {and} \bibinfo{person}{Yuhao Zhu}.} \bibinfo{year}{2023}\natexlab{}.
\newblock \showarticletitle{Olive: Accelerating large language models via hardware-friendly outlier-victim pair quantization}. In \bibinfo{booktitle}{\emph{Proceedings of ACM/IEEE ISCA}}. \bibinfo{pages}{1--15}.
\newblock


\bibitem[Hackborn(2013)]%
        {HOME_APP_ADJ}
\bibfield{author}{\bibinfo{person}{Dianne Hackborn}.} \bibinfo{year}{2013}\natexlab{}.
\newblock \bibinfo{title}{ProcessList.java}.
\newblock \bibinfo{howpublished}{\url{https://android.googlesource.com/platform/frameworks/base/+/6285a32/services/java/com/android/server/am/ProcessList.java}}.
\newblock
\newblock
\shownote{Accessed: 2023.12}.


\bibitem[Han and Shin(2020)]%
        {han2020command}
\bibfield{author}{\bibinfo{person}{Kyuhwa Han} {and} \bibinfo{person}{Dongkun Shin}.} \bibinfo{year}{2020}\natexlab{}.
\newblock \showarticletitle{Command queue-aware host I/O stack for mobile flash storage}.
\newblock \bibinfo{journal}{\emph{Journal of systems architecture}}  \bibinfo{volume}{109} (\bibinfo{year}{2020}), \bibinfo{pages}{101758}.
\newblock


\bibitem[Heo et~al\mbox{.}(2023)]%
        {heo2023rethinking}
\bibfield{author}{\bibinfo{person}{Jung~Hwan Heo}, \bibinfo{person}{Jeonghoon Kim}, \bibinfo{person}{Beomseok Kwon}, \bibinfo{person}{Byeongwook Kim}, \bibinfo{person}{Se~Jung Kwon}, {and} \bibinfo{person}{Dongsoo Lee}.} \bibinfo{year}{2023}\natexlab{}.
\newblock \showarticletitle{Rethinking channel dimensions to isolate outliers for low-bit weight quantization of large language models}.
\newblock \bibinfo{journal}{\emph{arXiv preprint arXiv:2309.15531}} (\bibinfo{year}{2023}).
\newblock


\bibitem[Horvath et~al\mbox{.}(2021)]%
        {horvath2021fjord}
\bibfield{author}{\bibinfo{person}{Samuel Horvath}, \bibinfo{person}{Stefanos Laskaridis}, \bibinfo{person}{Mario Almeida}, \bibinfo{person}{Ilias Leontiadis}, \bibinfo{person}{Stylianos Venieris}, {and} \bibinfo{person}{Nicholas Lane}.} \bibinfo{year}{2021}\natexlab{}.
\newblock \showarticletitle{Fjord: Fair and accurate federated learning under heterogeneous targets with ordered dropout}.
\newblock \bibinfo{journal}{\emph{Advances in Neural Information Processing Systems}}  \bibinfo{volume}{34} (\bibinfo{year}{2021}), \bibinfo{pages}{12876--12889}.
\newblock


\bibitem[Huang et~al\mbox{.}(2020)]%
        {huang2020swapadvisor}
\bibfield{author}{\bibinfo{person}{Chien-Chin Huang}, \bibinfo{person}{Gu Jin}, {and} \bibinfo{person}{Jinyang Li}.} \bibinfo{year}{2020}\natexlab{}.
\newblock \showarticletitle{Swapadvisor: Pushing deep learning beyond the gpu memory limit via smart swapping}. In \bibinfo{booktitle}{\emph{Proceedings of ACM ASPLOS}}. \bibinfo{pages}{1341--1355}.
\newblock


\bibitem[Ignatov et~al\mbox{.}(2023)]%
        {AIBenchmark}
\bibfield{author}{\bibinfo{person}{Andrey Ignatov}, \bibinfo{person}{Radu Timofte}, \bibinfo{person}{Andrei Kulik}, \bibinfo{person}{Seungsoo Yang}, \bibinfo{person}{Ke Wang}, \bibinfo{person}{Felix Baum}, \bibinfo{person}{Max Wu}, \bibinfo{person}{Lirong Xu}, {and} \bibinfo{person}{Luc Van~Gool}.} \bibinfo{year}{2023}\natexlab{}.
\newblock \bibinfo{title}{{AI Benchmark: All About Deep Learning on Smartphones.}}
\newblock \bibinfo{howpublished}{\url{https://ai-benchmark.com/ranking.html}}.
\newblock
\newblock
\shownote{Accessed: 2023.12}.


\bibitem[Jain et~al\mbox{.}(2020)]%
        {checkmate_mlsys2020_196}
\bibfield{author}{\bibinfo{person}{Paras Jain}, \bibinfo{person}{Ajay Jain}, \bibinfo{person}{Aniruddha Nrusimha}, \bibinfo{person}{Amir Gholami}, \bibinfo{person}{Pieter Abbeel}, \bibinfo{person}{Joseph Gonzalez}, \bibinfo{person}{Kurt Keutzer}, {and} \bibinfo{person}{Ion Stoica}.} \bibinfo{year}{2020}\natexlab{}.
\newblock \showarticletitle{Checkmate: Breaking the Memory Wall with Optimal Tensor Rematerialization}.
\newblock In \bibinfo{booktitle}{\emph{Proceedings of Machine Learning and Systems 2020}}. \bibinfo{pages}{497--511}.
\newblock


\bibitem[Jiang et~al\mbox{.}(2020)]%
        {alibaba2020mnn}
\bibfield{author}{\bibinfo{person}{Xiaotang Jiang}, \bibinfo{person}{Huan Wang}, \bibinfo{person}{Yiliu Chen}, \bibinfo{person}{Ziqi Wu}, \bibinfo{person}{Lichuan Wang}, \bibinfo{person}{Bin Zou}, \bibinfo{person}{Yafeng Yang}, \bibinfo{person}{Zongyang Cui}, \bibinfo{person}{Yu Cai}, \bibinfo{person}{Tianhang Yu}, \bibinfo{person}{Chengfei Lv}, {and} \bibinfo{person}{Zhihua Wu}.} \bibinfo{year}{2020}\natexlab{}.
\newblock \showarticletitle{MNN: A Universal and Efficient Inference Engine}. In \bibinfo{booktitle}{\emph{Proceedings of MLSys}}.
\newblock


\bibitem[Kairouz et~al\mbox{.}(2021)]%
        {kairouz2021advances}
\bibfield{author}{\bibinfo{person}{Peter Kairouz}, \bibinfo{person}{H~Brendan McMahan}, \bibinfo{person}{Brendan Avent}, \bibinfo{person}{Aur{\'e}lien Bellet}, \bibinfo{person}{Mehdi Bennis}, \bibinfo{person}{Arjun~Nitin Bhagoji}, \bibinfo{person}{Kallista Bonawitz}, \bibinfo{person}{Zachary Charles}, \bibinfo{person}{Graham Cormode}, \bibinfo{person}{Rachel Cummings}, {et~al\mbox{.}}} \bibinfo{year}{2021}\natexlab{}.
\newblock \showarticletitle{Advances and open problems in federated learning}.
\newblock \bibinfo{journal}{\emph{Foundations and Trends{\textregistered} in Machine Learning}} \bibinfo{volume}{14}, \bibinfo{number}{1--2} (\bibinfo{year}{2021}), \bibinfo{pages}{1--210}.
\newblock


\bibitem[Kirisame et~al\mbox{.}(2020)]%
        {kirisame2020dynamic}
\bibfield{author}{\bibinfo{person}{Marisa Kirisame}, \bibinfo{person}{Steven Lyubomirsky}, \bibinfo{person}{Altan Haan}, \bibinfo{person}{Jennifer Brennan}, \bibinfo{person}{Mike He}, \bibinfo{person}{Jared Roesch}, \bibinfo{person}{Tianqi Chen}, {and} \bibinfo{person}{Zachary Tatlock}.} \bibinfo{year}{2020}\natexlab{}.
\newblock \showarticletitle{Dynamic tensor rematerialization}.
\newblock \bibinfo{journal}{\emph{arXiv preprint arXiv:2006.09616}} (\bibinfo{year}{2020}).
\newblock


\bibitem[Kuznetsova et~al\mbox{.}(2020)]%
        {kuznetsova2020open}
\bibfield{author}{\bibinfo{person}{Alina Kuznetsova}, \bibinfo{person}{Hassan Rom}, \bibinfo{person}{Neil Alldrin}, \bibinfo{person}{Jasper Uijlings}, \bibinfo{person}{Ivan Krasin}, \bibinfo{person}{Jordi Pont-Tuset}, \bibinfo{person}{Shahab Kamali}, \bibinfo{person}{Stefan Popov}, \bibinfo{person}{Matteo Malloci}, \bibinfo{person}{Alexander Kolesnikov}, {et~al\mbox{.}}} \bibinfo{year}{2020}\natexlab{}.
\newblock \showarticletitle{The open images dataset v4: Unified image classification, object detection, and visual relationship detection at scale}.
\newblock \bibinfo{journal}{\emph{International Journal of Computer Vision}} \bibinfo{volume}{128}, \bibinfo{number}{7} (\bibinfo{year}{2020}), \bibinfo{pages}{1956--1981}.
\newblock


\bibitem[Lai et~al\mbox{.}(2022)]%
        {lai2022fedscale}
\bibfield{author}{\bibinfo{person}{Fan Lai}, \bibinfo{person}{Yinwei Dai}, \bibinfo{person}{Sanjay Singapuram}, \bibinfo{person}{Jiachen Liu}, \bibinfo{person}{Xiangfeng Zhu}, \bibinfo{person}{Harsha Madhyastha}, {and} \bibinfo{person}{Mosharaf Chowdhury}.} \bibinfo{year}{2022}\natexlab{}.
\newblock \showarticletitle{Fedscale: Benchmarking model and system performance of federated learning at scale}. In \bibinfo{booktitle}{\emph{Proceedings of ACM ICML}}. PMLR, \bibinfo{pages}{11814--11827}.
\newblock


\bibitem[Lai et~al\mbox{.}(2021)]%
        {lai2021oort}
\bibfield{author}{\bibinfo{person}{Fan Lai}, \bibinfo{person}{Xiangfeng Zhu}, \bibinfo{person}{Harsha~V Madhyastha}, {and} \bibinfo{person}{Mosharaf Chowdhury}.} \bibinfo{year}{2021}\natexlab{}.
\newblock \showarticletitle{Oort: Efficient federated learning via guided participant selection}. In \bibinfo{booktitle}{\emph{15th $\{$USENIX$\}$ Symposium on Operating Systems Design and Implementation ($\{$OSDI$\}$ 21)}}. \bibinfo{pages}{19--35}.
\newblock


\bibitem[Lan et~al\mbox{.}(2019)]%
        {lan2019albert}
\bibfield{author}{\bibinfo{person}{Zhenzhong Lan}, \bibinfo{person}{Mingda Chen}, \bibinfo{person}{Sebastian Goodman}, \bibinfo{person}{Kevin Gimpel}, \bibinfo{person}{Piyush Sharma}, {and} \bibinfo{person}{Radu Soricut}.} \bibinfo{year}{2019}\natexlab{}.
\newblock \showarticletitle{Albert: A lite bert for self-supervised learning of language representations}.
\newblock \bibinfo{journal}{\emph{arXiv preprint arXiv:1909.11942}} (\bibinfo{year}{2019}).
\newblock


\bibitem[Lee et~al\mbox{.}(2021)]%
        {lee2021internal}
\bibfield{author}{\bibinfo{person}{Gyeongyong Lee}, \bibinfo{person}{Jaewook Kwak}, \bibinfo{person}{Joonyong Jeong}, \bibinfo{person}{Daeyong Lee}, \bibinfo{person}{Moonseok Jang}, \bibinfo{person}{Jungwook Choi}, {and} \bibinfo{person}{Yong~Ho Song}.} \bibinfo{year}{2021}\natexlab{}.
\newblock \showarticletitle{Internal Task-Aware Command Scheduling to Improve Read Performance of Embedded Flash Storage Systems}.
\newblock \bibinfo{journal}{\emph{IEEE Access}}  \bibinfo{volume}{9} (\bibinfo{year}{2021}), \bibinfo{pages}{71638--71650}.
\newblock


\bibitem[Li et~al\mbox{.}(2021b)]%
        {li2021hermes}
\bibfield{author}{\bibinfo{person}{Ang Li}, \bibinfo{person}{Jingwei Sun}, \bibinfo{person}{Pengcheng Li}, \bibinfo{person}{Yu Pu}, \bibinfo{person}{Hai Li}, {and} \bibinfo{person}{Yiran Chen}.} \bibinfo{year}{2021}\natexlab{b}.
\newblock \showarticletitle{Hermes: an efficient federated learning framework for heterogeneous mobile clients}. In \bibinfo{booktitle}{\emph{Proceedings of the 27th Annual International Conference on Mobile Computing and Networking}}. \bibinfo{pages}{420--437}.
\newblock


\bibitem[Li et~al\mbox{.}(2022)]%
        {li2022pyramidfl}
\bibfield{author}{\bibinfo{person}{Chenning Li}, \bibinfo{person}{Xiao Zeng}, \bibinfo{person}{Mi Zhang}, {and} \bibinfo{person}{Zhichao Cao}.} \bibinfo{year}{2022}\natexlab{}.
\newblock \showarticletitle{PyramidFL: A fine-grained client selection framework for efficient federated learning}. In \bibinfo{booktitle}{\emph{Proceedings of ACM MobiCom}}. \bibinfo{pages}{158--171}.
\newblock


\bibitem[Li et~al\mbox{.}(2021a)]%
        {li2021understanding}
\bibfield{author}{\bibinfo{person}{Tong Li}, \bibinfo{person}{Yali Fan}, \bibinfo{person}{Yong Li}, \bibinfo{person}{Sasu Tarkoma}, {and} \bibinfo{person}{Pan Hui}.} \bibinfo{year}{2021}\natexlab{a}.
\newblock \showarticletitle{Understanding the long-term evolution of mobile app usage}.
\newblock \bibinfo{journal}{\emph{IEEE Transactions on Mobile Computing}} \bibinfo{volume}{22}, \bibinfo{number}{2} (\bibinfo{year}{2021}), \bibinfo{pages}{1213--1230}.
\newblock


\bibitem[Liang et~al\mbox{.}(2020)]%
        {liang2020acclaim}
\bibfield{author}{\bibinfo{person}{Yu Liang}, \bibinfo{person}{Jinheng Li}, \bibinfo{person}{Rachata Ausavarungnirun}, \bibinfo{person}{Riwei Pan}, \bibinfo{person}{Liang Shi}, \bibinfo{person}{Tei-Wei Kuo}, {and} \bibinfo{person}{Chun~Jason Xue}.} \bibinfo{year}{2020}\natexlab{}.
\newblock \showarticletitle{Acclaim: Adaptive memory reclaim to improve user experience in android systems}. In \bibinfo{booktitle}{\emph{Proceedings of USENIX ATC}}. \bibinfo{pages}{897--910}.
\newblock


\bibitem[Liang et~al\mbox{.}(2022)]%
        {liang2022cachesifter}
\bibfield{author}{\bibinfo{person}{Yu Liang}, \bibinfo{person}{Riwei Pan}, \bibinfo{person}{Tianyu Ren}, \bibinfo{person}{Yufei Cui}, \bibinfo{person}{Rachata Ausavarungnirun}, \bibinfo{person}{Xianzhang Chen}, \bibinfo{person}{Changlong Li}, \bibinfo{person}{Tei-Wei Kuo}, {and} \bibinfo{person}{Chun~Jason Xue}.} \bibinfo{year}{2022}\natexlab{}.
\newblock \showarticletitle{$\{$CacheSifter$\}$: Sifting Cache Files for Boosted Mobile Performance and Lifetime}. In \bibinfo{booktitle}{\emph{20th USENIX Conference on File and Storage Technologies (FAST 22)}}. \bibinfo{pages}{445--459}.
\newblock


\bibitem[Lim et~al\mbox{.}(2023)]%
        {lim2023swam}
\bibfield{author}{\bibinfo{person}{Geunsik Lim}, \bibinfo{person}{Donghyun Kang}, \bibinfo{person}{MyungJoo Ham}, {and} \bibinfo{person}{Young~Ik Eom}.} \bibinfo{year}{2023}\natexlab{}.
\newblock \showarticletitle{SWAM: Revisiting Swap and OOMK for Improving Application Responsiveness on Mobile Devices}.
\newblock \bibinfo{journal}{\emph{arXiv preprint arXiv:2306.08345}} (\bibinfo{year}{2023}).
\newblock


\bibitem[Lin et~al\mbox{.}(2023)]%
        {lin2023awq}
\bibfield{author}{\bibinfo{person}{Ji Lin}, \bibinfo{person}{Jiaming Tang}, \bibinfo{person}{Haotian Tang}, \bibinfo{person}{Shang Yang}, \bibinfo{person}{Xingyu Dang}, {and} \bibinfo{person}{Song Han}.} \bibinfo{year}{2023}\natexlab{}.
\newblock \showarticletitle{Awq: Activation-aware weight quantization for llm compression and acceleration}.
\newblock \bibinfo{journal}{\emph{arXiv preprint arXiv:2306.00978}} (\bibinfo{year}{2023}).
\newblock


\bibitem[Liu et~al\mbox{.}(2022)]%
        {liu2022gact}
\bibfield{author}{\bibinfo{person}{Xiaoxuan Liu}, \bibinfo{person}{Lianmin Zheng}, \bibinfo{person}{Dequan Wang}, \bibinfo{person}{Yukuo Cen}, \bibinfo{person}{Weize Chen}, \bibinfo{person}{Xu Han}, \bibinfo{person}{Jianfei Chen}, \bibinfo{person}{Zhiyuan Liu}, \bibinfo{person}{Jie Tang}, \bibinfo{person}{Joey Gonzalez}, {et~al\mbox{.}}} \bibinfo{year}{2022}\natexlab{}.
\newblock \showarticletitle{GACT: Activation compressed training for generic network architectures}. In \bibinfo{booktitle}{\emph{Proceedings of ACM ICML}}. PMLR, \bibinfo{pages}{14139--14152}.
\newblock


\bibitem[McMahan et~al\mbox{.}(2017)]%
        {mcmahan2017communication}
\bibfield{author}{\bibinfo{person}{Brendan McMahan}, \bibinfo{person}{Eider Moore}, \bibinfo{person}{Daniel Ramage}, \bibinfo{person}{Seth Hampson}, {and} \bibinfo{person}{Blaise~Aguera y Arcas}.} \bibinfo{year}{2017}\natexlab{}.
\newblock \showarticletitle{Communication-efficient learning of deep networks from decentralized data}. In \bibinfo{booktitle}{\emph{Proceedings of AISTATS}}. PMLR, \bibinfo{pages}{1273--1282}.
\newblock


\bibitem[McMahan and Ramage(2017)]%
        {mcmahan2017federated}
\bibfield{author}{\bibinfo{person}{Brendan McMahan} {and} \bibinfo{person}{Daniel Ramage}.} \bibinfo{year}{2017}\natexlab{}.
\newblock \showarticletitle{Federated learning: Collaborative machine learning without centralized training data}.
\newblock \bibinfo{journal}{\emph{Google Research Blog}}  \bibinfo{volume}{3} (\bibinfo{year}{2017}).
\newblock


\bibitem[Monsoon~Solutions(2023)]%
        {High_voltage_power_moniter}
\bibfield{author}{\bibinfo{person}{Inc Monsoon~Solutions}.} \bibinfo{year}{2023}\natexlab{}.
\newblock \bibinfo{title}{High voltage power moniter}.
\newblock \bibinfo{howpublished}{\url{https://www.msoon.com/}}.
\newblock
\newblock
\shownote{Accessed: 2023.12}.


\bibitem[Oliner et~al\mbox{.}(2013)]%
        {oliner2013carat}
\bibfield{author}{\bibinfo{person}{Adam~J Oliner}, \bibinfo{person}{Anand~P Iyer}, \bibinfo{person}{Ion Stoica}, \bibinfo{person}{Eemil Lagerspetz}, {and} \bibinfo{person}{Sasu Tarkoma}.} \bibinfo{year}{2013}\natexlab{}.
\newblock \showarticletitle{Carat: Collaborative energy diagnosis for mobile devices}. In \bibinfo{booktitle}{\emph{Proceedings of the 11th ACM conference on embedded networked sensor systems}}. \bibinfo{pages}{1--14}.
\newblock


\bibitem[Paulik et~al\mbox{.}(2021)]%
        {paulik2021federated}
\bibfield{author}{\bibinfo{person}{Matthias Paulik}, \bibinfo{person}{Matt Seigel}, \bibinfo{person}{Henry Mason}, \bibinfo{person}{Dominic Telaar}, \bibinfo{person}{Joris Kluivers}, \bibinfo{person}{Rogier van Dalen}, \bibinfo{person}{Chi~Wai Lau}, \bibinfo{person}{Luke Carlson}, \bibinfo{person}{Filip Granqvist}, \bibinfo{person}{Chris Vandevelde}, {et~al\mbox{.}}} \bibinfo{year}{2021}\natexlab{}.
\newblock \showarticletitle{Federated evaluation and tuning for on-device personalization: System design \& applications}.
\newblock \bibinfo{journal}{\emph{arXiv preprint arXiv:2102.08503}} (\bibinfo{year}{2021}).
\newblock


\bibitem[Peng et~al\mbox{.}(2020)]%
        {peng2020capuchin}
\bibfield{author}{\bibinfo{person}{Xuan Peng}, \bibinfo{person}{Xuanhua Shi}, \bibinfo{person}{Hulin Dai}, \bibinfo{person}{Hai Jin}, \bibinfo{person}{Weiliang Ma}, \bibinfo{person}{Qian Xiong}, \bibinfo{person}{Fan Yang}, {and} \bibinfo{person}{Xuehai Qian}.} \bibinfo{year}{2020}\natexlab{}.
\newblock \showarticletitle{Capuchin: Tensor-based gpu memory management for deep learning}. In \bibinfo{booktitle}{\emph{Proceedings of ACM ASPLOS}}. \bibinfo{pages}{891--905}.
\newblock


\bibitem[Sandler et~al\mbox{.}(2018)]%
        {sandler2018mobilenetv2}
\bibfield{author}{\bibinfo{person}{Mark Sandler}, \bibinfo{person}{Andrew Howard}, \bibinfo{person}{Menglong Zhu}, \bibinfo{person}{Andrey Zhmoginov}, {and} \bibinfo{person}{Liang-Chieh Chen}.} \bibinfo{year}{2018}\natexlab{}.
\newblock \showarticletitle{Mobilenetv2: Inverted residuals and linear bottlenecks}. In \bibinfo{booktitle}{\emph{Proceedings of IEEE CVPR}}. \bibinfo{pages}{4510--4520}.
\newblock


\bibitem[Shin et~al\mbox{.}(2022)]%
        {shin2022fedbalancer}
\bibfield{author}{\bibinfo{person}{Jaemin Shin}, \bibinfo{person}{Yuanchun Li}, \bibinfo{person}{Yunxin Liu}, {and} \bibinfo{person}{Sung-Ju Lee}.} \bibinfo{year}{2022}\natexlab{}.
\newblock \showarticletitle{FedBalancer: data and pace control for efficient federated learning on heterogeneous clients}. In \bibinfo{booktitle}{\emph{Proceedings of the 20th Annual International Conference on Mobile Systems, Applications and Services}}. \bibinfo{pages}{436--449}.
\newblock


\bibitem[Tu et~al\mbox{.}(2018)]%
        {tu2018your}
\bibfield{author}{\bibinfo{person}{Zhen Tu}, \bibinfo{person}{Runtong Li}, \bibinfo{person}{Yong Li}, \bibinfo{person}{Gang Wang}, \bibinfo{person}{Di Wu}, \bibinfo{person}{Pan Hui}, \bibinfo{person}{Li Su}, {and} \bibinfo{person}{Depeng Jin}.} \bibinfo{year}{2018}\natexlab{}.
\newblock \showarticletitle{Your apps give you away: distinguishing mobile users by their app usage fingerprints}.
\newblock \bibinfo{journal}{\emph{Proceedings of the ACM on Interactive, Mobile, Wearable and Ubiquitous Technologies}} \bibinfo{volume}{2}, \bibinfo{number}{3} (\bibinfo{year}{2018}), \bibinfo{pages}{1--23}.
\newblock


\bibitem[Wang et~al\mbox{.}(2018)]%
        {wang2018superneurons}
\bibfield{author}{\bibinfo{person}{Linnan Wang}, \bibinfo{person}{Jinmian Ye}, \bibinfo{person}{Yiyang Zhao}, \bibinfo{person}{Wei Wu}, \bibinfo{person}{Ang Li}, \bibinfo{person}{Shuaiwen~Leon Song}, \bibinfo{person}{Zenglin Xu}, {and} \bibinfo{person}{Tim Kraska}.} \bibinfo{year}{2018}\natexlab{}.
\newblock \showarticletitle{Superneurons: Dynamic GPU memory management for training deep neural networks}. In \bibinfo{booktitle}{\emph{Proceedings of ACM PPoPP}}. \bibinfo{pages}{41--53}.
\newblock


\bibitem[Wang et~al\mbox{.}(2022)]%
        {wang2022melon}
\bibfield{author}{\bibinfo{person}{Qipeng Wang}, \bibinfo{person}{Mengwei Xu}, \bibinfo{person}{Chao Jin}, \bibinfo{person}{Xinran Dong}, \bibinfo{person}{Jinliang Yuan}, \bibinfo{person}{Xin Jin}, \bibinfo{person}{Gang Huang}, \bibinfo{person}{Yunxin Liu}, {and} \bibinfo{person}{Xuanzhe Liu}.} \bibinfo{year}{2022}\natexlab{}.
\newblock \showarticletitle{Melon: Breaking the memory wall for resource-efficient on-device machine learning}. In \bibinfo{booktitle}{\emph{Proceedings of ACM MobiSys}}. \bibinfo{pages}{450--463}.
\newblock


\bibitem[{Wikipedia contributors}(ited)]%
        {6895997rule}
\bibfield{author}{\bibinfo{person}{{Wikipedia contributors}}.} \bibinfo{year}{Year the page was last edited}\natexlab{}.
\newblock \bibinfo{booktitle}{\emph{68–95–99.7 rule}}.
\newblock
\urldef\tempurl%
\url{https://en.wikipedia.org/wiki/68%E2%80%9395%E2%80%9399.7_rule}
\showURL{%
\tempurl}
\newblock
\shownote{Accessed: Access date}.


\bibitem[Xue et~al\mbox{.}(2024)]%
        {xue2024powerinfer}
\bibfield{author}{\bibinfo{person}{Zhenliang Xue}, \bibinfo{person}{Yixin Song}, \bibinfo{person}{Zeyu Mi}, \bibinfo{person}{Le Chen}, \bibinfo{person}{Yubin Xia}, {and} \bibinfo{person}{Haibo Chen}.} \bibinfo{year}{2024}\natexlab{}.
\newblock \showarticletitle{PowerInfer-2: Fast Large Language Model Inference on a Smartphone}.
\newblock \bibinfo{journal}{\emph{arXiv preprint arXiv:2406.06282}} (\bibinfo{year}{2024}).
\newblock


\bibitem[Zhang et~al\mbox{.}(2020)]%
        {zhang2020batchcrypt}
\bibfield{author}{\bibinfo{person}{Chengliang Zhang}, \bibinfo{person}{Suyi Li}, \bibinfo{person}{Junzhe Xia}, \bibinfo{person}{Wei Wang}, \bibinfo{person}{Feng Yan}, {and} \bibinfo{person}{Yang Liu}.} \bibinfo{year}{2020}\natexlab{}.
\newblock \showarticletitle{$\{$BatchCrypt$\}$: Efficient homomorphic encryption for $\{$Cross-Silo$\}$ federated learning}. In \bibinfo{booktitle}{\emph{2020 USENIX annual technical conference (USENIX ATC 20)}}. \bibinfo{pages}{493--506}.
\newblock


\bibitem[Zhang et~al\mbox{.}(2015)]%
        {zhang2015character}
\bibfield{author}{\bibinfo{person}{Xiang Zhang}, \bibinfo{person}{Junbo Zhao}, {and} \bibinfo{person}{Yann LeCun}.} \bibinfo{year}{2015}\natexlab{}.
\newblock \showarticletitle{Character-level convolutional networks for text classification}.
\newblock \bibinfo{journal}{\emph{Advances in neural information processing systems}}  \bibinfo{volume}{28} (\bibinfo{year}{2015}).
\newblock


\bibitem[Zhang et~al\mbox{.}(2018)]%
        {zhang2018shufflenet}
\bibfield{author}{\bibinfo{person}{Xiangyu Zhang}, \bibinfo{person}{Xinyu Zhou}, \bibinfo{person}{Mengxiao Lin}, {and} \bibinfo{person}{Jian Sun}.} \bibinfo{year}{2018}\natexlab{}.
\newblock \showarticletitle{Shufflenet: An extremely efficient convolutional neural network for mobile devices}. In \bibinfo{booktitle}{\emph{Proceedings of IEEE CVPR}}. \bibinfo{pages}{6848--6856}.
\newblock


\end{thebibliography}


\end{document}